\documentclass[12pt]{article}

\usepackage{amsmath,amsthm, amsfonts, amssymb, amsxtra, amsopn}
\usepackage{pgfplots}
\pgfplotsset{compat=1.13}
\usepackage{pgfplotstable}
\usepackage{graphicx,grffile}
\usepackage{multirow}
\usepackage{booktabs}
\usepackage{listings}
\usepackage{cmap}
\usepackage{colortbl}
\usepackage{adjustbox}
\usepackage{epsfig}

\usepackage[tableposition=top,font=small,skip=5pt]{caption}
\usepackage{subcaption}
\usepackage{makecell} 

\usepgfplotslibrary{external} 
\def\nest{{\mbox{\texttt{nesterov}}}}
\def\mom{{m}}

\PassOptionsToPackage{hyphens}{url}

\usepackage{hyperref}
\hypersetup{colorlinks=true,linkcolor=black,citecolor=black,urlcolor=blue,filecolor=black}
\hypersetup{pdfpagemode=UseNone,pdfstartview=}

\usepackage{enumitem}
\setlist[itemize]{noitemsep, topsep=0pt}

\advance\oddsidemargin by -0.22in
\advance\textwidth by 0.44in

\advance\topmargin by -0.275in
\advance\textheight by 0.55in

\def\eref#1{(\ref{#1})}

\long\def\symbolfootnotetext[#1]#2{\begingroup%
\def\thefootnote{\fnsymbol{footnote}}\footnotetext[#1]{#2}\endgroup}

\def\hh{\hspace*{0.25in}}

\def\cO{{\cal O}}

\def\zz{\phantom{0}}

\title{Hidden Markov Models with Momentum}

\author{Andrew Miller\footnotemark[1]\ \ \ 
Fabio Di Troia\footnotemark[1]\ \ \
Mark Stamp\footnotemark[1]\,\,\footnotemark[2]}

\begin{document}

\symbolfootnotetext[1]{Department of Computer Science, San Jose State University}
\symbolfootnotetext[2]{mark.stamp$@$sjsu.edu}

\maketitle

\abstract
Momentum is a popular technique for improving convergence rates during gradient descent.  In this research, we experiment with adding momentum to the Baum-Welch expectation-maximization algorithm for training Hidden Markov Models.  
We compare discrete Hidden Markov Models trained with and without momentum on English text and malware opcode data.
The effectiveness of momentum is determined by measuring the changes in model score and classification accuracy due to momentum.
Our extensive experiments indicate that adding momentum to Baum-Welch can reduce the number of iterations required for initial convergence during HMM training, particularly in cases where the model is slow to converge.  
However, momentum does not seem to improve the final model performance at a high number of iterations.

\section{Introduction}

Momentum is a popular extension to the gradient descent optimization algorithm for training machine learning models, being integrated into popular and widely used optimizers such as ADAM~\cite{gd_overview}.  Momentum can accelerate training by smoothing the effects of noisy gradients, adjusting the gradient step based on an exponentially decaying combination of past gradients.

This research applies the concept of momentum to another machine learning method: the Hidden Markov Model~(HMM).  HMMs are used to model Markov processes in which some or all states cannot be directly observed.  HMMs learn about these hidden states by observing a secondary sequence where each observation is dependent on the corresponding hidden state.  HMMs have been used for a wide variety of applications, including speech recognition~\cite{speechRecHMM}, biological sequences~\cite{YoonBio}, cryptanalysis~\cite{stampHMMCiphers}, and malware detection~\cite{dhanasekar}.

HMMs are often trained using the Baum-Welch algorithm to efficiently reestimate model parameters~\cite{stampHMM}.  Baum-Welch is a hill climb algorithm and will always converge to a local maximum, making it heavily dependent on the initialization of the HMM parameters for achieving good results.  A common solution is to perform a multitude of random restarts, training many HMMs with varying initializations and selecting the model with the best score.  However, this dramatically increases the amount of time required for training an effective HMM.  In this paper, we add a momentum term to the Baum-Welch algorithm.  If subsequent Baum-Welch updates trend in the same direction, momentum may also reduce training time by lowering the number of iterations required for convergence.  The addition of momentum may also cause the algorithm to overshoot local maxima, decreasing model score in favor of potential better final maximum.  On the other hand, momentum makes the Baum-Welch algorithm no longer a true hill-climb, and requires additional parameter tuning.

We use a classic English text problem as a test case for initial experiments comparing model scores of HMMs trained with and without momentum.  Our findings show that at early iterations, the addition of momentum to Baum-Welch reduces the number of iterations required to achieve a given score, speeding up training.  In particular, initializations which take a significant amount of time to begin converging show major reductions in training time.  However, once training slows, momentum is limited to minor changes in average model score.

Further experiments train HMMs on malware samples and observe the effects of momentum on model performance for malware classification.  The classification of malware into individual families can assist in learning features that can make it easier to detect difficult types of malware~\cite{malwarebook}.  HMMs are trained on sequences of opcodes extracted from malware executables from several malware families.  Samples are scored against these models and used for classification. These experiments demonstrate that the increases in model scores with momentum can translate into tangible improvements in model performance.

The remaining paper is organized as follows:
Section~\ref{chap:background} reviews related work and provides background for the technologies used in this paper.  
Section~\ref{chap:implementation} details how momentum is implemented for Baum-Welch training.  
Section~\ref{chap:results} discusses experiments comparing HMMs with and without momentum utilizing English text and malware opcode data, and analyzes the results. 
Finally, Section~\ref{chap:conclusion} summarizes the findings and suggests future directions for research.

\section{Background}\label{chap:background}

Our main focus in this section is to introduce the basic concepts of hidden Markov models.
We also discuss several topics that are related to momentum and the use of momentum
in HMM training, and we briefly consider the malware problem that forms the basis for
extensive experiments in Section~\ref{chap:results}, below.

\subsection{Hidden Markov Models}

A Markov process is a sequence in which the probability of the state at each position in the sequence depends solely on the state at the previous position.  Hidden Markov Models are statistical models capable of modelling these sequences in cases where the states are unable to be measured directly.  By using a secondary sequence of observable values which are dependent on the original sequence, an HMM can make predictions about the most probable states of the original state sequence.  Note that while there exists continuous variants of HMMs~\cite{599022,malwareGHMM}, only the discrete case is considered for this paper.

HMMs function under the assumption that the probability of an observation depends only on its corresponding state, and is independent of other observations and states.  Because the underlying state sequence is a Markov process, HMMs assume that the probability of a hidden state depends only on the previous hidden state.  
Higher order Markov processes that depend on more than one previous state exist, but are more complex and are not considered for this report.  

An HMM is composed of three matrices: the transition probability matrix, observation or emission probability matrix, and the initial state probability matrix.  The transition matrix determines the likelihood of transitioning from one state to another at each position in the sequence.  The observation matrix contains the likelihood of each observation being emitted by each state.  The initial state matrix contains the probability of each state as being the first state in the sequence.  The
HMM notation in Table~\ref{tab:HMMnotation} is used throughout this paper.

\begin{table}[!htb]
\caption{HMM notation~\cite{stampHMM}}\label{tab:HMMnotation}
\centering
\advance\tabcolsep by -2pt
\adjustbox{scale=0.85}{
\begin{tabular}{rcl}\midrule\midrule
$N$ &=& $\mbox{number of hidden states}$ \\
$M$ &=& $\mbox{number of observed states}$ \\
$\cO$ &=& $\mbox{observation sequence}$ \\
$T$ &=& $\mbox{length of } \cO$ \\
$Q$ &=& $\mbox{set of possible hidden states}$ \\
$V$ &=& $\mbox{set of possible observations}$ \\
$A$ &=& $\mbox{state transition probability matrix } (N\times N)$ \\
$B$ &=& $\mbox{emission/observation probability matrix } (N\times M)$ \\
$\pi$ &=& $\mbox{initial state probability } (1\times N)$ \\
$\lambda$ &=& $\mbox{model } (A,B,\pi)$ \\ \midrule\midrule
\end{tabular}
}
\end{table}

HMMs are useful for solving the following three problems, and efficient algorithms exist for each~\cite{stampHMM}.
\begin{enumerate}
\item\label{problem1} Given a model $\lambda(A,B,\pi)$ and an observation sequence $\cO$, find $P(\cO | \lambda)$.  This can also be thought of as scoring $\cO$ using $\lambda$.
\item\label{problem2} Given a model $\lambda(A,B,\pi)$ and an observation sequence $\cO$, find an optimal sequence of hidden states matching that observation sequence.  More explicitly, find the sequence with the maximum number of correct states according to the model.
\item\label{problem3} Given hyperparameters $N$ and $M$, and an observation sequence $\cO$, train a model $\lambda(A,B,\pi)$ which maximizes the probability of the observation sequence $\cO$. 
\end{enumerate}
This report is primarily concerned with Problem~\ref{problem3}: training a model to best fit some observation sequence.  However, the solutions to the other two problems are necessary for the training problem.    

\subsection{Problem 1: Score an Observation Sequence}\label{hmm_scoring}

Na\"{i}vely computing $P(\cO | \lambda)$ would require $O({N^{T}}T)$ operations~\cite{stampHMM}, so a more efficient algorithm is required. An improved solution to Problem 1 utilizes the following forward algorithm, reducing it down to $O({N^2}T)$ operations.


First, let $q_i$ be the hidden state at time $t$.  We define $\alpha_t(i)$ as the likelihood of the observation sequence up until time $t$ using the equation
\begin{equation*}
\alpha_t(i) = P(\cO_0,\cO_1,\ldots,\cO_{T-1}, x_t=q_i |\lambda)
\end{equation*}
The forward algorithm takes the following steps: 
\begin{enumerate}
\item Initialization: \\
for $i=0,1,\ldots,N-1$, set  
\[\alpha_0(i) = {\pi_i}{{b_i}(\cO_0)}\]
\item Recursion: \\
for $i=0,1,\ldots,N-1$ and $t=1,2,\ldots, T-1$, compute \\
\[\alpha_t(i) = \sum_{j=0}^{N-1} {\alpha_{t-1}}(j) a_{ji} b_{i}(\cO_t) \]
\item Completion: \\
\[P(\cO | \lambda) = \sum_{j=0}^{N-1} {\alpha_{T-1}}(j)\]
\end{enumerate}

However, as sequence length $T$ grows, the multiplication of a large number of small probabilities causes computational issues with underflow.  To solve this, a constant factor $c_t$ is introduced to scale the probability at each $t$.  The scaling constant $c_t$ at sequence position $t$ is computed as the sum of the $\alpha$'s of each state $i$ for $i = 0, 1, \ldots, N-1$ using 
\begin{equation*}
\label{constant_calc}
    c_t = \frac{1}{\displaystyle\sum_{i=0}^{N-1}{\alpha_{t}(i)}}
\end{equation*}
The $\alpha$ values are then scaled by these constants, avoiding underflow during future calculations by computing
\begin{equation*}
    \alpha_{t}(i) = c_{t}\alpha_{t}(i)
\end{equation*}

Finally, rather than using the probability $P(\cO | \lambda)$, scoring can be performed by computing the log likelihood using the scaling constants as 
\begin{equation*}
    \log_{}[P(\cO | \lambda)] = -\sum_{t=0}^{T-1}{\log_{}{c_t}}
\end{equation*}

\subsection{Problem 2: Most Likely Hidden State Sequence}

Finding the most likely sequence of hidden states requires determining the probability of each state occurring at each time step.  We dub these gammas, and define the probability of state $q_i$ occurring at time step $t$ as 
\begin{equation*}
\gamma_t(i) = P(x_t=q_i | \cO, \lambda)
\end{equation*}
Computing $P(x_t=q_i |\cO, \lambda)$ requires both the probability of the sequence up to $t$, as well as the probability of the remaining sequence following $t$.  The forward algorithm is used to compute $\alpha_t(i)$ for the sequence prior to $t$.  The similar \emph{backwards algorithm} is used to compute the probability for the remaining part of the sequence, defined as
\begin{equation*}
\beta_t(i) = P(\cO_{t+1},\cO_{t+2},\ldots\cO_{T-1}, x_t=q_i |\lambda)
\end{equation*}
\begin{enumerate}
\item for $i=0,1,\ldots,N-1$, set
\[\beta_{T-1}(i) = 1.0\]
\item for $i=0,1,\ldots,N-1$ and $t=T-2,T-3,\ldots,0$, compute \\
\[\beta_t(i) = \sum_{j=0}^{N-1} {\beta_{t+1}}(j) a_{ij} b_{j}(\cO_{t+1}) \]
\end{enumerate}
Then for a given state $i$ and time $t$, $\gamma_t(i)$ can be computed by
\[\gamma_t(i) = \frac{\alpha_t(i)\beta_t(i)}{P(\cO|\lambda)} \]
The most likely state $q_i$ at $t$ will be that with the maximum $\gamma_t(i)$.

\subsection{Problem 3: HMM Training}\label{Baum-Welch}

A common way of solving Problem~\ref{problem3} is the Baum-Welch reestimation algorithm.  Baum-Welch is a version of the Expectation Maximization(EM) algorithm for maximum likelihood estimation~\cite{acceleratedEM}.  The EM algorithm works by iteratively using the model parameters to compute a probability distribution for latent variables, then using that distribution to update the parameters~\cite{acceleratedEM}.  Baum-Welch uses this technique to efficiently train the $A$, $B$, and $\pi$ matrices of an HMM by iteratively adjusting these HMM parameters to best fit the observation sequence.  Baum-Welch is a hill climb algorithm, and will therefore always converge to a local maximum.

While not explicitly a part of Baum-Welch, the first step in training an HMM is to initialize the model's $A$, $B$, and $\pi$ matrices.  This can be done randomly, or by using prior knowledge of the problem.  Stamp~\cite{stampHMM} recommends random values close to~$1/N$.  The matrices must be made row stochastic after initialization.  

Because hill climb algorithms will only converge to a local maximum, numerous random restarts are often performed when training HMMs.  By using different random initializations for each restart, more of the problem surface can be covered in search of a global maximum.  This makes the efficiency of Baum-Welch important, as many models may be trained and only the best selected for use.  

The first step of Baum-Welch is to compute both the $\alpha$'s and $\beta$'s for each time step $t$ using the forward and backward algorithms from the previous solutions.  These are then used to calculate $\gamma_t(i)$ as in Problem 2 for each $t$ and state $q_i$.

Baum-Welch also computes the so-called di-gammas, $\gamma_{t}(i,j)$, which is defined as the probability of being in state $q_i$ at time $t$ and transitioning to state $q_j$ at $t+1$.  The di-gammas can be computed as
\begin{equation*}
\gamma_t(i,j) = \frac{\alpha_t(i) a_{ij} b_j(\cO_{t+1}) \beta_{t+1}(j)}{P(\cO|\lambda)}
\end{equation*}

Finally, the model parameters $A$, $B$, and $\pi$ can be reestimated using the gammas and di-gammas.  
The new probability $a_{ij}$ is found by taking the predicted number of transitions from state $q_i$ to state $q_j$ and dividing by the total number of transitions out of $q_i$.
Similarly, the new probability $b_i(j)$ is the predicted number of times the model emits observation $\cO_j$ when at state $q_i$, divided by the number of times the model is in that state~\cite{stampHMM}.  These are computed using the following equations:
\begin{equation}\label{bw_update}
\begin{array}{l}
\mbox{for } i=0,1,\ldots,N-1\\
\hh \pi_{i} = \gamma_0(i)\\
\mbox{for } i=0,1,\ldots,N-1 \mbox{ and } j=0,1,\ldots,N-1\\ 
\hh a_{ij} = \displaystyle\sum_{t=0}^{T-2}{\gamma_t(i,j)} \bigg/ \displaystyle\sum_{t=0}^{T-2}{\gamma_t(i)} \\
\mbox{for } i=0,1,\ldots,N-1 \mbox{ and } j=0,1,\ldots,M-1 \mbox{ where } \cO_t = j \\
\hh b_i(j) = \displaystyle\sum_{t=0, \cO_t=j}^{T-1}{\gamma_t(i)} \bigg/ \displaystyle\sum_{t=0}^{T-1}{\gamma_t(i)}
\end{array}
\end{equation}
To summarize the training Baum-Welch training process is as follows:
\begin{enumerate}
    \item Compute $\alpha$'s, $\beta$'s, $\gamma$'s, and di-gammas using the current $A$, $B$, and~$\pi$ matrices 
    	and the observation sequence~$\cO$.
    \item Reestimate $A$, $B$, and~$\pi$ using the~$\gamma$'s and di-gammas.
    \item Repeat for a specified number of iterations, or until~$P(\cO|\lambda)$ does not increase significantly,
    	or other stopping criteria is met
\end{enumerate}

\subsection{Gradient Descent}

Gradient descent(GD) is an algorithm for iteratively optimizing a function using its first-order derivative~\cite{Du_2019}.  It follows the principle that by continuously moving in the direction of steepest descent, one will eventually reach a local minimum.  Given a function and set of parameters, at each iteration gradient descent computes the gradient of the function at that point, then updates the parameters by taking a step in the direction of the gradient.  Note that when performing gradient descent, the direction of the step is the negative of the gradient.  The term gradient ascent is used for algorithms which move in the direction of the gradient.  The size of the step is called the learning rate $\eta$.  A learning rate that is too large can cause the algorithm to overshoot a minimum, while a too-small learning rate can cause it to undershoot, requiring more iterations to reach a minimum.  Once a minimum is reached with a gradient of 0, the algorithm is complete.  

Given learning rate $\eta$, an objective function $f$ to be minimized, model parameters ${\theta_t}$ at step $t$ are computed as
\begin{equation*}\label{eq_gd}
    {\theta_{t}} = {\theta_{t-1}} - {\eta}{\Delta}f({\theta_{t-1}}) 
\end{equation*}

\subsection{Momentum in Gradient Descent}

The gradient descent algorithm may struggle when near saddle points or ravines~\cite{gd_overview}.  The algorithm may get stuck and be unable to converge, or converge slowly towards the optimum while oscillating between the sides of a ravine.  Momentum is a popular modification to classic gradient descent used to improve performance these cases and accelerate training.  The momentum term carries over a portion of previous gradients vectors, making the direction at a given step a combination of current and recent gradients~\cite{sutskever2013}.  

Momentum in gradient descent can be thought of as similar to momentum in Newtonian physics~\cite{qianMomentum1999}.  As a an object rolls downwards on a slope, it accelerates, gaining velocity in the direction of the slope.  Upon reaching the base of the slope, it will continue moving until its momentum is exhausted.  If the object encounters a small hill with sufficient momentum, it will continue over said hill while losing momentum.  Momentum in GD behaves similarly; although the gradient at each individual step may differ, movement in a general direction will build momentum in that direction.  This can result in faster convergence in ravines, as well as potentially overshooting to escape a local minimum.  However, if momentum is too high, it can overshoot a good minimum, resulting in worse performance.

\begin{figure}[!htb] 
\centering
\includegraphics[width=0.8\textwidth]{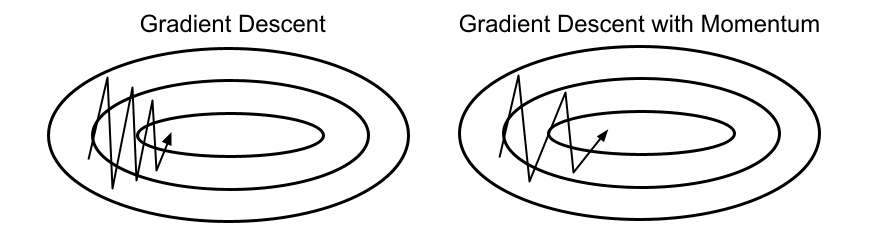}
\caption{Simplified gradient descent with and without momentum}\label{fig:momentumPhysics}
\end{figure}

The classic version of momentum is calculated as a sum of exponentially decaying previous gradients.  This is done using a momentum factor $\mu$.  $\mu$ is a hyperparameter defining how much momentum should be carried over at each step with a range of~$[0,1)$.  A $\mu$ of $0$ is equivalent to standard gradient descent without momentum.  $\mu$ is often set around 0.9~\cite{Du_2019,gd_overview}.  Adjusting the gradient descent algorithm to include the momentum term gives
\begin{equation*}
\label{eq_momentum}
\begin{aligned}
{\theta_{t}} &= {\theta_{t-1}} - {\eta}{\Delta}f({\theta_{t-1}}) + {\mu}{v_{t-1}} \\
v_{t} &= {\mu}{v_{t-1}} - {\eta}{\Delta}f({\theta_{t-1}})
\end{aligned}
\end{equation*}

\subsubsection{Nesterov Accelerated Gradient}

Nesterov momentum or Nesterov accelerated gradient(NAG)~\cite{nesterov1983}, is a popular alternative to the usual momentum algorithm.  NAG outperforms the standard momentum implementation in many situations due to increased stability and responsiveness~\cite{sutskever2013}.  NAG is very similar to momentum, but reverses the order of operations: adding the momentum vector first, then computing the gradient from the new point.  NAG can be formulated as
\begin{equation*}
\label{eq_NAG}
\begin{aligned}
{\theta_{t}} &= {\theta_{t-1}} - {\eta}{\Delta}f({\theta_{t-1}} + {\mu}{v_{t-1}}) + {\mu}{v_{t-1}} \\
v_{t} &= {\mu}{v_{t-1}} - {\eta}{\Delta}f({\theta_{t-1}} + {\mu}{v_{t-1}})
\end{aligned}
\end{equation*}

\begin{figure}[!htb]
\centering
\includegraphics[width=0.7\textwidth]{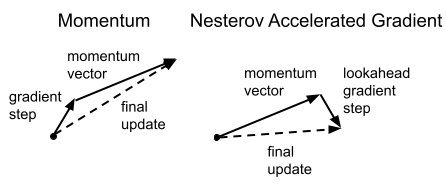}
\caption{Momentum update vs Nesterov update}\label{ref:MOMNAG}
\end{figure}

Figure~\ref{ref:MOMNAG} illustrates how the different gradient at the new position can change the final update vector, even with an identical momentum vector.  This look-ahead gradient used by NAG can be thought of as correcting the change in position due to momentum.  In the case where the momentum vector points in a poor direction, NAG will produce a better update vector without waiting until the next iteration for a correction.  The difference may be small, but compounds over many iterations~\cite{sutskever2013}.  

\subsection{Parameterized EM}\label{paramEM}

While momentum is commonly utilized by practical applications utilizing gradient descent, it does not seem to have been explicitly studied for EM-based algorithms.
However, Xu~\cite{xuMEM} discusses a modified EM algorithm which scales the magnitude of each EM update step using a scaling factor $\eta$.  While Xu dubs this the Momentum EM algorithm~(MEM), the algorithm is closer to EM with a learning rate than EM with momentum.  $\eta$ functions similarly to the learning rate factor in gradient descent, scaling the step taken at a given iteration.  At $\eta=1$, MEM is equivalent to EM.  Xu proves that the MEM algorithm is capable of improving convergence rates, and suggests a heuristic for choosing $\eta$.  For $\eta > 0$ and parameters $\theta$ at time $t$, MEM can be described as
\begin{equation*}\label{eq_paramEM}
\begin{aligned}
{\theta_{t+1}} &= {\theta_{t}} + {\eta}\Delta\theta_{t}
\end{aligned}
\end{equation*}
This approach can be described as parameterized EM~\cite{acceleratedEM}, and is capable of improving convergence speed when close to a solution~\cite{acceleratedEM}.  Parameterized EM could be considered momentum for just the previous time step.  

Furthermore, Xu and Jordan~\cite{xuJordan} demonstrate a connection between EM algorithms and gradient descent for Gaussian mixtures, showing that the EM step can be related to the gradient using a projection matrix.  Based on these findings and the potential relationship between gradient descent and EM~\cite{xuJordan,xuMEM}, it seems plausible that momentum could be applied to the Baum-Welch EM algorithm for training HMMs.

\subsection{HMMs for Malware Classification}

There have been several experiments utilizing HMMs for malware classification.  In~\cite{annachhatre}, models were trained using opcode sequences extracted from executables generated by several standard compilers, along with hand-written assembly and executables produced by two metamorphic malware generators.  Malware samples from the Malicia dataset~\cite{malicia} were then scored against each model.  K-means clustering was used on the resulting combination of scores to produce predicted groupings of the Malicia samples.  The HMM clustering method was able to classify the malware samples with reasonable accuracy, despite not being trained on specific families.  Models were trained for 800 iterations with $N=2$, as it was found that the number of hidden states did not affect classification.  

The authors of~\cite{kale} and~\cite{Chandak2021} each experimented with classification based on the models themselves, using the flattened $B$ matrix of each model as a feature vector.  This method was called HMM2Vec after the word embedding technique Word2Vec.  As hidden state order is not guaranteed to be consistent between HMMs, matrices were rearranged in an attempt to maintain a consistent state order for the final vectors.  Both found that HMM2Vec features produced accuracy above 90\% when classifying using support-vector machines~(SVMs), random forests, and neural networks.  k-nearest neighbors performed similarly for~\cite{Chandak2021}, but only reached 79\% accuracy for~\cite{kale}.  In addition,~\cite{Chandak2021} compares HMM2Vec feature vectors to vectors of scores on HMMs trained on each family.  While score features performed similarly with the neighborhood-based classifiers, the multilayer perceptron and SVM accuracy dropped to 44\% and 79\% accuracy respectively.  
 
A similar approach to constructing feature vectors was used by~\cite{singh} for classification via clustering of malware families.  HMMs were trained on opcode sequences of length $10{,}000$ for only 50 iterations, then clustered using k-means and k-medioids.  The effectiveness of this approach varied by family, with some families being well clustered and others being highly split.

In~\cite{boostedHMMs}, the classification accuracy of HMMs trained with multiple random restarts was compared to that of combinations of multiple HMMs using Adaboost.  In these tests, boosting showed little improvement over performing a similar number of restarts except in the most challenging cases.  Classification was performed by scoring samples against trained models for each family, rather than using the HMM matrices as features for another model.  

Finally,~\cite{malwareGHMM} compared discrete and continuous HMMs using Gaussian mixture model-hidden Markov models(GMM-HMMs).  For opcode sequences, GMM-HMMs performed similarly to discrete HMMs, while requiring additional tuning.  However, GMM-HMMs proved superior in the case of continuous data using entropy-based features. 

\section{Implementation}\label{chap:implementation}

In this section we introduce two types of momentum. We then consider the issue of missing observations, which leads us to the topic of smoothing.
 
\subsection{Momentum for Baum-Welch} 

The Baum-Welch version of momentum is based on momentum in gradient descent.  In gradient descent, the current update is a product of the learning rate and current gradient, and momentum is a function of these recent updates~\cite{gd_overview}.  However, due to the lack of a gradient in Baum-Welch, the update step is instead computed using the difference in model parameters before and after Baum-Welch reestimation.  This could be considered a sort of discrete gradient.  The modified Baum-Welch algorithm with momentum can be summarized as 
\begin{enumerate}
    \item Run Baum-Welch and reestimate $A$, $B$, $\pi$ as normal
    \item{\label{bw_m_step2} Compute the difference between parameters before and after reestimation}
    \item Add the current momentum to the reestimated matrices
    \item Update momentum using the difference found in step~\ref{bw_m_step2}
    \item Repeat for each iteration
\end{enumerate}

Momentum is tracked individually for each parameter.  Three matrices are created, corresponding to the $A$, $B$, and $\pi$ matrices respectively, with matching dimensions.  Each momentum matrix element is initialized to 0 at the start of training, as no momentum exists prior to the first iteration.  A hyperparameter $m$ is used to control the amount of momentum carried over at each iteration.  

The Baum-Welch update with momentum can be formulated as follows.  Let $v_t$ be the current momentum at time $t$, with $v(A)$, $v(B)$, $v(\pi)$ being the momentum matrices for a model $\lambda(A,B,\pi)$.

First, the usual Baum-Welch update is performed.  We define $F(\lambda)$ as the function performing the Baum-Welch update described in Section~\ref{bw_update} for a model $\lambda(A,B,\pi)$.  The Baum-Welch update at iteration~$t$ can then be written as
\begin{equation*}
\lambda_{t} = F(\lambda_{t-1})
\end{equation*}

Next, the change in parameters before and after reestimation is computed using
\begin{equation*}
\Delta{\lambda} = \lambda_{t} - \lambda_{t-1}
\end{equation*}

The current momentum $v$ is then added to the reestimated parameters to produce the final updated model parameters
\begin{equation*}
\lambda'_{t} = \lambda_{t} + v_{t-1}
\end{equation*}

Finally, the momentum vector $v$ is updated as a function of momentum hyperparameter $m$ and the change in parameters prior to the inclusion of momentum by
\begin{equation*}
v_t = m (v_{t-1} + \Delta{\lambda})
\end{equation*}

For example, the full update for a parameter $a_{ij}$ is computed as
\begin{equation*}\label{a_update_example}
\begin{aligned}
\Delta{a}_{ij} &= 
\Biggl(
\displaystyle\sum_{t=0}^{T-2}{\gamma_t(i,j)}
\Big/
\displaystyle\sum_{t=0}^{T-2}{\gamma_t(i)}
\Biggr)
- a_{ij} \\ 
a_{ij} &= a_{ij} + \Delta{a}_{ij} \\
a'_{ij} &= a_{ij} + v(A)_{ij} \\
v(A)_{ij} &= m(v(A)_{ij} + \Delta{a}_{ij})
\end{aligned}
\end{equation*}

Following the addition of momentum to each of the $A$, $B$, and $\pi$ matrices, an additional step is required to fix the updated values.  Because negative momentum may cause a parameter to become less than zero, any negative values are changed to a small positive value $\varepsilon$ close to 0.  This removes any negative probabilities while also avoiding adding zero probabilities to the model.  Finally, to ensure each matrix remains row stochastic, each row of the $A$, $B$, and $\pi$ matrices is normalized to make the final probabilities sum to 1.  

Finally, it is worth noting that with the addition of momentum, the Baum-Welch algorithm is no longer a true hill-climb.  Momentum may cause the algorithm to overshoot a local maximum resulting in a potential decrease in model score.  
     
\subsection{Nesterov Momentum} 

The Nesterov accelerated gradient approach to momentum is implemented similarly, but with some changes to when the momentum is applied.  Rather than adding and updating momentum at the end of each iteration, the momentum vector is added at the start of each iteration, prior to the execution of Baum-Welch.  The Nesterov momentum implementation can be summarized as
\begin{enumerate}
    \item Add momentum to the matrices
    \item Run Baum-Welch and reestimate $A$, $B$, $\pi$ from these updated matrices
    \item Compute the difference between parameters before and after reestimation
    \item Update momentum using the difference
    \item Repeat for each iteration
\end{enumerate}

For a model~$\lambda$, momentum matrices $v$, momentum hyperparameter $m$, and Baum-Welch update~$F$, Nesterov momentum is computed at iteration~$t$ by
\begin{equation*}
\begin{aligned}
\widehat{\lambda}_{t} &= \lambda_{t-1} + v_{t-1} \\
\lambda_{t} &= F(\widehat{\lambda}_{t}) \\
\Delta{\lambda} &= \lambda_t - \lambda_{t-1} \\
v_t &= m (v_{t-1} + \Delta{\lambda})
\end{aligned}
\end{equation*}

As with the base momentum implementation, any negative probabilities are replaced with a small positive $\varepsilon$, and each row is normalized after applying momentum.  

\subsection{Smoothing} 

Issues can occur with the Baum-Welch algorithm when zero probabilities occur in the model.  Zero probabilities can cause division by zero errors and eliminate potential observation sub-sequences that did not occur in training, but may occur when using the trained model.  For example, say a model $\lambda$ is trained on an observation sequence which does not contain a specific observation.  This will likely result in zero chance of any state producing that observation in the final model.  Scoring another test sequence which does contain the observation may then cause an error, such as a division by zero when computing constants as in~\eref{constant_calc}.   

To prevent these issues, an option for additive smoothing is included in the HMM implementation.  Additive smoothing works by adding some designated value to each count, ensuring each count is greater than zero when computing probabilities.  The larger the smoothing value, the more uniform each probability becomes.  Smoothing is performed during the update step of Baum-Welch, changing the matrix update functions for a smoothing value of~$s$ to
\begin{align*}   
    \pi_{i} &= \frac{\gamma_0(i) + s}{Ns} \\
    a_{ij} &= \Biggl(s + \displaystyle\sum_{t=0}^{T-2}{\gamma_t(i,j)}\Biggr)
    \Big/
    \Biggl(Ns + \displaystyle\sum_{t=0}^{T-2}{\gamma_t(i)}\Biggr) \\
    b_i(j) &= \Biggl(s + \displaystyle\sum_{t=0, \cO_t=j}^{T-1}{\gamma_t(i)}\Biggr)
    \Big/
    \Biggl(Ms + \displaystyle\sum_{t=0}^{T-1}{\gamma_t(i)}\Biggr) 
\end{align*}
For the multi-sequence case, the smoothing value is applied to the total count across all observation sequences, rather than per sequence.  Smoothing is applied prior to any momentum updates.  By smoothing in this manner, zero probabilities are avoided while keeping the matrices row stochastic.

\section{Experiments and Results}\label{chap:results}

In this section, we provide empirical results of HMMs trained with momentum. 
First, we consider the use of momentum in a classic HMM English text model. We then present
extensive experimental results where HMMs are applied to the malware classification
problem.

\subsection{Momentum and English Text}\label{momentum_results}

We begin by training HMMs on English text as an initial test case for Baum-Welch with momentum.  This English text
problem was first discussed in~\cite{CaveNeu} and appears in the tutorial~\cite{stampHMM}, for example, where it is 
used to introduce some of the key concepts of HMMs.

We measure the effectiveness of momentum by comparing training scores with and without momentum, using otherwise
identical hyperparameters.  English text samples are constructed by extracting character sequences of length~$T$ from the Brown Corpus of English text~\cite{BrownCorpus}.  
A vocabulary consisting of alphabetic characters~A through~Z plus word space is used for all experiments.  
Upper and lowercase characters are considered identical, for a total of of~$M=27$ possible observations.  
The corpus version used does not include spaces between words separated by line breaks, so an additional space character is added at the end of each line.  Any metadata sections and characters are excluded.

For our experiments, HMMs are trained with various momentum values and compared to baseline HMMs trained without momentum, using the same parameters and initializations.  Unless otherwise specified, models are initialized randomly using a continuous uniform distribution, then normalized to ensure each matrix is row stochastic. Each model is trained for~500 iterations with~100 random restarts.  
The number of hidden states~$N=27$ was chosen as the default for tests to match the number of vocabulary characters used.  

\begin{figure}[!htb]
\centering
\begin{subfigure}{.385\textwidth}
  \centering
  \includegraphics[width=1.0\linewidth]{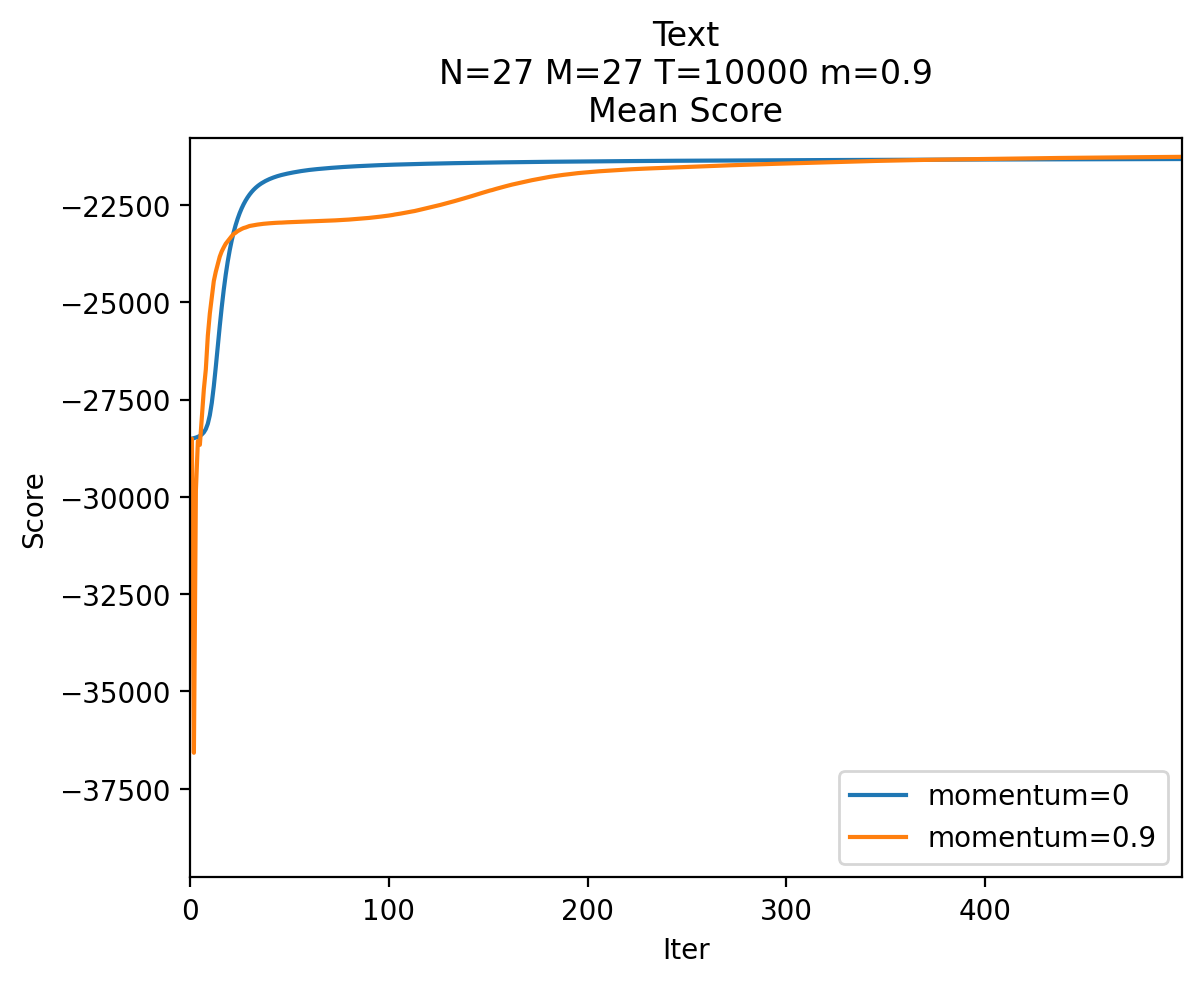}
  \label{fig:sfig1}
\end{subfigure}%
\begin{subfigure}{.385\textwidth}
  \centering
  \includegraphics[width=1.0\linewidth]{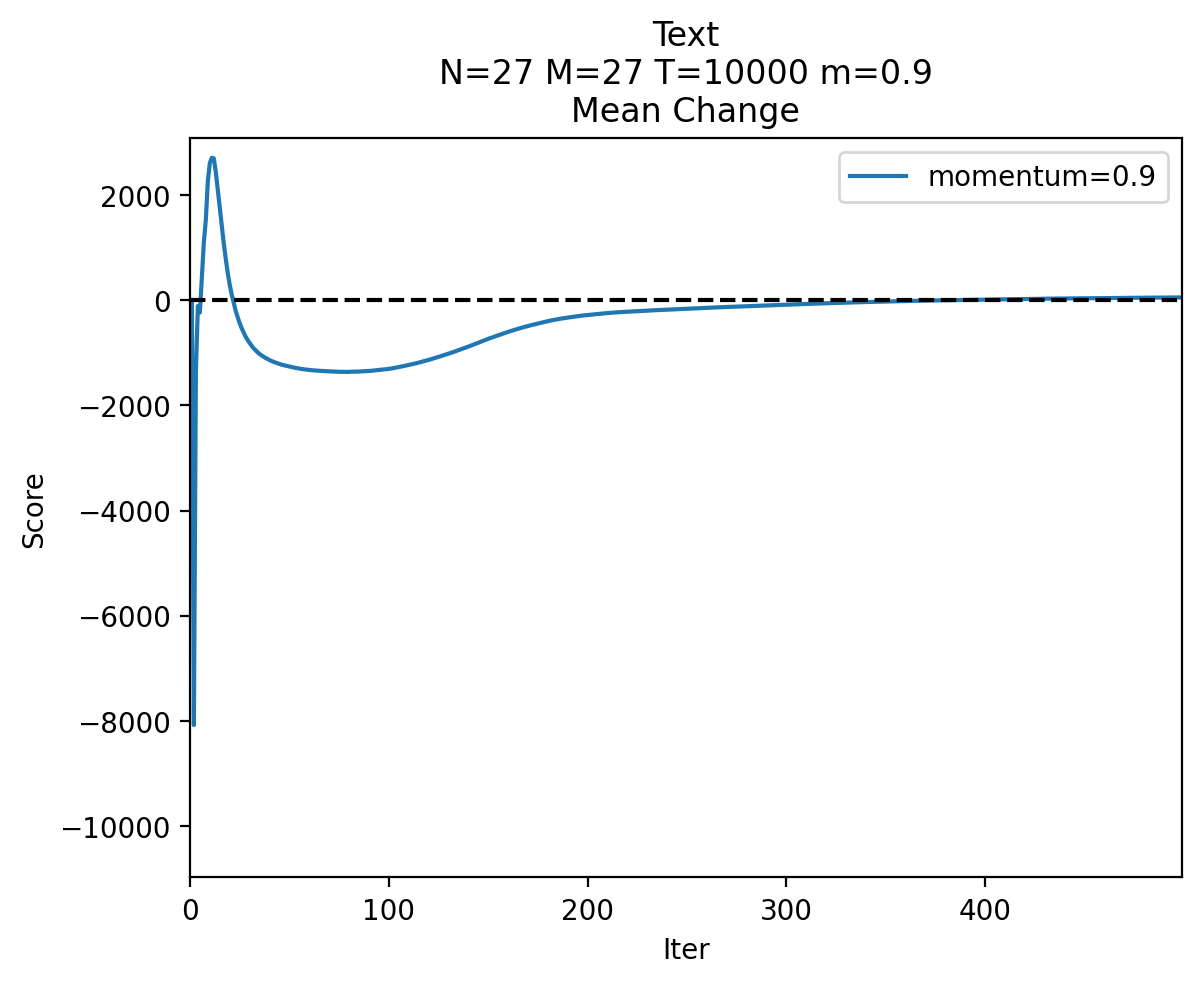}
  \label{fig:sfig2}
\end{subfigure}
\begin{subfigure}{.385\textwidth}
  \centering
  \includegraphics[width=1.0\linewidth]{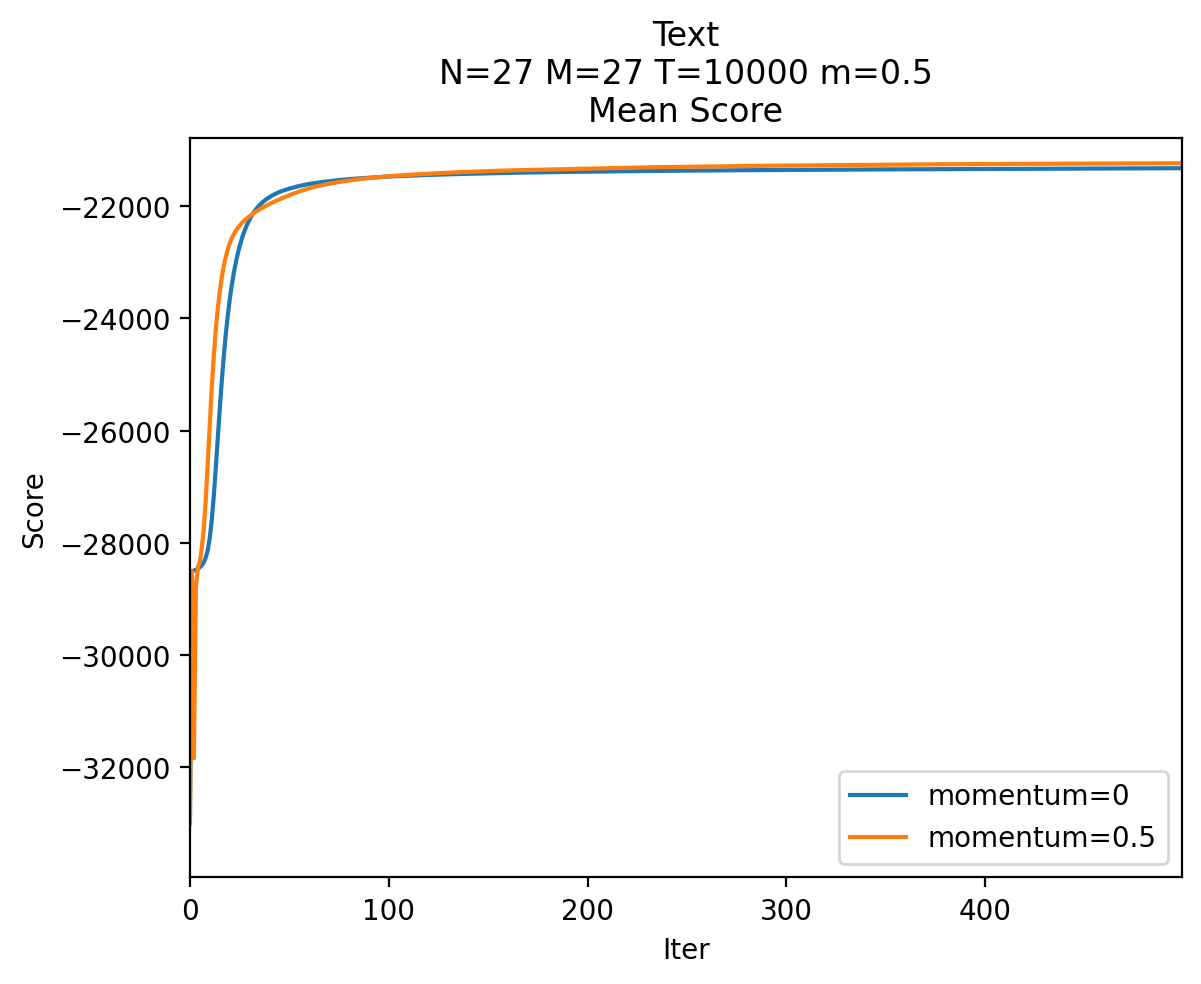}
  \label{fig:sfig3}
\end{subfigure}%
\begin{subfigure}{.385\textwidth}
  \centering
  \includegraphics[width=1.0\linewidth]{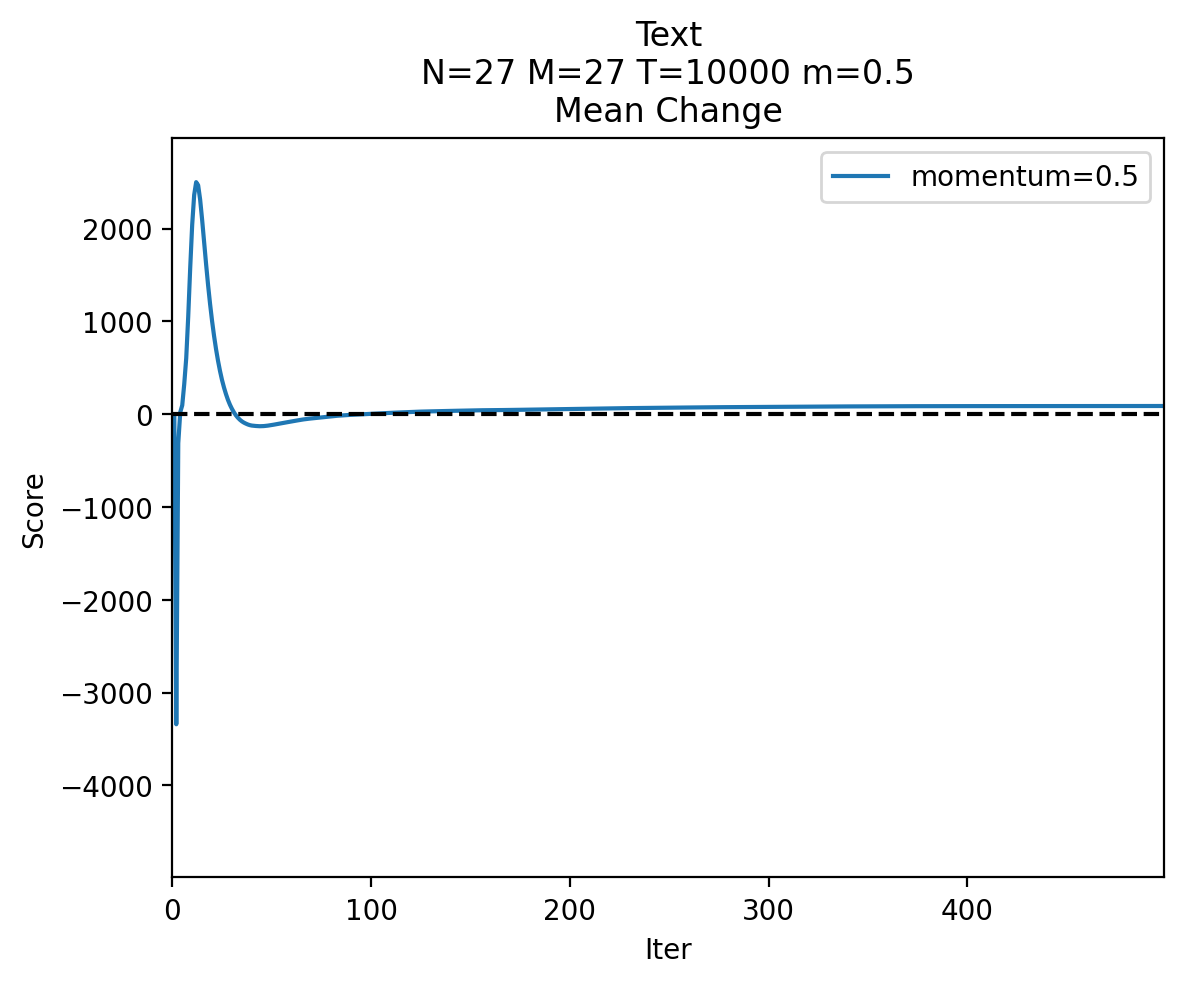}
  \label{fig:sfig4}
\end{subfigure}
\begin{subfigure}{.385\textwidth}
  \centering
  \includegraphics[width=1.0\linewidth]{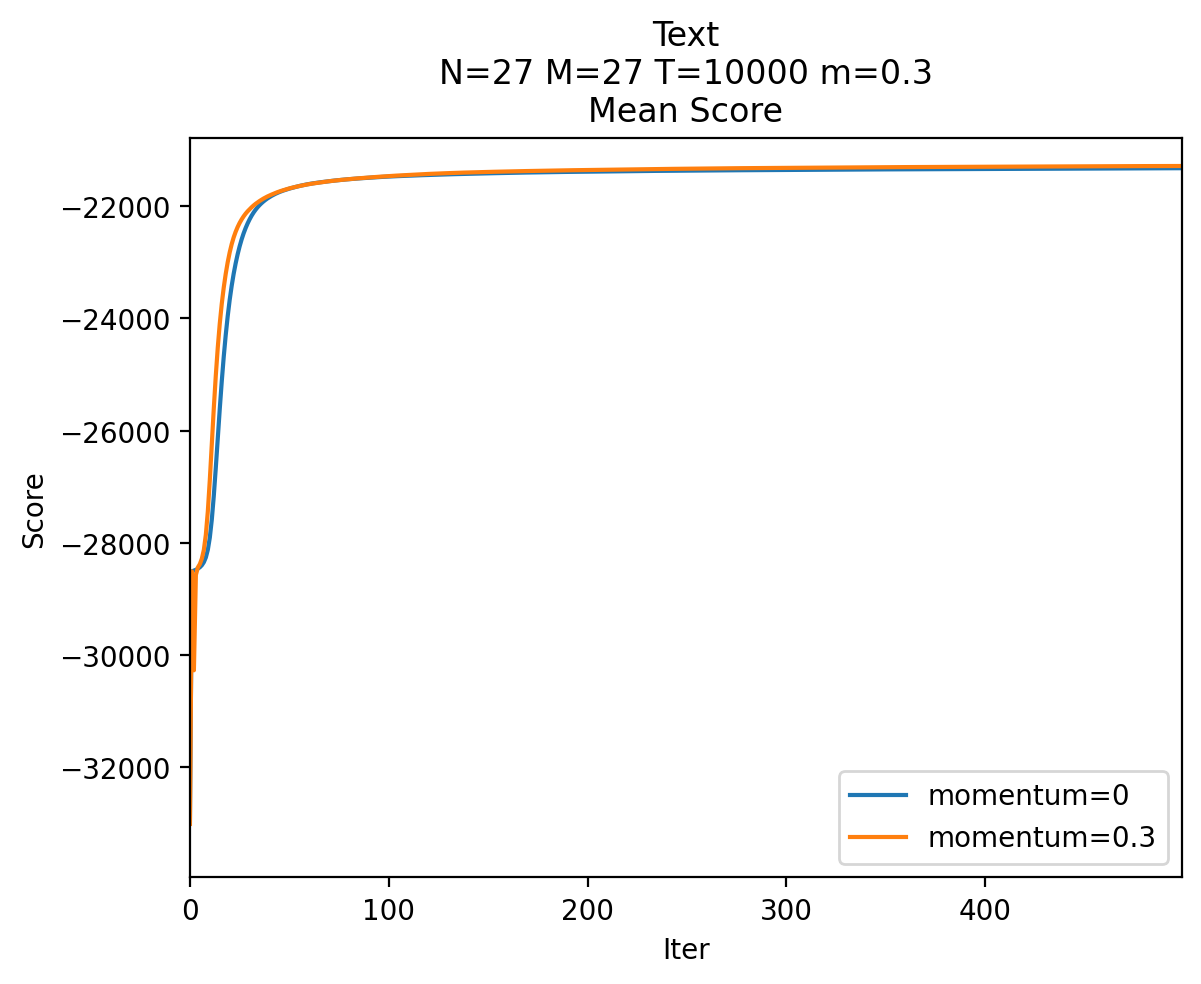}
  \label{fig:sfig5}
\end{subfigure}%
\begin{subfigure}{.385\textwidth}
  \centering
  \includegraphics[width=1.0\linewidth]{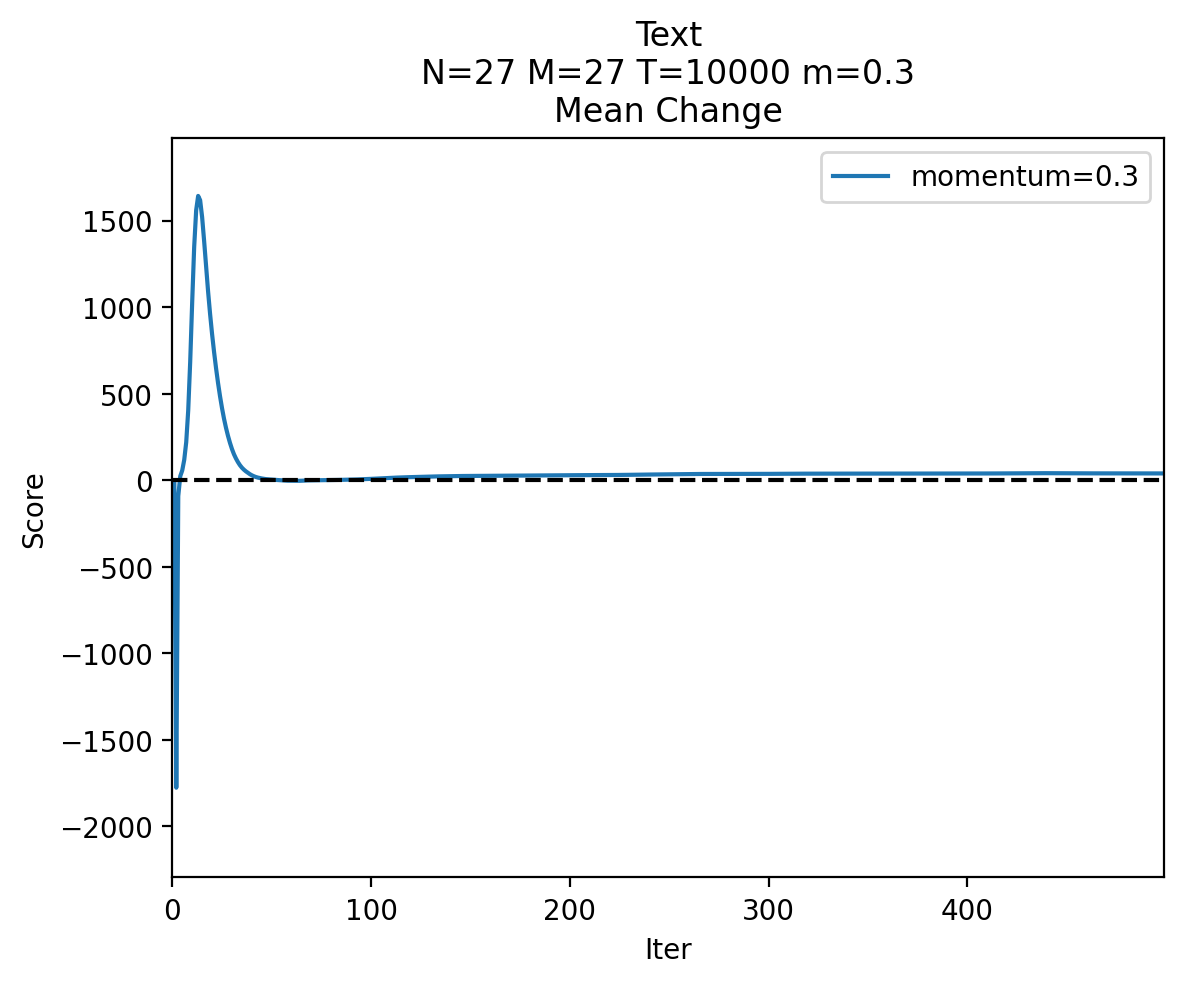}
  \label{fig:sfig6}
\end{subfigure}
\caption{Change in score with momentum for $T=10{,}000$}
\label{fig:text_momentum_T=10000}
\end{figure}

Figure~\ref{fig:text_momentum_T=10000} demonstrates the mean difference in score across~100 restarts 
for~$T=10{,}000$ with varying levels of momentum.  On average, models with momentum converge slightly faster than those without, with similar or slightly improved final scores.  However, higher momentum values tend to overshoot 
as convergence slows, resulting in worse scores for some number of iterations.  
The higher the momentum, the larger the overshoot period and the longer it takes to recover.  The high momentum value of~$m=0.9$ results in a larger 
increase in mean score during early iterations, but a lower momentum of~$m=0.3$ produces a more stable result.  

\begin{figure}[!htb]
\centering
\begin{subfigure}{.385\textwidth}
  \centering
  \includegraphics[width=1.0\linewidth]{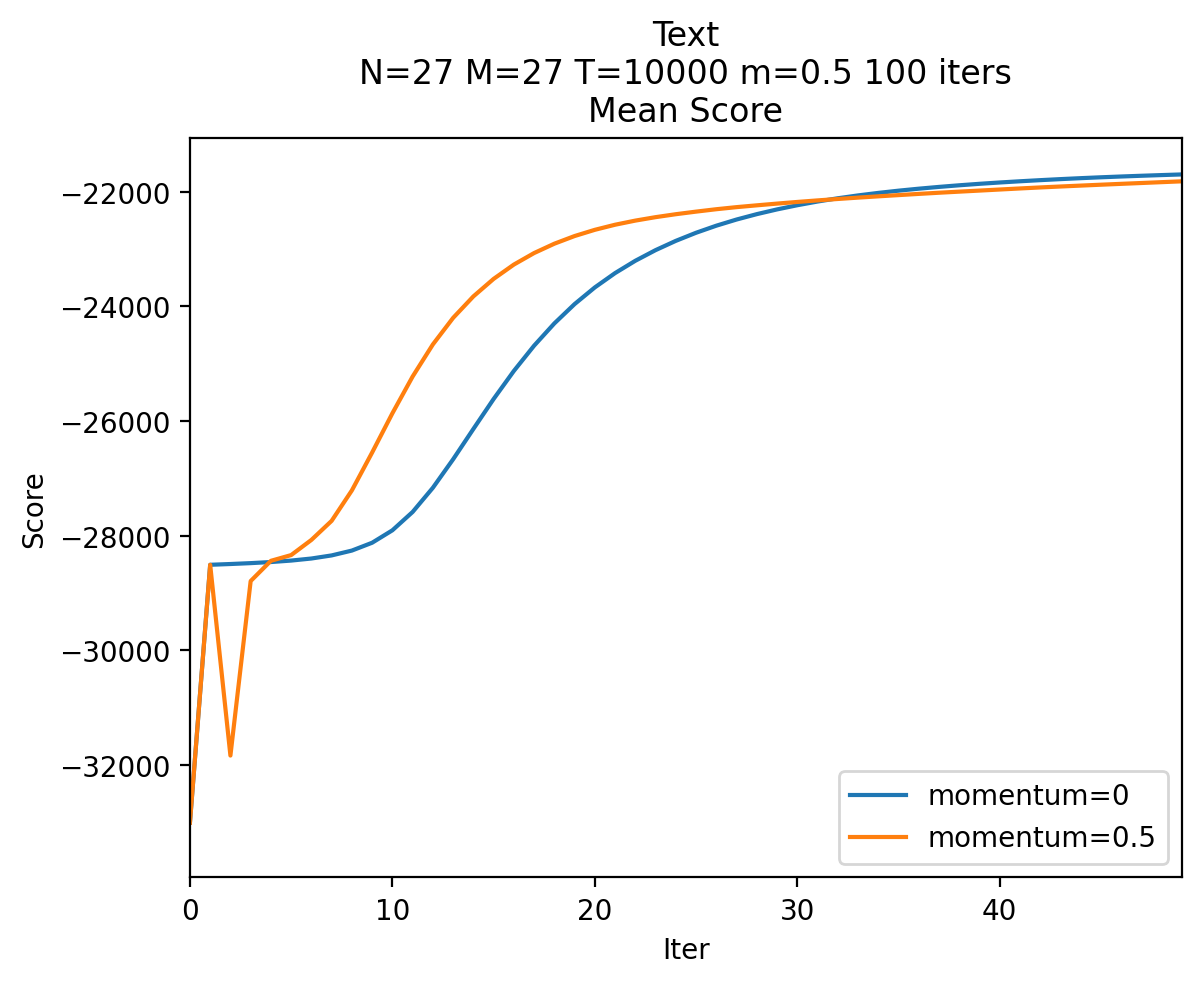}
\end{subfigure}%
\begin{subfigure}{.385\textwidth}
  \centering
  \includegraphics[width=1.0\linewidth]{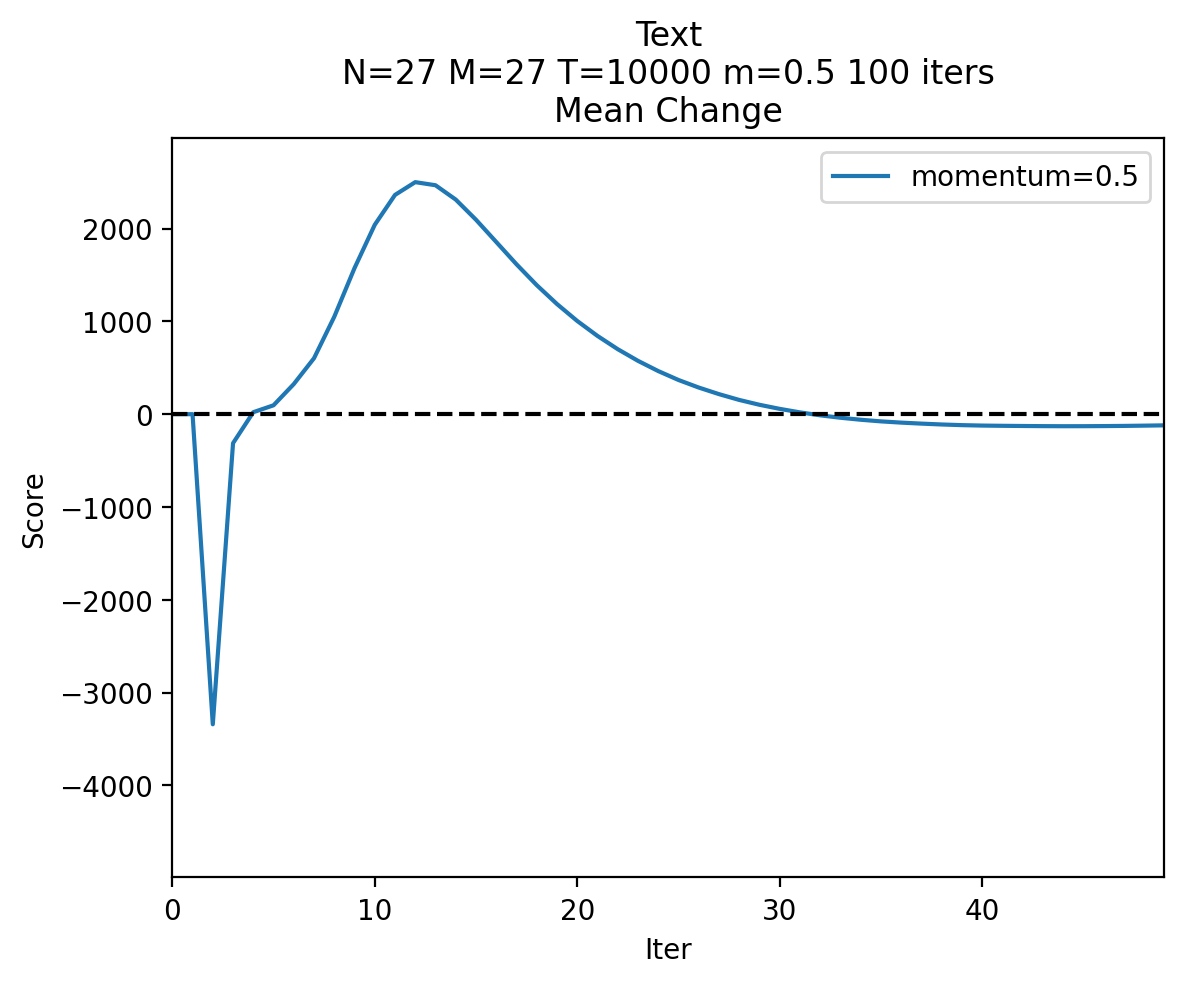}
\end{subfigure}
\caption{First 50 iterations for $T=10{,}000$ with $m=0.5$}
\label{fig:text_momentum_50iters}
\end{figure}

Also of note is the large dip in momentum scores during the first few iterations, which quickly recovers.  Figure~\ref{fig:text_momentum_50iters} emphasizes this by showing only the first fifty iterations for~$m=0.5$.  
At iteration~0, the models with and without momentum perform the same, as no momentum has been generated.  
However, as with the later overshoot, the large changes in the model during the first iterations creates excessive momentum, causing a major decrease in score for one iteration. 

\begin{figure}[!htb]
\centering
\begin{subfigure}{.385\textwidth}
  \centering
  \includegraphics[width=1.0\linewidth]{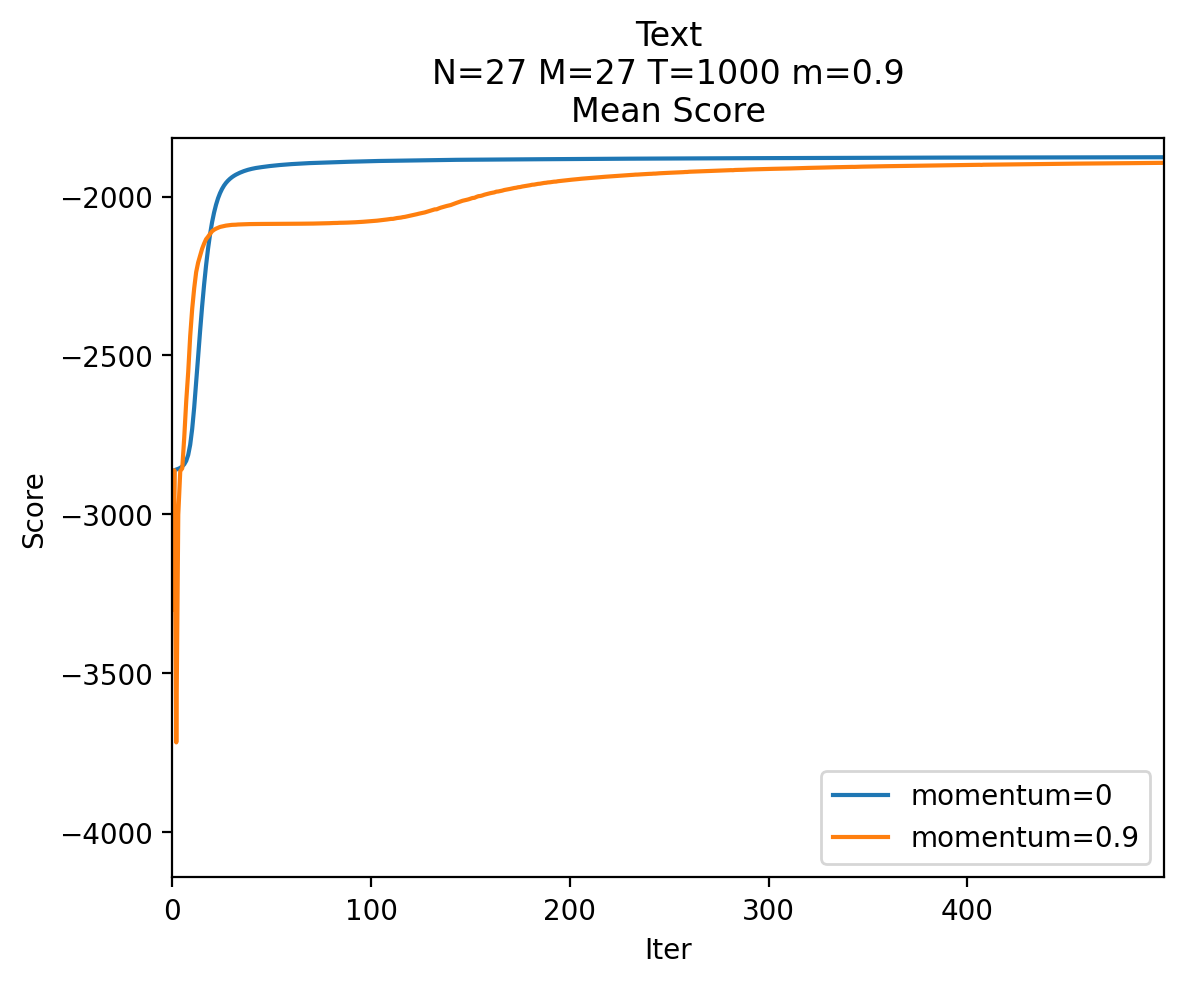}
\end{subfigure}%
\begin{subfigure}{.385\textwidth}
  \centering
  \includegraphics[width=1.0\linewidth]{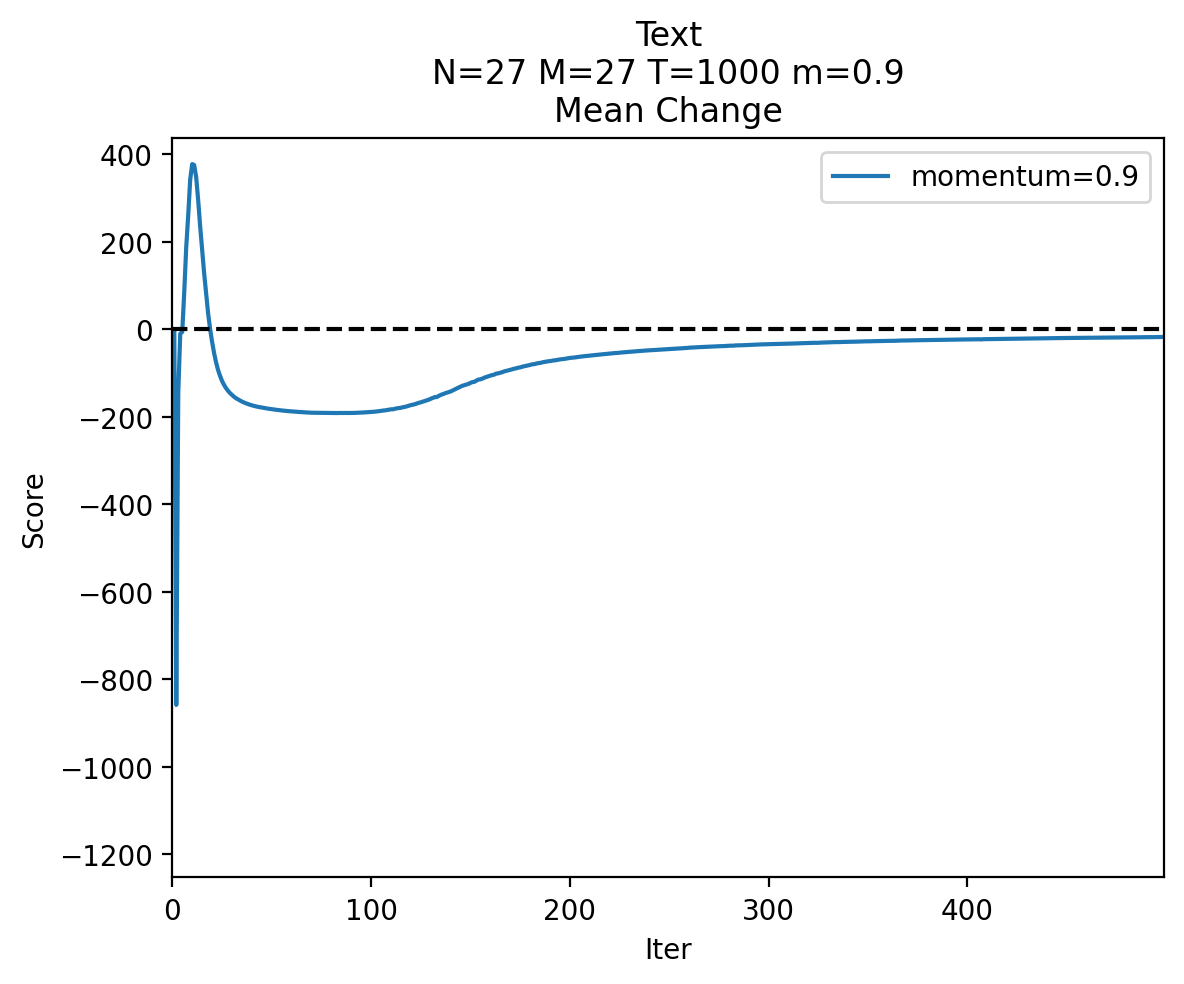}
\end{subfigure}
\begin{subfigure}{.385\textwidth}
  \centering
  \includegraphics[width=1.0\linewidth]{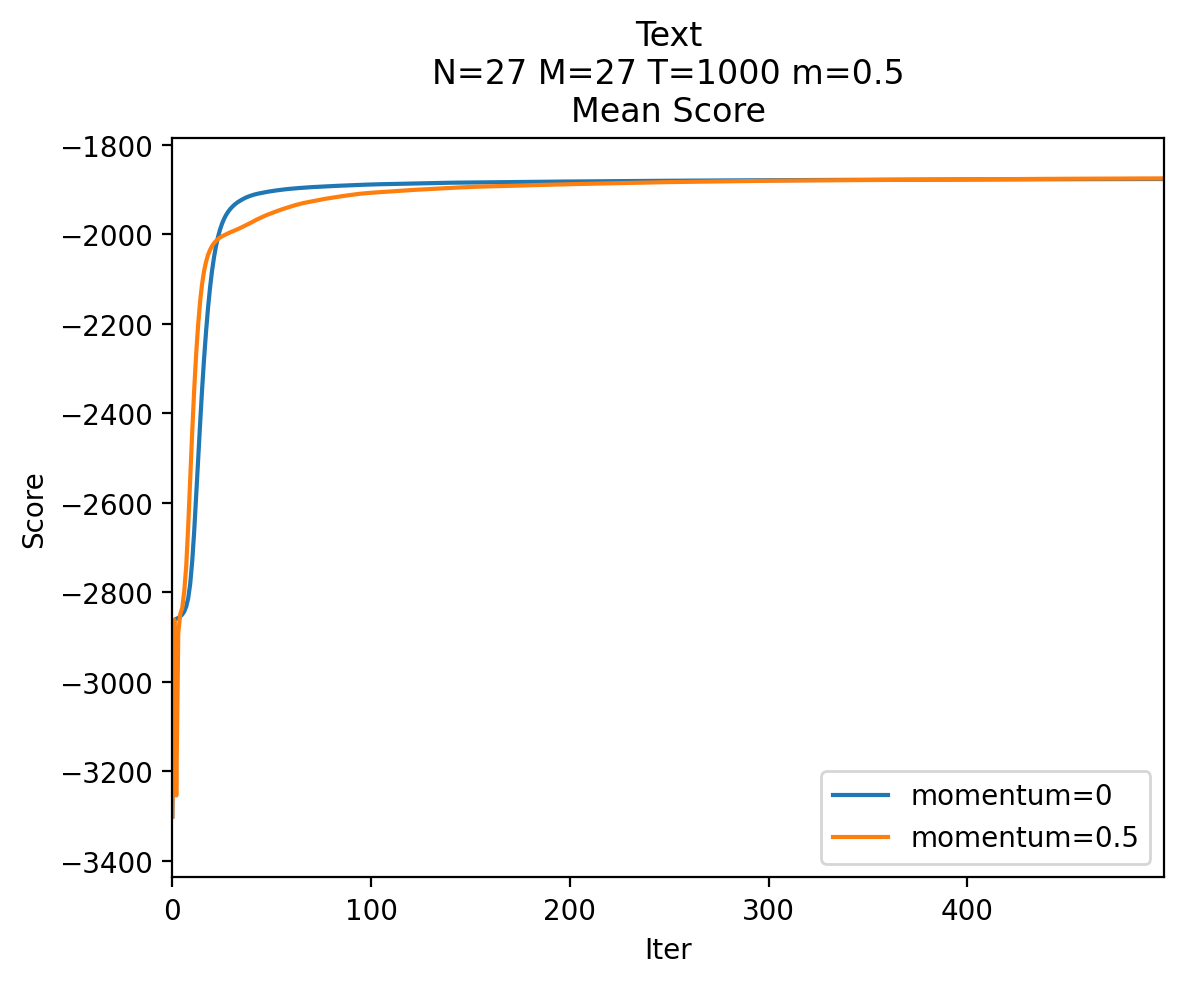}
\end{subfigure}%
\begin{subfigure}{.385\textwidth}
  \centering
  \includegraphics[width=1.0\linewidth]{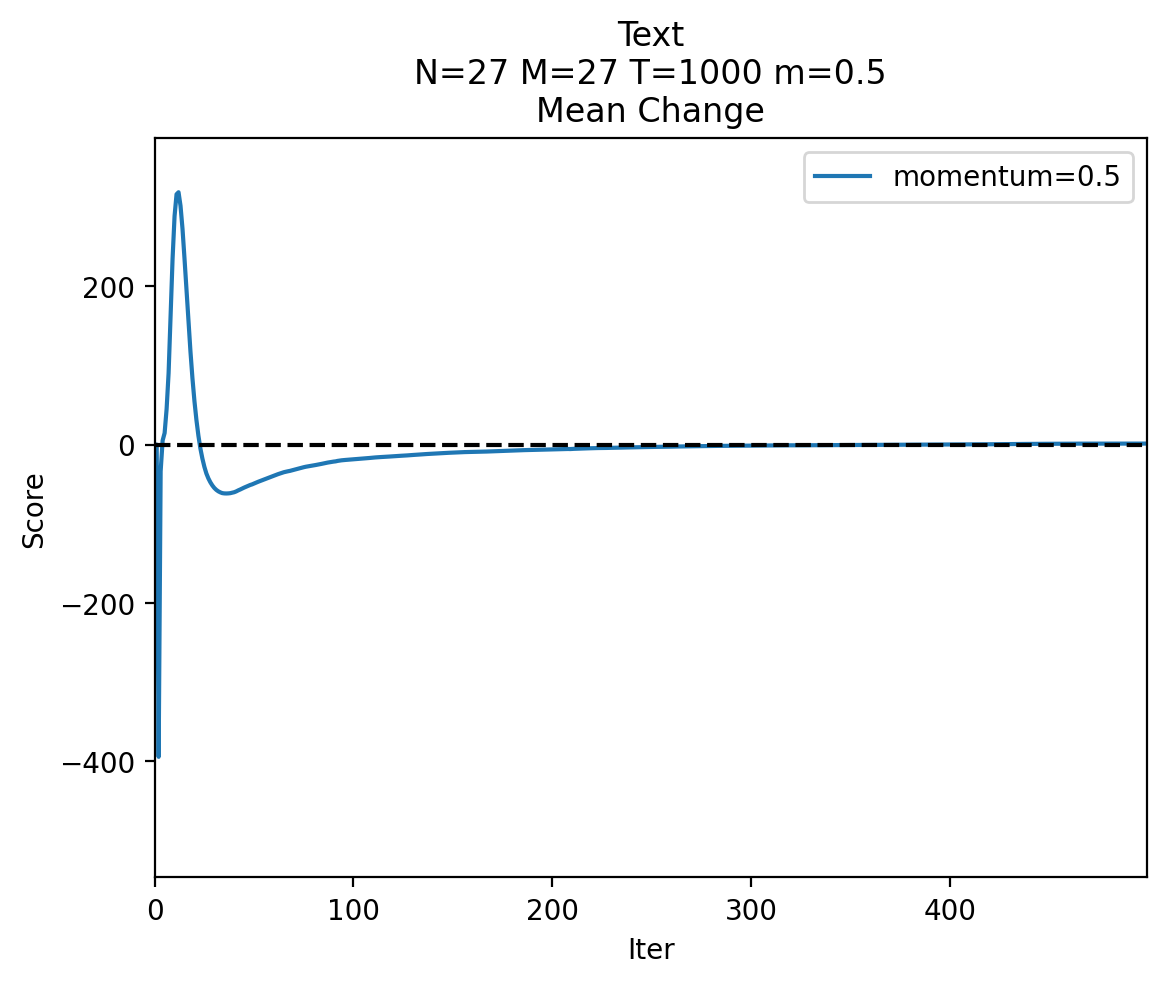}
\end{subfigure}
\begin{subfigure}{.385\textwidth}
  \centering
  \includegraphics[width=1.0\linewidth]{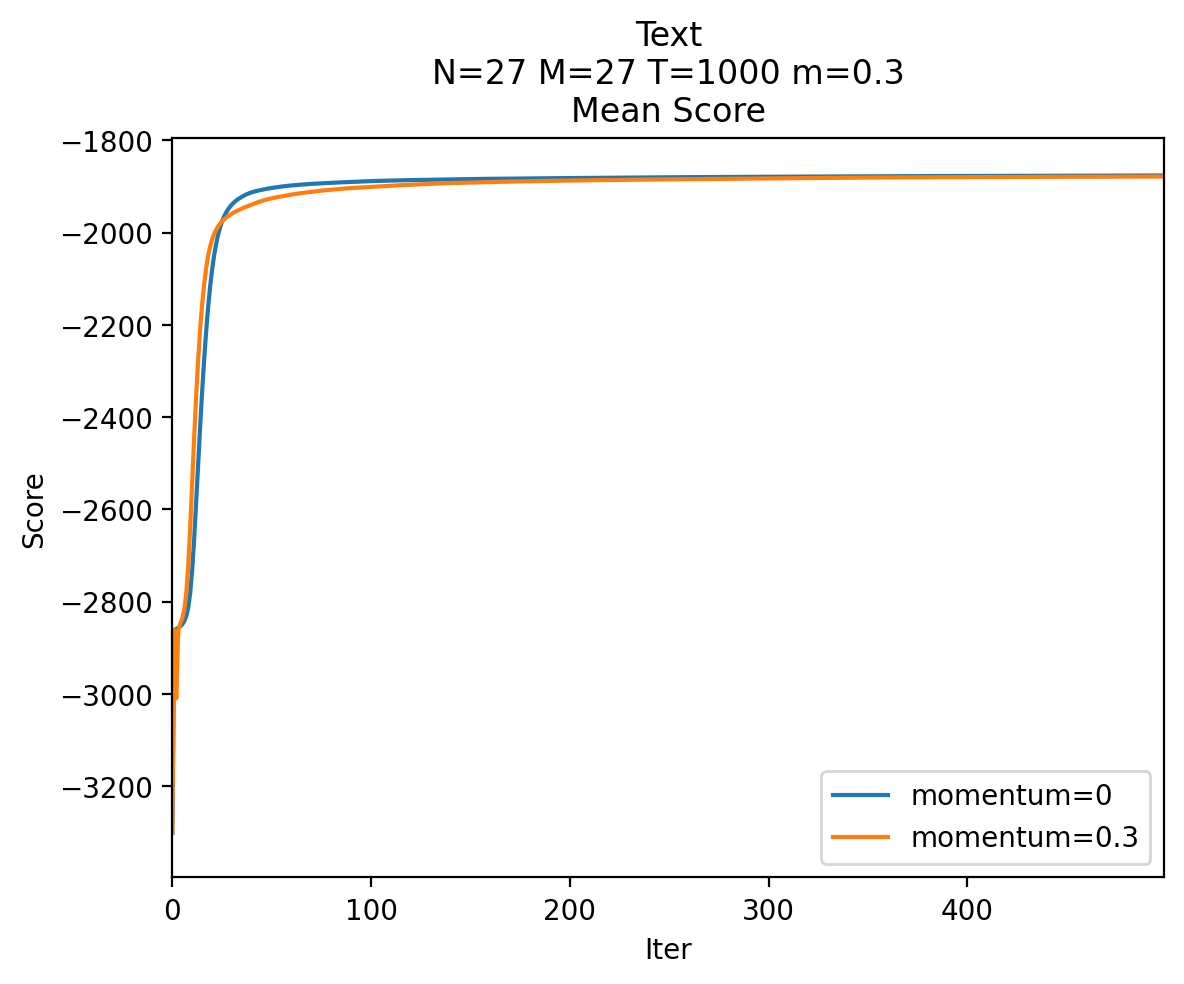}
\end{subfigure}%
\begin{subfigure}{.385\textwidth}
  \centering
  \includegraphics[width=1.0\linewidth]{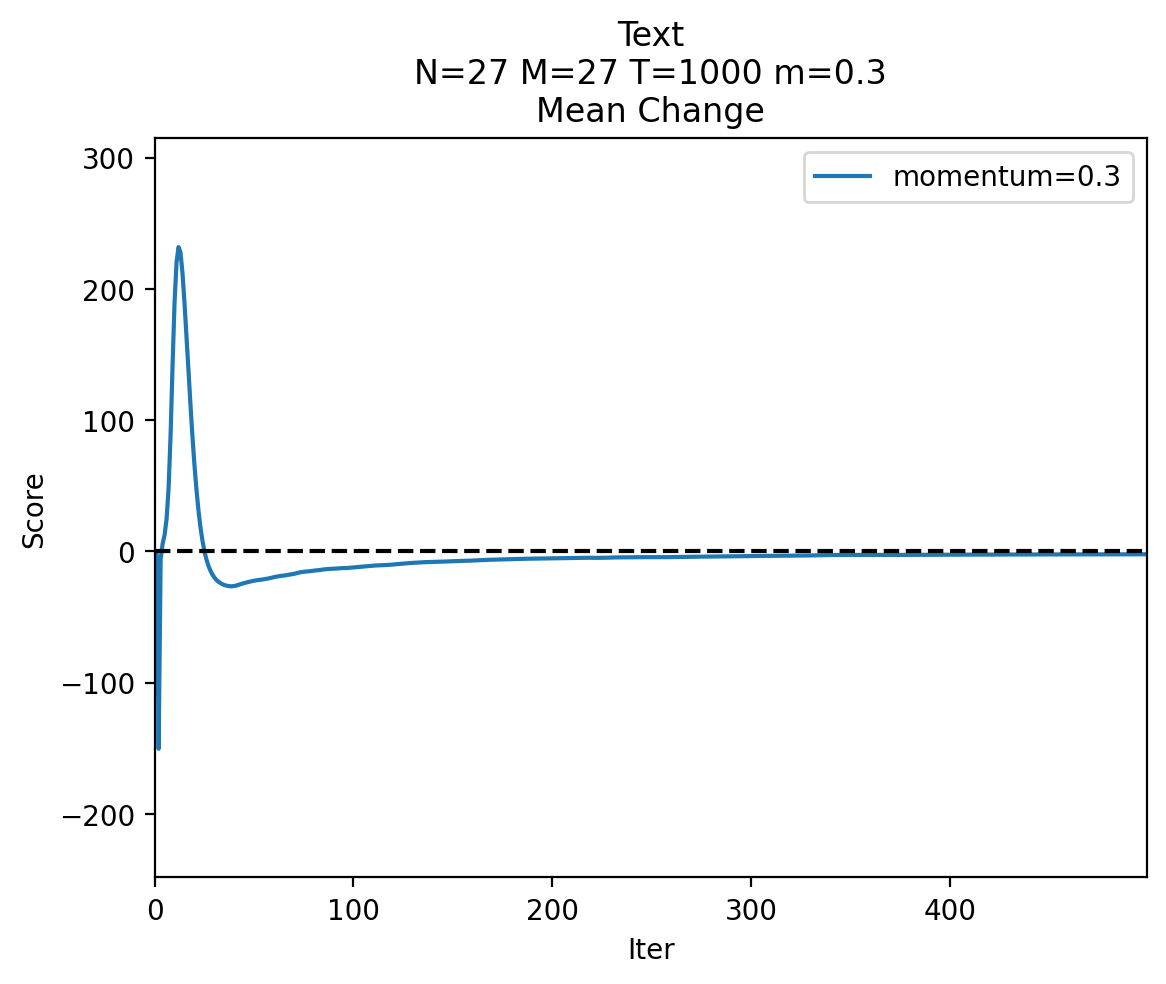}
\end{subfigure}
\caption{Change in score with momentum for $T=1000$}
\label{fig:text_momentum_T=1000}
\end{figure}

Reducing the observation sequence length from~$T=10{,}000$ to~$T=1000$ does not significantly 
change the behavior, as shown in Figure~\ref{fig:text_momentum_T=1000}.  The trend of an initial 
increase in score remains consistent across momentum values, but as the curve levels off, 
momentum at lower sequence lengths results in more negative changes in score.  This may indicate 
that with less data, the update direction at each iteration is less optimal, and therefore the momentum carried over is less beneficial.   

\begin{figure}[!htb]
\centering
\includegraphics[trim={0 0 0 0.2cm},clip,width=0.5\linewidth]{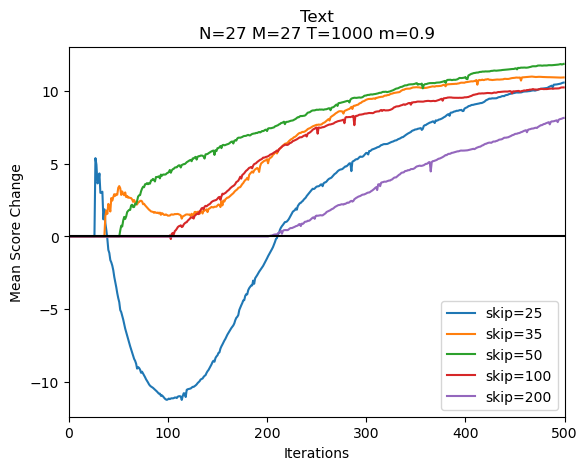}
\caption{Tail-only momentum score change}\label{fig:tailskip}
\end{figure}

As mentioned above, momentum built up during initial convergence can result in lingering negative effects on model score during the latter iterations of training.  In order to isolate and observe momentum for just the tail, 
we train models with momentum disabled for varying numbers of iterations at the start of training.  
These tests use a training sequence length of $T=1000$ and a momentum value of $m=0.9$.  Figure~\ref{fig:tailskip} shows the difference in scores over time when disabling momentum for the first~25, 35, 50, 100, and~200 iterations.  These results seem to indicate that momentum has a small positive effect on model scores at later iterations when not influenced by the higher magnitude past gradients.  
For this test case, it takes on average between~35 and~50 iterations before momentum stops overshooting.  

\subsubsection{Nesterov Momentum}

\begin{figure}[!htb]
\centering
\begin{subfigure}{.385\textwidth}
  \centering
  \includegraphics[width=1.0\linewidth]{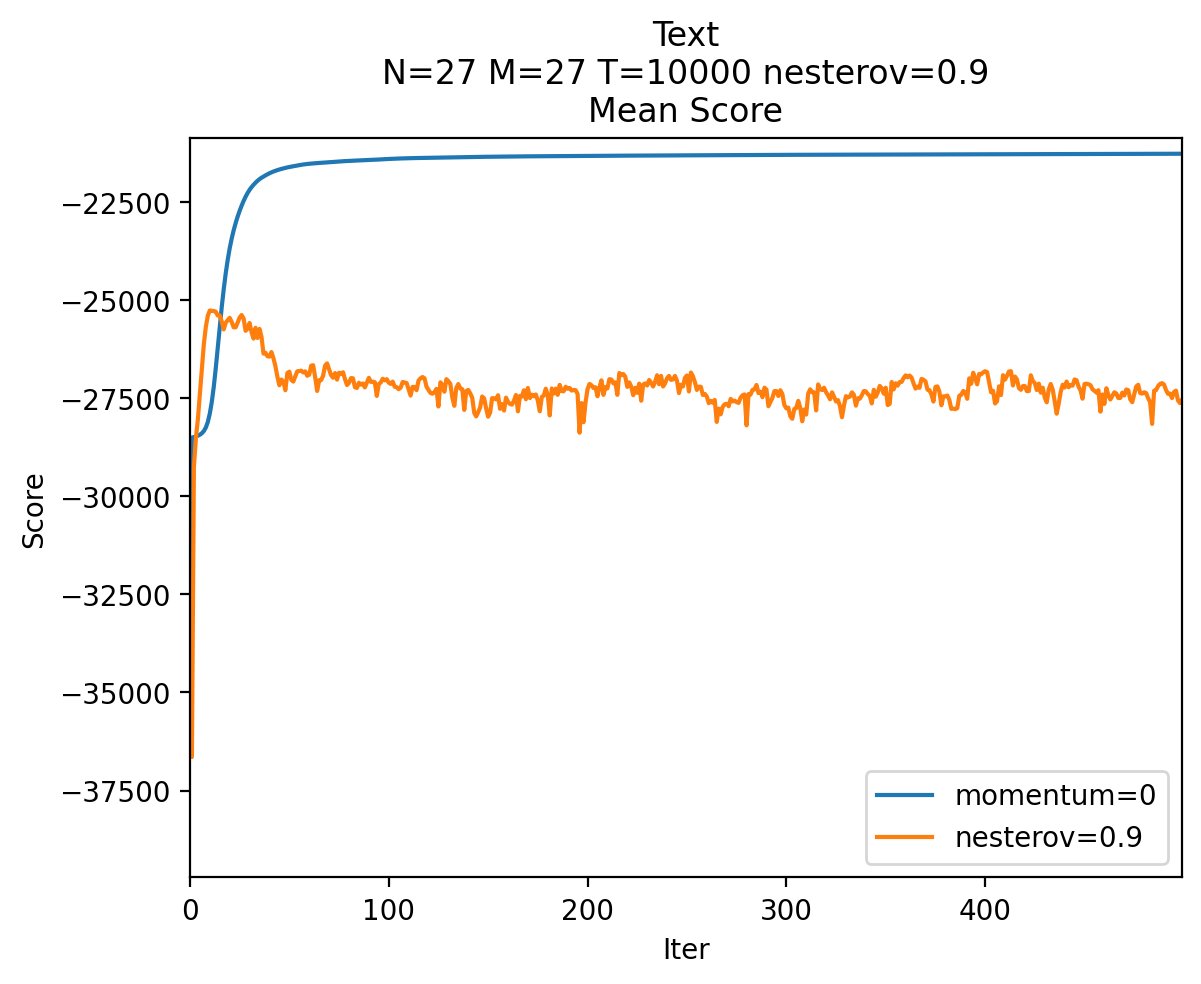}
  \label{fig:nest9mean}
\end{subfigure}%
\begin{subfigure}{.385\textwidth}
  \centering
  \includegraphics[width=1.0\linewidth]{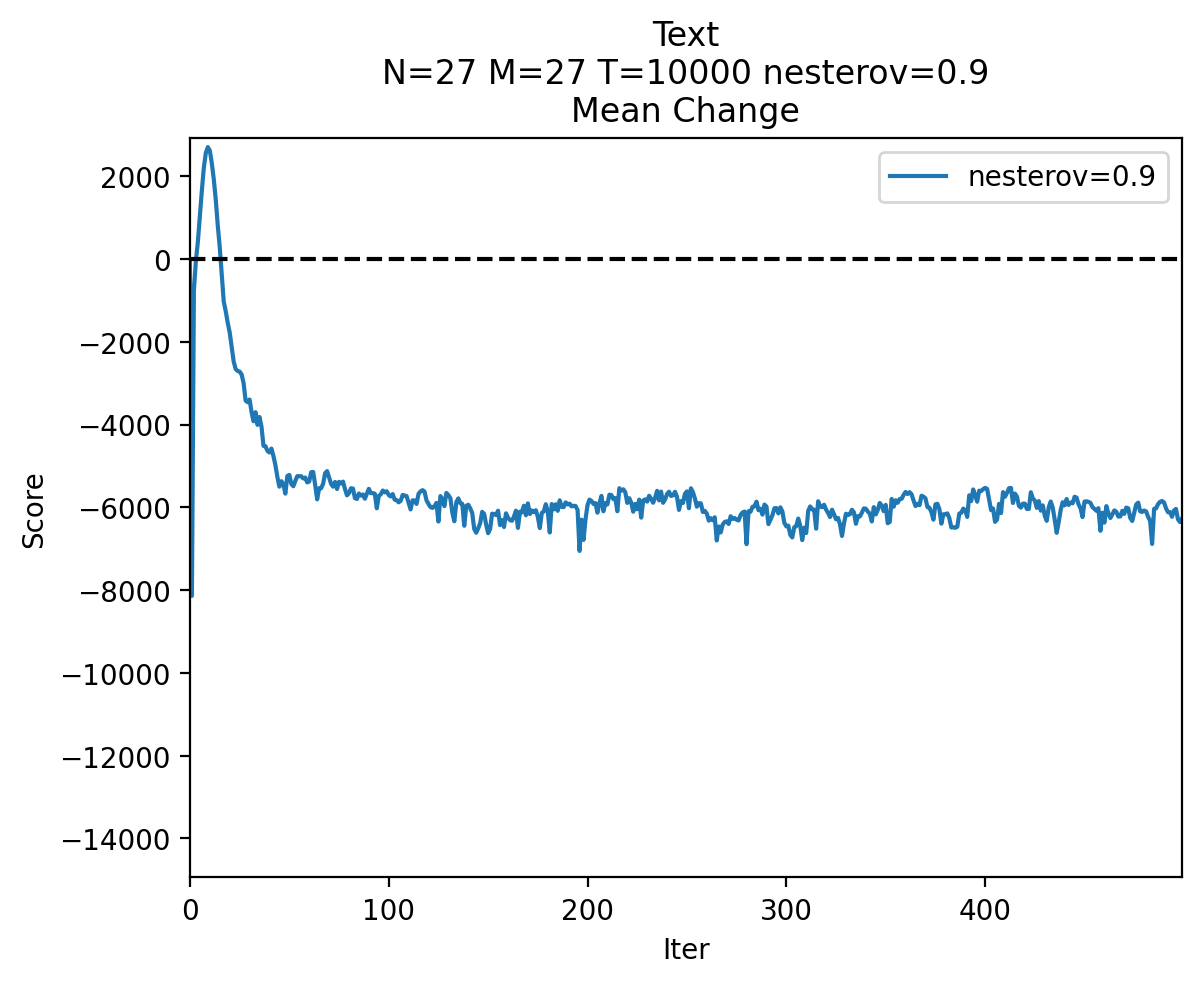}
  \label{fig:nest9diff}
\end{subfigure}
\begin{subfigure}{.385\textwidth}
  \centering
  \includegraphics[width=1.0\linewidth]{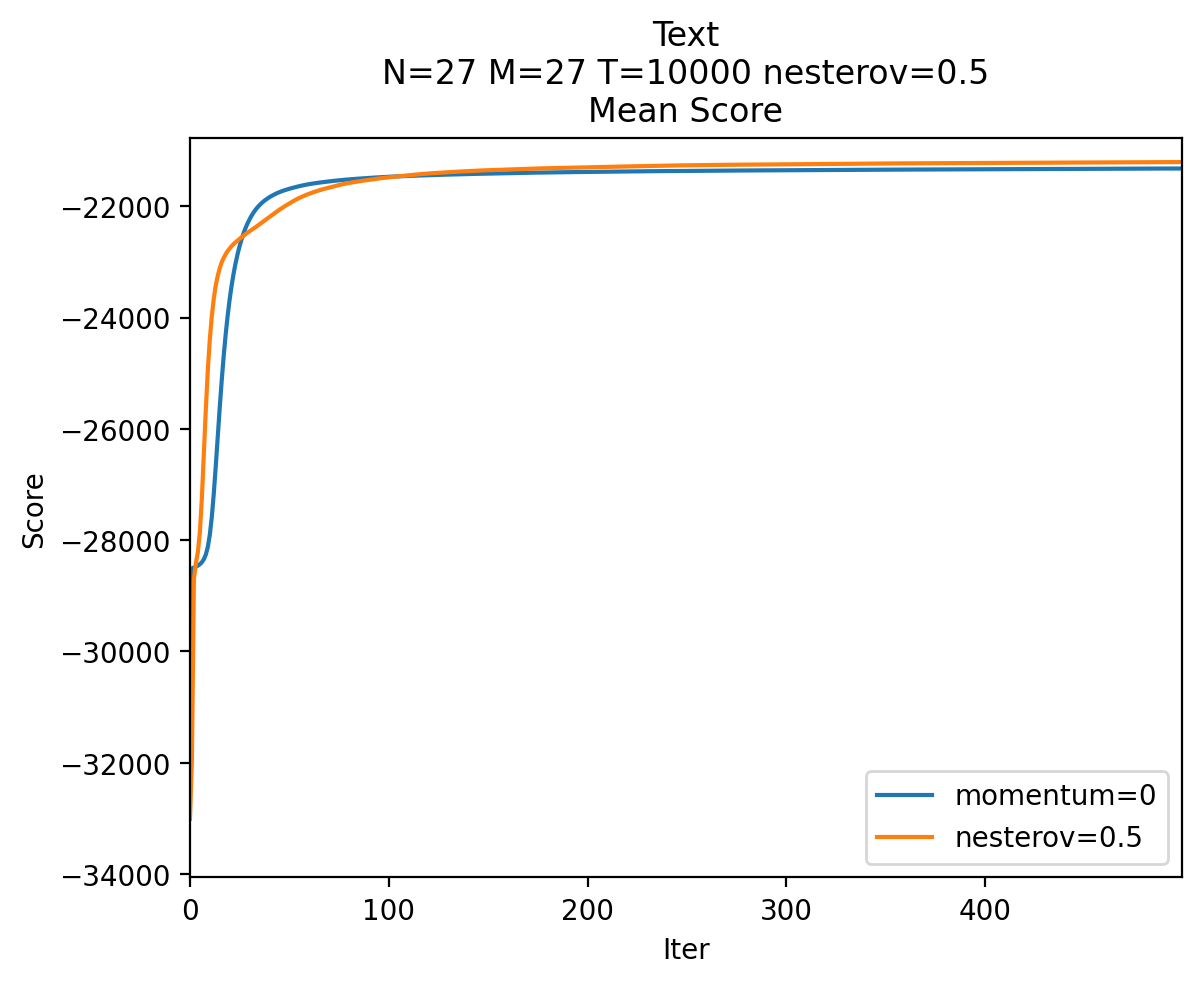}
  \label{fig:nest5mean}
\end{subfigure}%
\begin{subfigure}{.385\textwidth}
  \centering
  \includegraphics[width=1.0\linewidth]{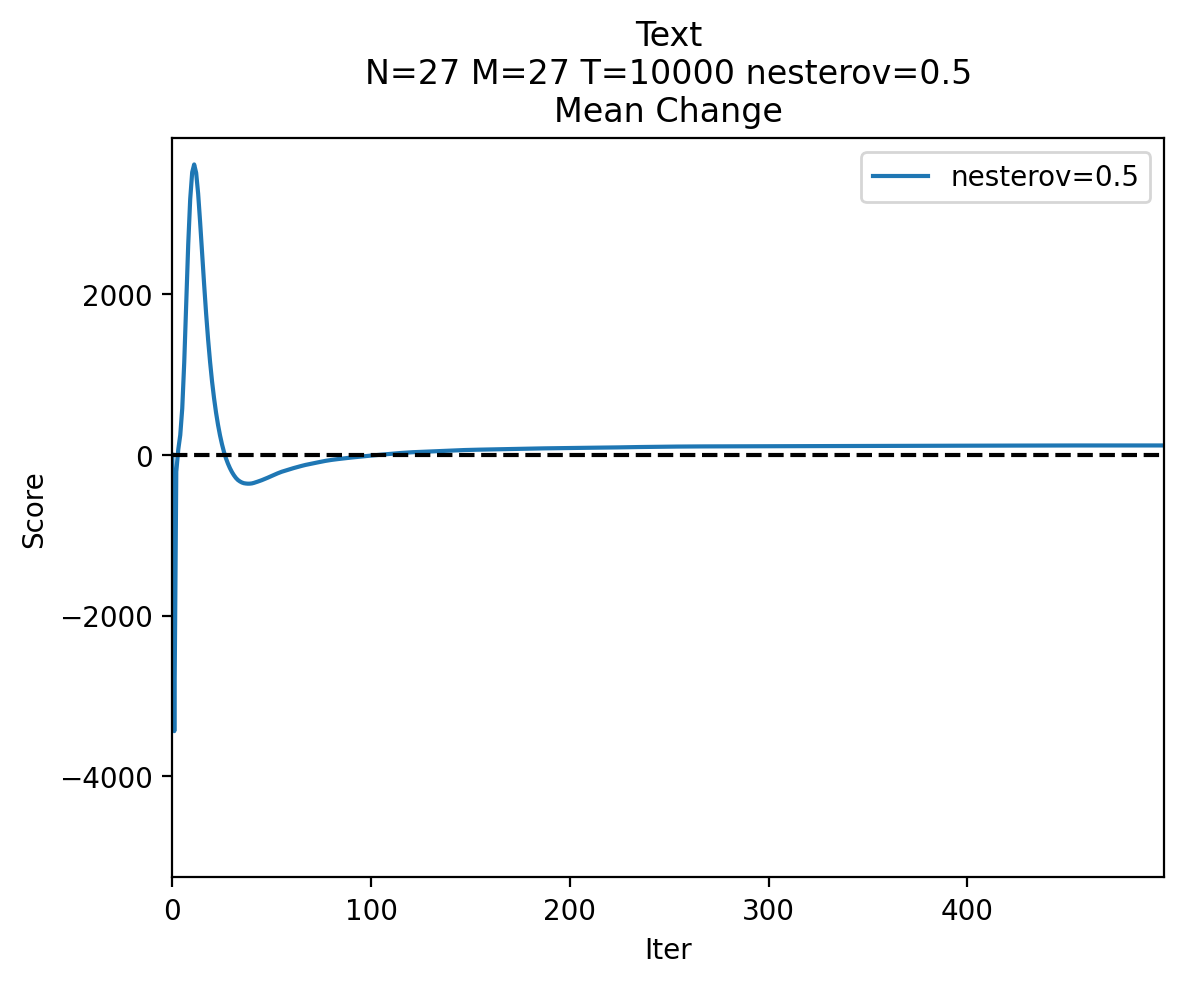}
  \label{fig:nest5diff}
\end{subfigure}
\begin{subfigure}{.385\textwidth}
  \centering
  \includegraphics[width=1.0\linewidth]{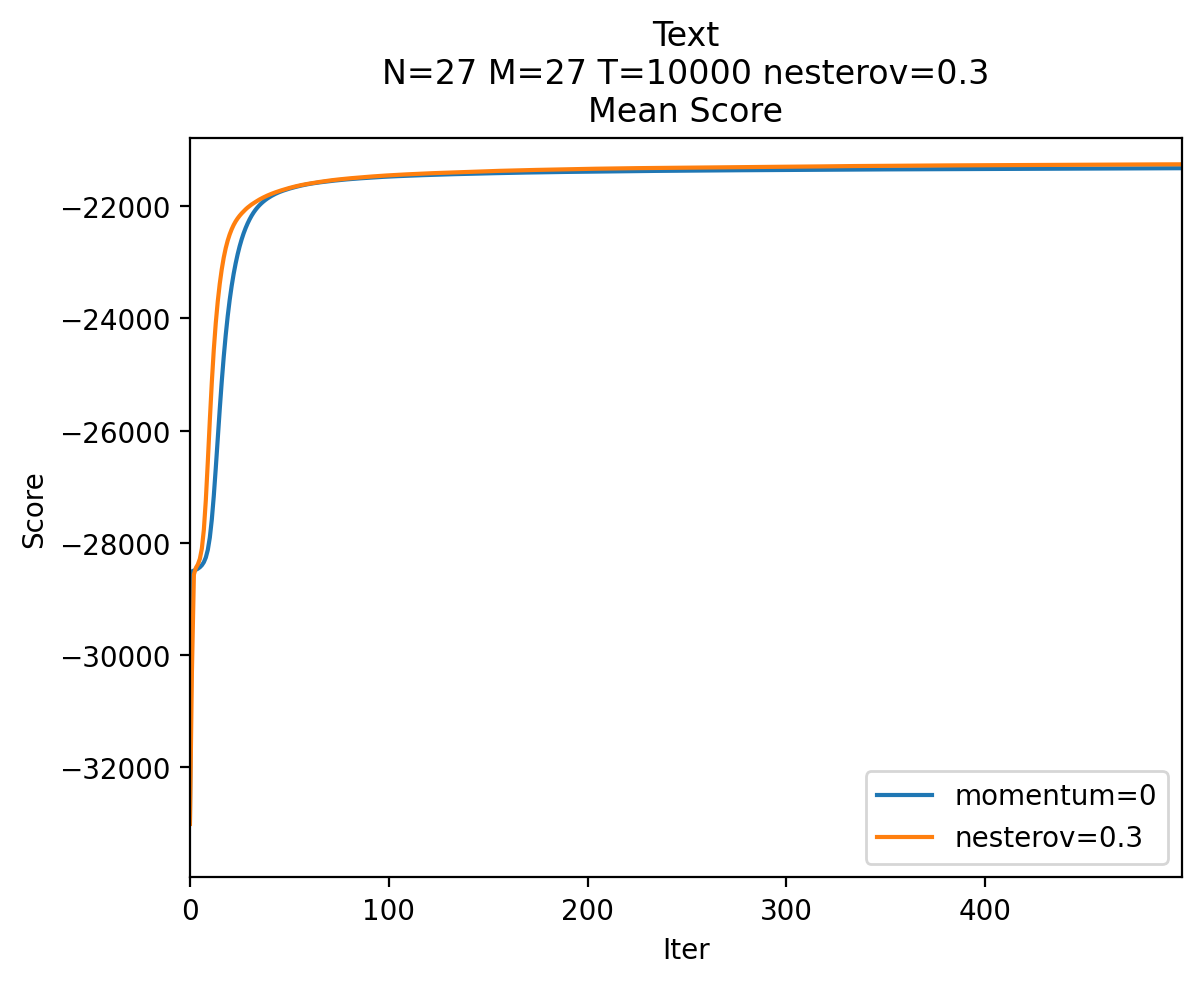}
  \label{fig:nest3mean}
\end{subfigure}%
\begin{subfigure}{.385\textwidth}
  \centering
  \includegraphics[width=1.0\linewidth]{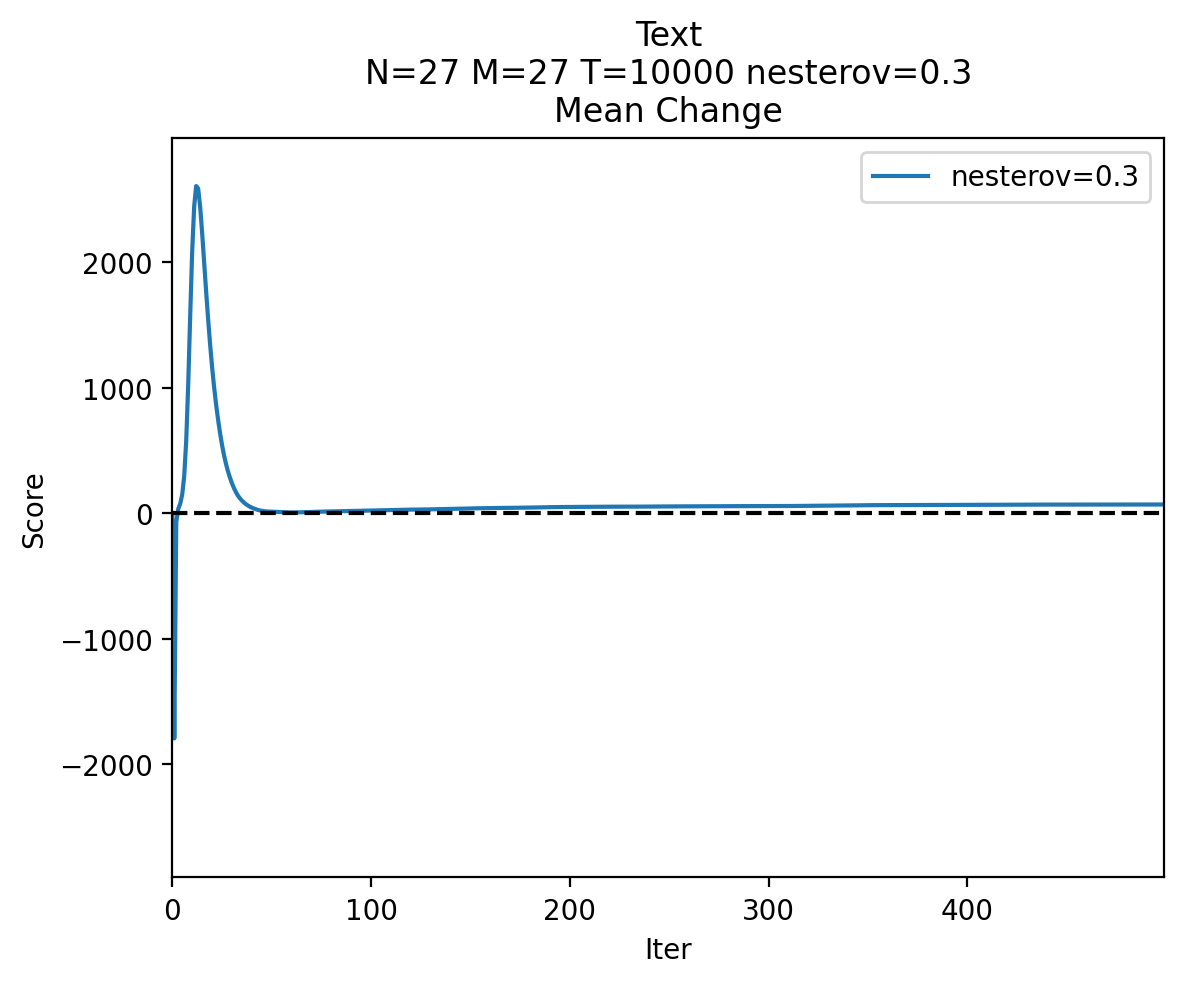}
  \label{fig:nest3diff}
\end{subfigure}
\caption{Nesterov results for $T=10{,}000$}
\label{fig:text_nesterov_T=10000}
\end{figure}

Reproducing the same experiments but with Nesterov momentum yields similar trends in training behavior.  However, at high momentum values, Nesterov momentum results in the model failing to converge.  Figure~\ref{fig:text_nesterov_T=10000} depicts training scores of models trained identically to those in Figure~\ref{fig:text_momentum_T=10000}, but instead using Nesterov momentum.  Nesterov momentum 
demonstrates the same overshoot behavior, with drops in score during the first few iterations and again after initial convergence.  While unusable at higher momentum, at lower values Nesterov momentum seem to generally produce larger peak score increases during early iterations.  Nesterov momentum also results in a greater mean final increase in score after a high number of iterations with equal momentum values.  

Table~\ref{tab:momentum_nesterov_comparison} compares the difference in scores at a select set of iterations.  
Nesterov momentum seems to generally outperform the standard momentum implementation, but risks being unable to converge if momentum is set too high.

\begin{table}[!htb]
    \caption{Change in score} 
    \label{tab:momentum_nesterov_comparison}
    \centering
    \adjustbox{scale=0.85}{
    \begin{tabular}{c|rrrrrrr}
        \midrule\midrule
        \multirow{2}{*}{Momentum} & \multicolumn{7}{c}{Iteration} \\ \cmidrule{2-8}
         & 5 & 15 & 25 & 50 & 100 & 200 & 500\\
        \midrule
        $\mom=0.9$ & -108 & 2108 & -318 & -1250 & -1309 & -282 & 56\\
        $\mom=0.5$ & 21 & 2313 & 465 & -120 & 4 & 55 & 88\\
        $\mom=0.3$ & 24 & 1618 & 486 & 5 & 8 & 29 & 40\\
        \midrule
        $\nest=0.5$ & 246 & 2890 & 241 & -275 & -12 & 84 & 116\\
        $\nest=0.3$ & 77 & 2443 & 640 & 13 & 21 & 50 & 69\\
        \midrule\midrule
    \end{tabular}
    }
\end{table}

\subsubsection{Number of Hidden States}\label{highN}

\begin{figure}[!htb]
\centering
\begin{subfigure}{.385\textwidth}
  \centering
  \includegraphics[width=1.0\linewidth]{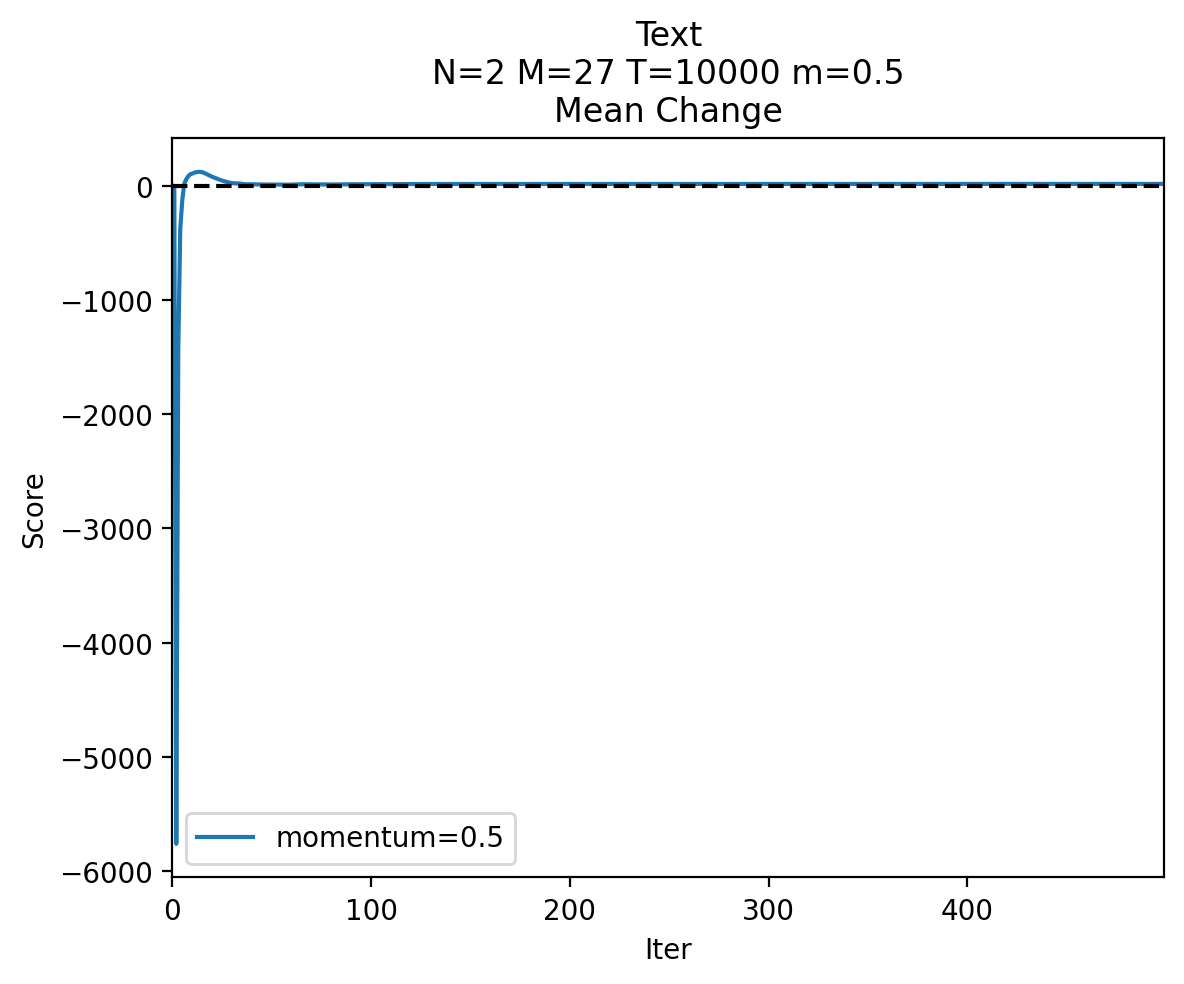}
\end{subfigure}
\begin{subfigure}{.385\textwidth}
  \centering
  \includegraphics[width=1.0\linewidth]{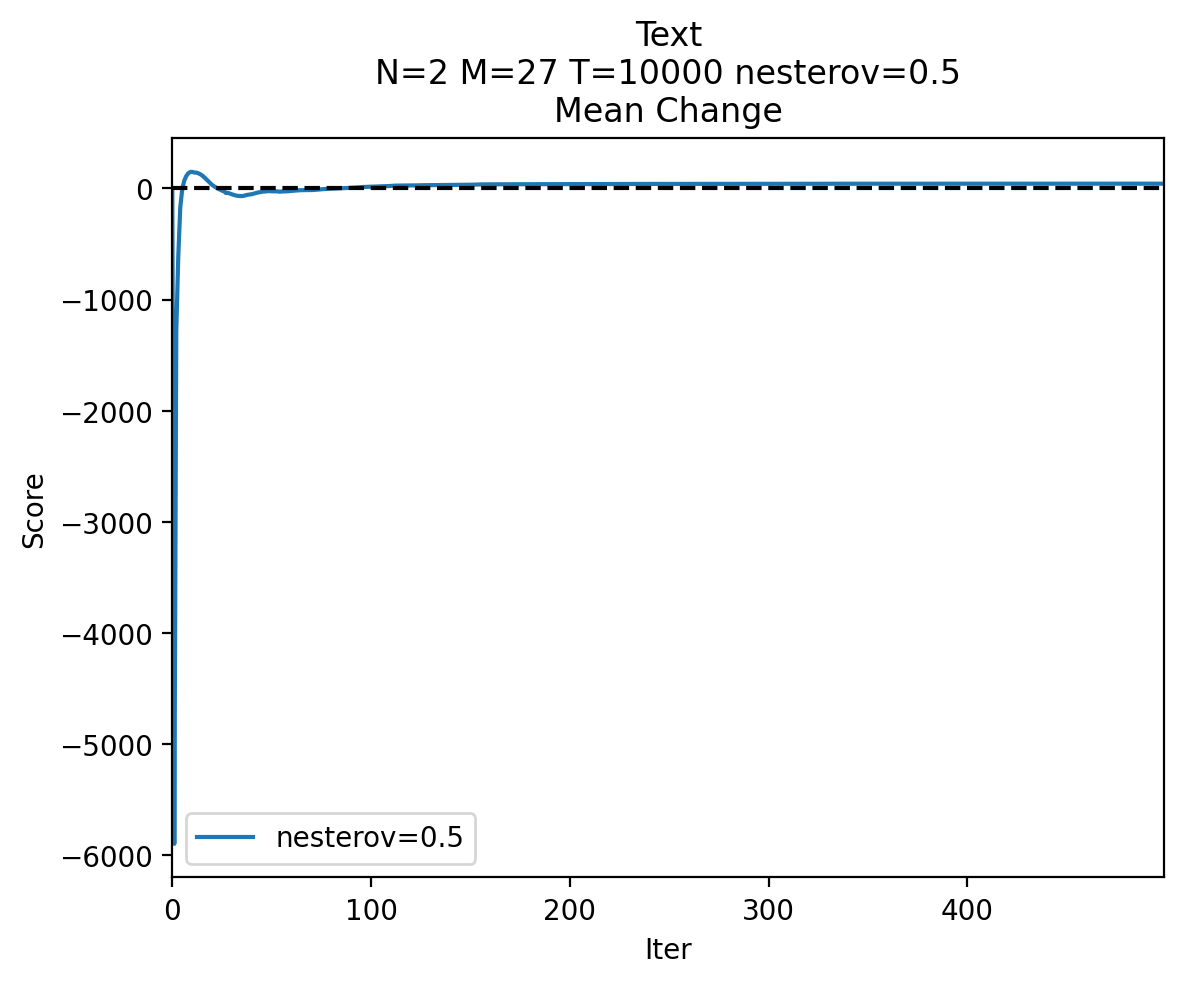}
\end{subfigure}
\caption{Momentum and Nesterov with $N=2$}\label{fig:lowN_N=2}
\end{figure}

To determine how the number of hidden states interacts with momentum, we compare the 
effects of momentum with~$N=2$, $N=10$, and~$N=27$.  
Tests with~$N=2$ show almost no difference with or without momentum, 
as demonstrated in Figure~\ref{fig:lowN_N=2}.  The previously observed behavior of a sharp initial dip followed by a positive spike continues, but excluding the initial negative spike, changes in score are negligible for both standard and Nesterov momentum.  

\begin{figure}[!htb]
\centering
\begin{subfigure}{.385\textwidth}
  \centering
  \includegraphics[width=1.0\linewidth]{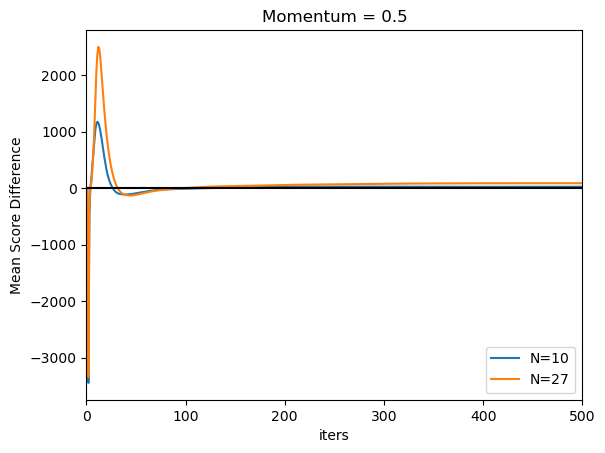}
\end{subfigure}
\begin{subfigure}{.385\textwidth}
  \centering
  \includegraphics[width=1.0\linewidth]{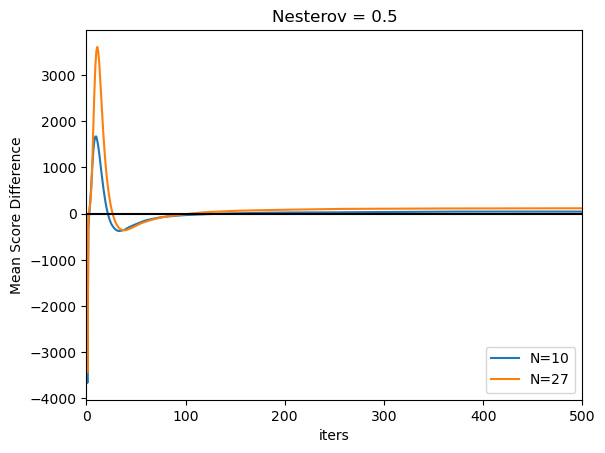}
\end{subfigure}
\caption{Momentum and Nesterov for $N=10$ and $N=27$}\label{fig:lowN_N=10}
\end{figure}

Increasing the number of hidden states to~$N=10$ and~$N=27$, as in Figure~\ref{fig:lowN_N=10}, 
produces much more significant changes in score compared to the baseline HMM.  These experiments
indicate that momentum may be more impactful at higher values of~$N$.  A possible explanation is that models with low numbers of hidden states are less complex and more easily optimized, leaving little opportunity for momentum to affect the score.

\subsubsection{Plateaus}\label{plateaus}

\begin{figure}[!htb]
\centering
\begin{subfigure}{.385\textwidth}
  \centering
  \includegraphics[width=1.0\linewidth]{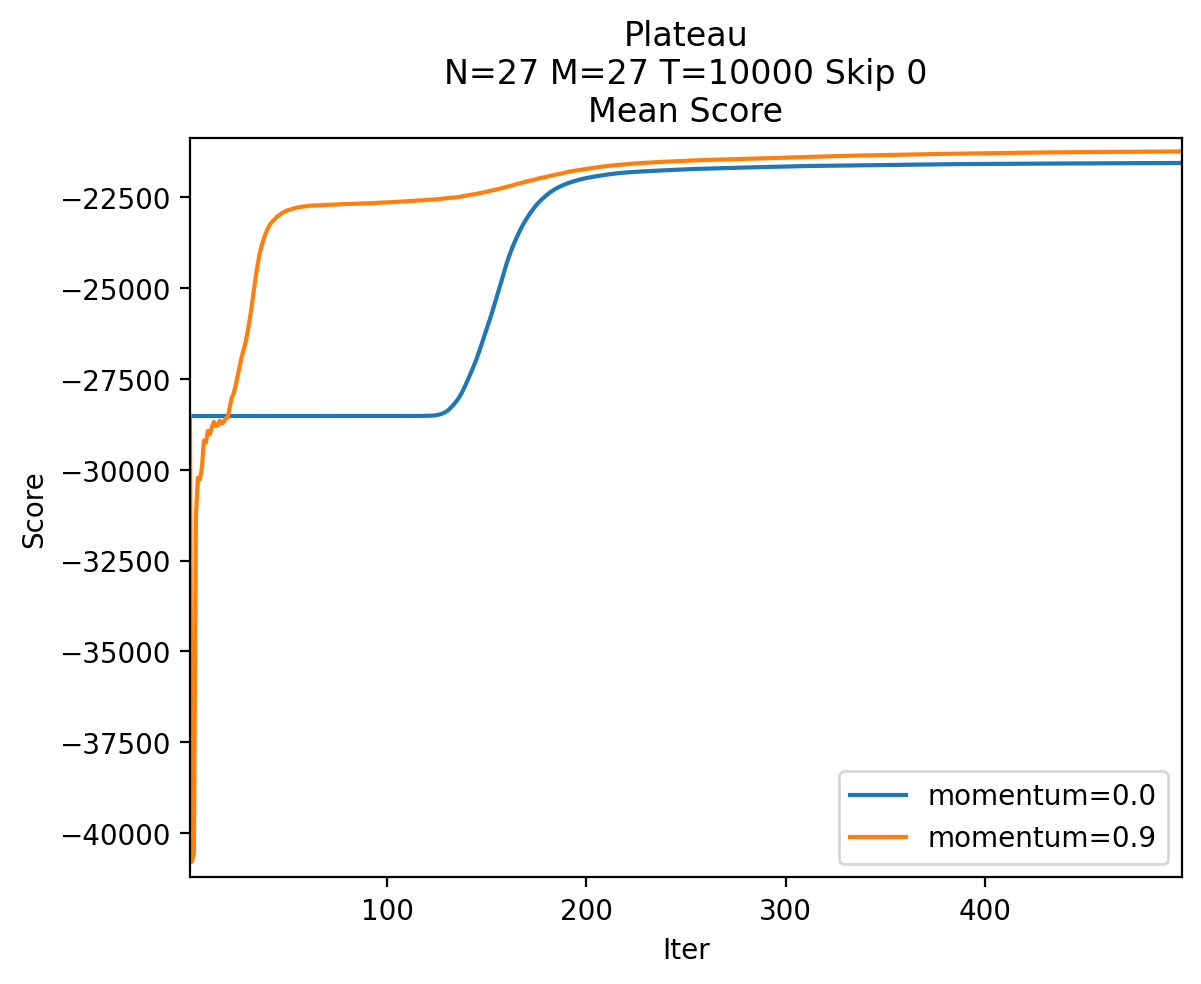}
\end{subfigure}
\begin{subfigure}{.385\textwidth}
  \centering
  \includegraphics[width=1.0\linewidth]{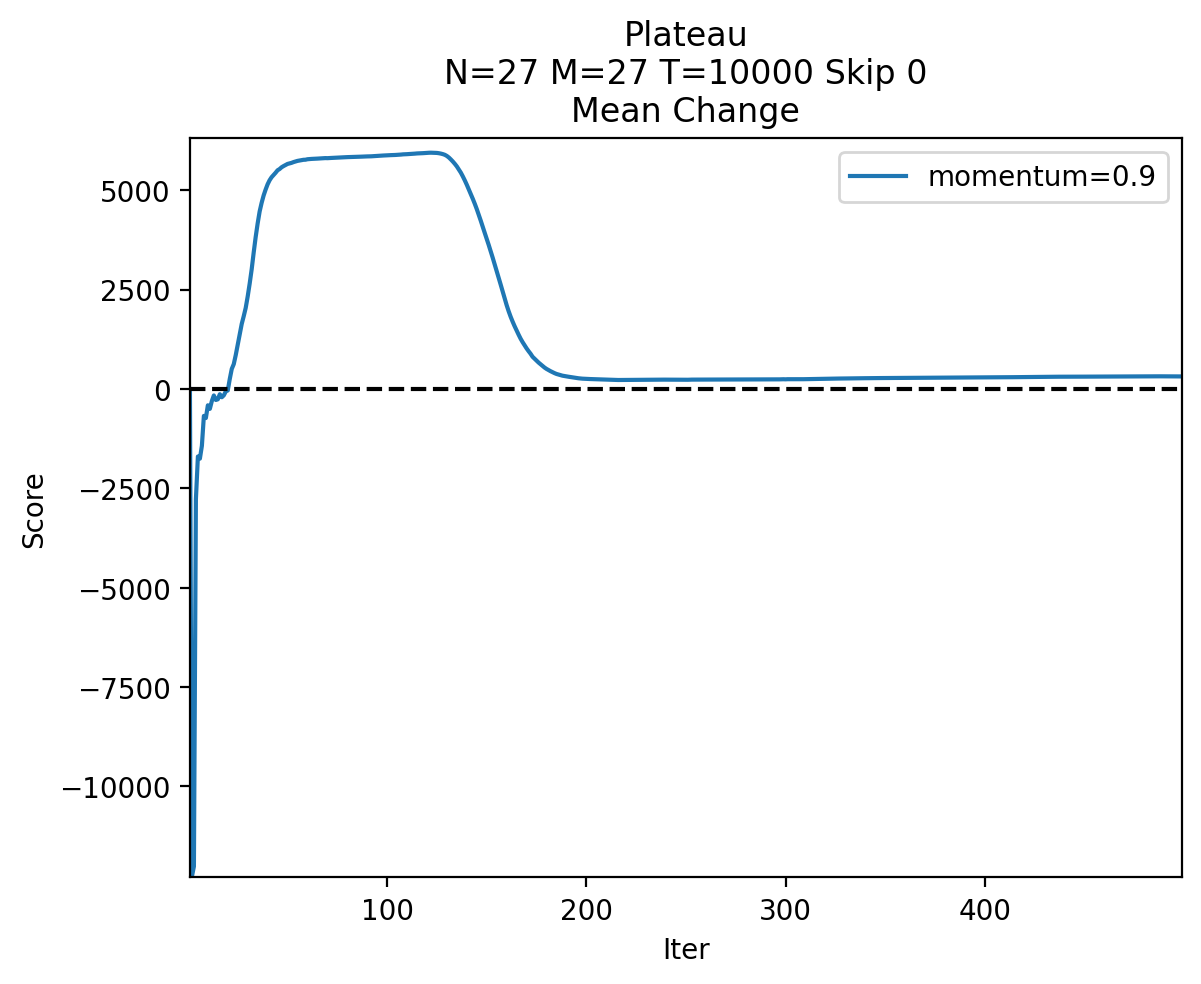}
\end{subfigure}
\caption{Momentum vs plateau at~$N=27$}\label{fig:plateau1}
\end{figure}

\begin{figure}[!htb]
\centering
\begin{subfigure}{.385\textwidth}
  \centering
  \includegraphics[width=1.0\linewidth]{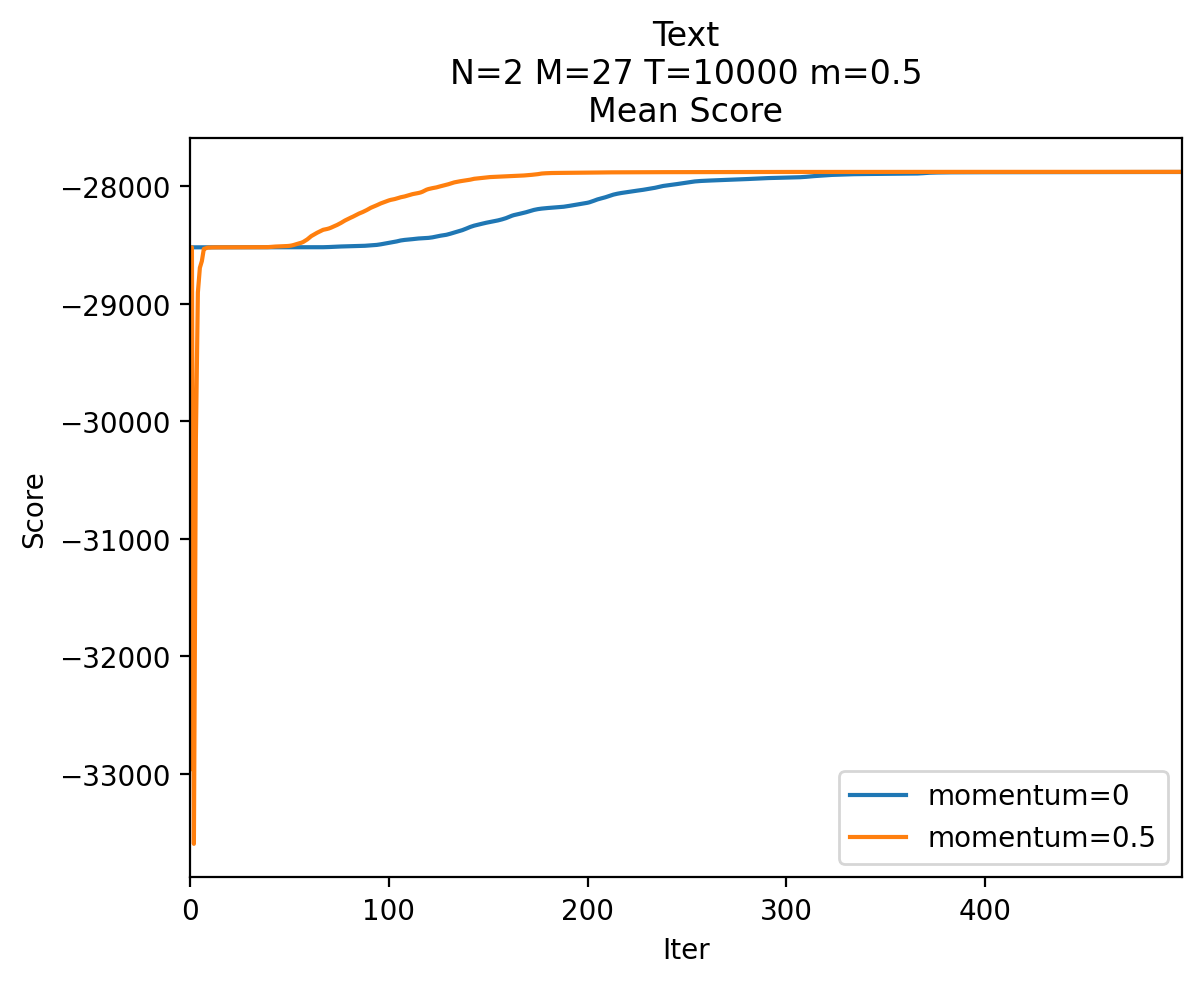}
\end{subfigure}
\begin{subfigure}{.385\textwidth}
  \centering
  \includegraphics[width=1.0\linewidth]{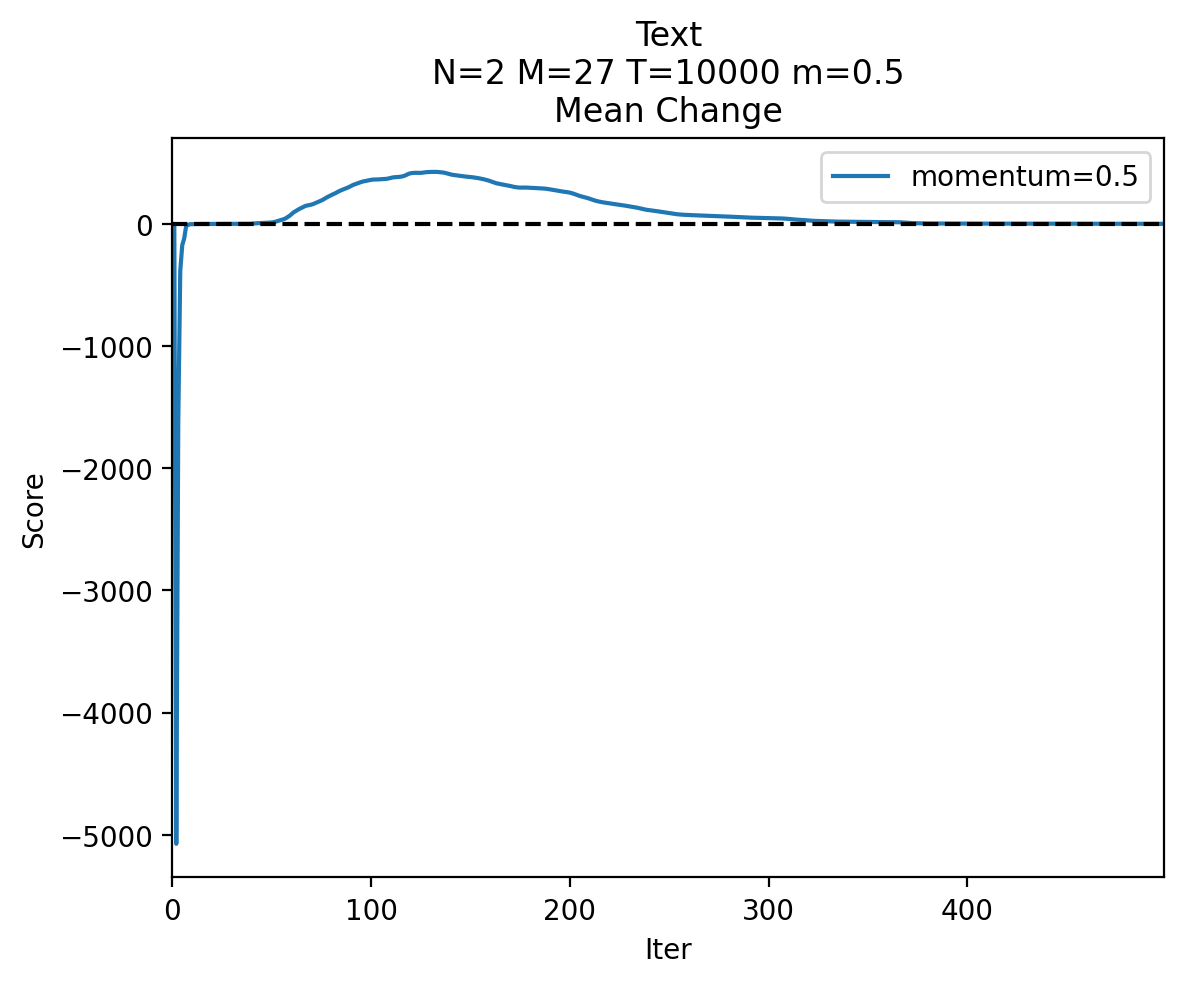}
\end{subfigure}
\caption{Momentum vs plateau, $N=2$}\label{fig:plateau2}
\end{figure}

While the result above shows that momentum can slightly improve the speed of initial convergence, the reduction in the amount of iterations before training levels off remains fairly small.  However, our experiments 
demonstrate significant speedup for cases in which training scores plateau early before eventually converging.  
Such a plateau can be produced by initializing English text models with values such 
that~$\pi_i \approx 1/N$, $a_{ij} \approx 1/N$, and~$b_i(j) \approx 1/M$, 
as recommended in~\cite{stampHMM}.  We train models using this initialization scheme 
for~500 iterations, 100 random restarts, $N=27$, $M=27$, and~$T=10{,}000$.  Figure~\ref{fig:plateau1} 
shows that momentum is able to reduce the number of necessary iterations for this case from~200 iterations
to about~50.  Momentum successfully breaks out of early plateaus even at the 
lowest value of~$N=2$, as shown in Figure~\ref{fig:plateau2}.  

\subsubsection{Momentum Scheduling}

In gradient descent, learning rate schedules are used to dynamically control the learning rate based on some predetermined function.  
Because the gradient descent momentum update is dependent on the learning rate, learning rate scheduling also indirectly influences the amount of momentum at each step.
While our previous Baum-Welch momentum experiments have used a static momentum value 
for simplicity and consistency, it is unlikely that a single momentum value will produce optimal results at all points in training.  Here, we test momentum scheduling as a potential solution.  

\begin{figure}[!htb]
\centering
\begin{subfigure}{.385\textwidth}
  \centering
  \includegraphics[width=1.0\linewidth]{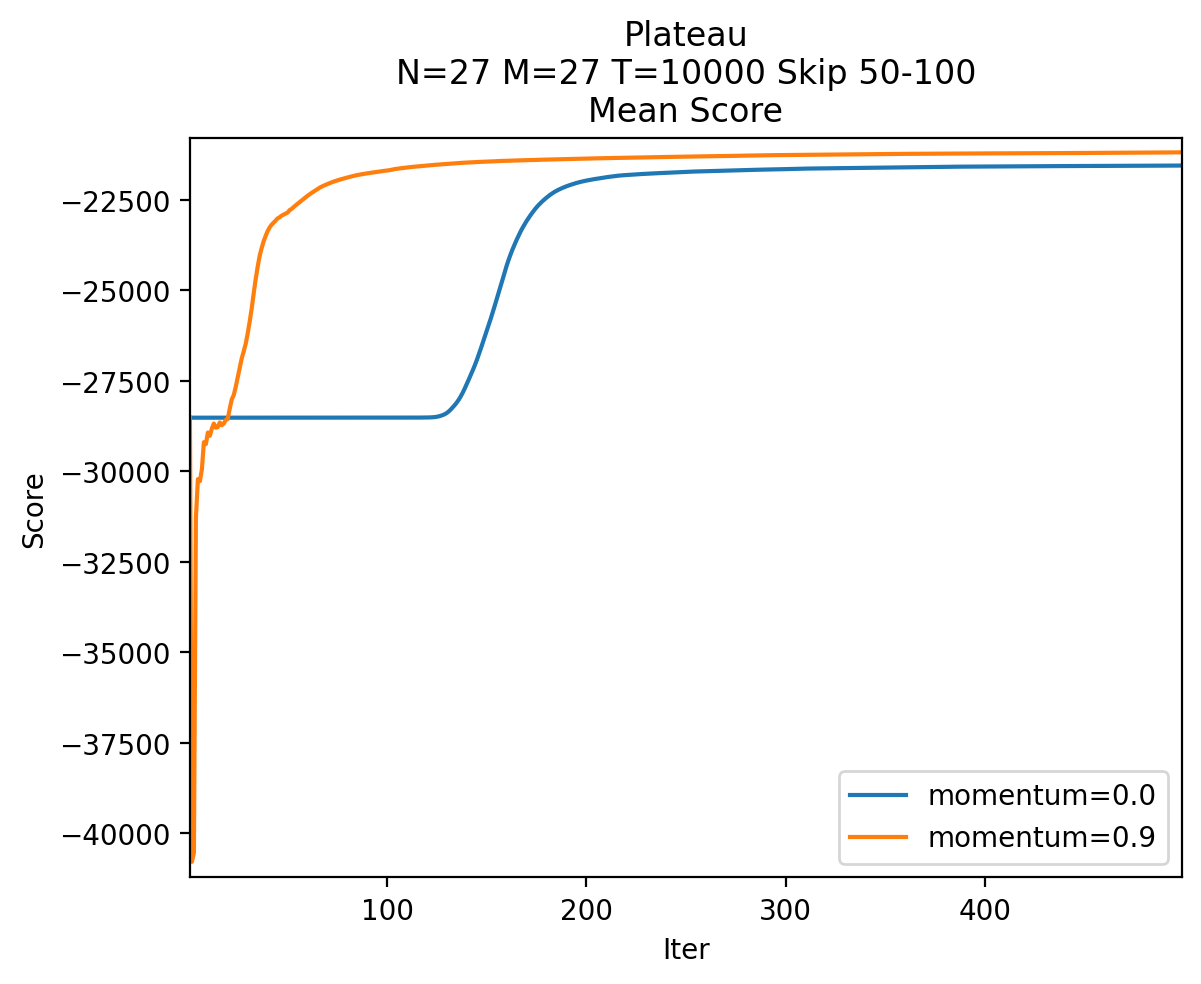}
\end{subfigure}
\begin{subfigure}{.385\textwidth}
  \centering
  \includegraphics[width=1.0\linewidth]{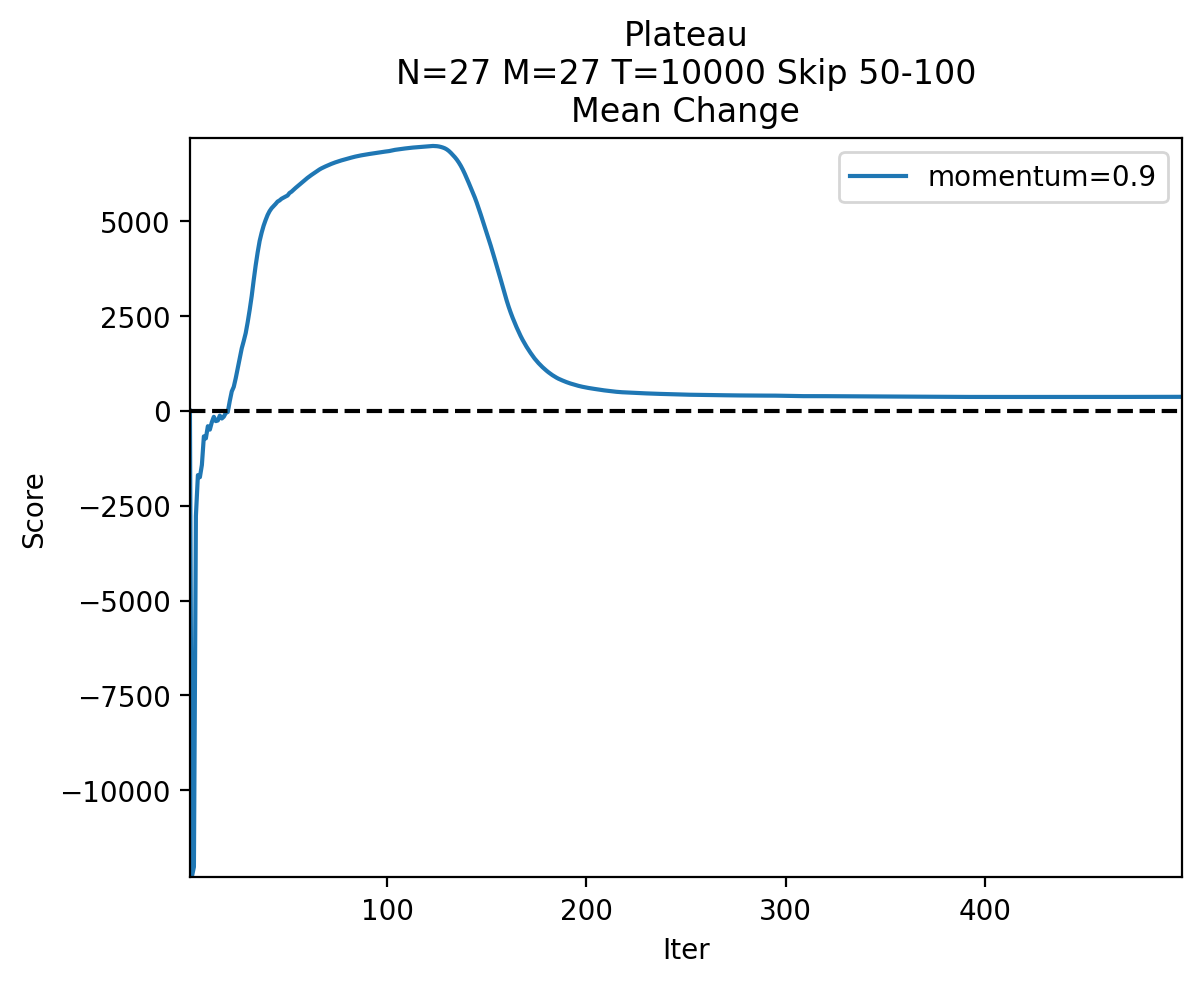}
\end{subfigure}
\caption{Plateau example, skipping iterations 50 to 100}\label{fig:sched_plateau1}
\end{figure}

As demonstrated in Figures~\ref{fig:text_momentum_T=10000}~and~\ref{fig:text_nesterov_T=10000}, momentum generally produces positive changes early and late in training, but tends to overshoot as training slows, particularly at higher momentum values.  As a na\"i{v}e solution, a momentum schedule is implemented in which the momentum is set to~0 for a predetermined range of iterations.  
Figure~\ref{fig:sched_plateau1} displays an example of the difference in training behavior with and without this na\"{i}ve schedule using the previous plateau example.  Disabling momentum between iterations~50 and~100 for this model produces a smoother curve, and effectively eliminates the overshoot.  

\begin{figure}[!htb]
\centering
\begin{subfigure}{.385\textwidth}
  \centering
  \includegraphics[width=1.0\linewidth]{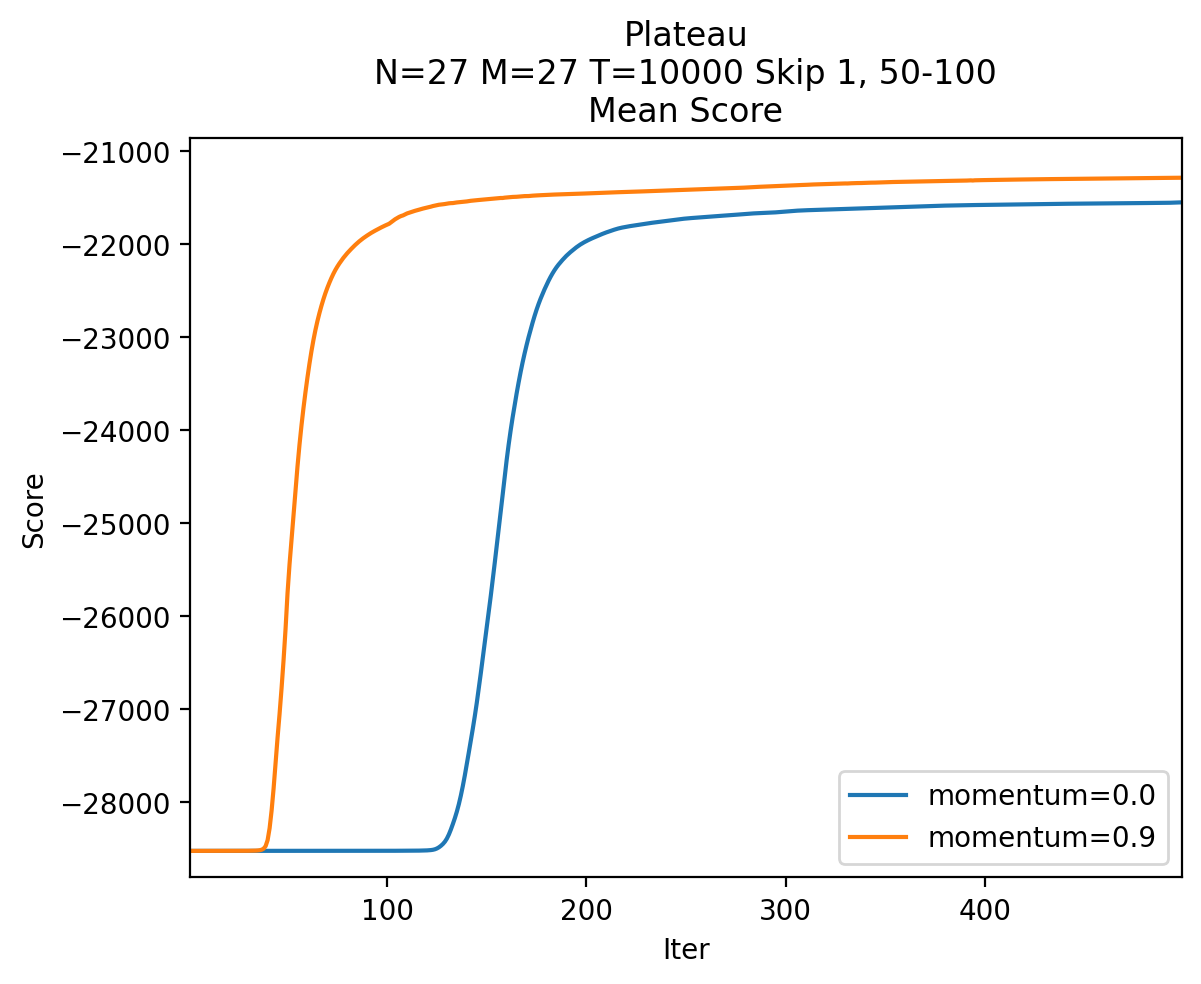}
\end{subfigure}
\begin{subfigure}{.385\textwidth}
  \centering
  \includegraphics[width=1.0\linewidth]{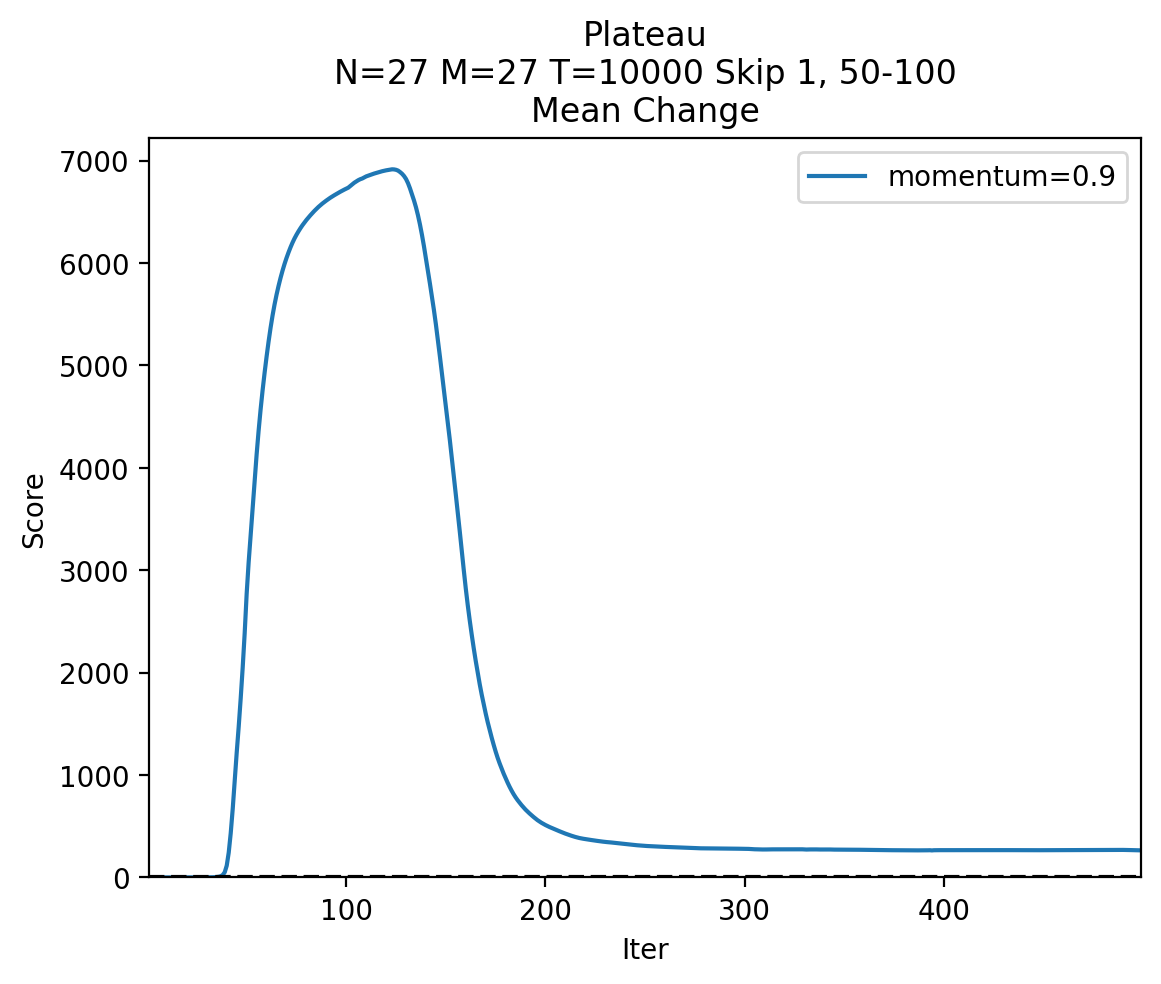}
\end{subfigure}
\caption{Plateau example, skipping iterations 1 and 50 to 100}\label{fig:sched_plateau2}
\end{figure}

Extending this method to also exclude momentum for the first iteration eliminates the negative score 
differential in the first few iterations.  As shown in Figure~\ref{fig:sched_plateau2}, the combination 
of skipping momentum for both problematic periods results in a smoother training curve.  Removing momentum 
at the first iteration causes the model to take slightly longer to start converging, but this is offset by higher 
peak increases in score.  The downside of such a manual scheduling approach is that it requires prior testing or knowledge of the general period in which overshoot occurs for the given model.  Ideally, an adaptive schedule capable of dynamically modifying momentum based on current training behavior would be used.

\subsection{Malware Classification}

We now consider HMMs trained on malware opcode sequences to test the effects of momentum on model classification accuracy. 
These malware classification experiments aim to determine if the increases in score shown with English text translate to improvements in a practical application.
We conduct two sets of experiments, involving distinct malware datasets. 

\subsubsection{Malicia Dataset}\label{malicia_desc}

Our first set of malware experiments involves the popular Malicia malware dataset~\cite{malicia}.  The Malicia variant used contains~8283 malware executables from various malware families, along with mnemonic opcode sequences for each executable.  These sequences were generated using the IDA Pro disassembler~\cite{idafree}.  The dataset is dominated by three families: Winwebsec, Zeroaccess, and Zbot.  As shown in Table~\ref{maliciaCounts}, these three families account for about~94\%\ of the dataset, with the next largest family containing only~74 executables.  Our Malicia experiments are based on these three dominant families.

\begin{table}[!htb]
\centering
\caption{Number of Malicia malware files by family}\label{maliciaCounts}
\adjustbox{scale=0.85}{
\begin{tabular}{c|c}
\midrule\midrule
Family & Samples \\
\midrule
Winwebsec & 4360\\
Zbot & 2136\\
Zeroaccess & 1305\\
All other & \zz482\\
\midrule\midrule
\end{tabular}
}
\end{table}

To minimize model complexity, and consistent with previous work,
only the top~29 most frequently occurring opcodes across the three families are used.  
All opcodes outside this top~29 are combined into a single ``other'' category, 
giving us a total of~30 distinct observation symbols.  This ``other'' opcode category 
contains less than~5\%\ of the total opcodes across the three families, 
as can be seen in Table~\ref{maliciaOpcodePercents}.  The effectiveness of grouping infrequently occurring opcodes was demonstrated by~\cite{malwareGHMM}.  Because of this grouping method, 
a vocabulary size of $M=30$ is used for all malware experiments in this section.

\begin{table}[!htb]
\centering
\caption{Percentage of opcodes in top 29}\label{maliciaOpcodePercents}
\adjustbox{scale=0.85}{
\begin{tabular}{c|c}
\midrule\midrule
Family & Percentage\\
\midrule
Winwebsec & 96.31\%\\
Zbot & 92.42\%\\
Zeroaccess & 95.00\%\\
\midrule
All samples & 95.34\%\\
\midrule\midrule
\end{tabular}
}
\end{table}

To determine the effectiveness of momentum on individual models, we train an HMM for each family using opcode sequences belonging to that family.  
Classification is performed by scoring a given test sample against each family's model.  The higher the score for a model, the higher the likelihood that the given test sample belongs to that model's family.  Because the magnitude of the score is dependent on sequence length, scores are normalized by the length of the test sequence to produce a log-likelihood per opcode score.  
After generating the three family scores for each test sample, the resulting score vectors are used to train a linear 
support vector machine~(SVM) for classification.  Five-fold stratified cross validation is used when training the SVM.  
The mean balanced accuracy across each fold is used as a metric to determine the effectiveness of a given trio of family models.  This process is repeated~100 times for each experiment with random HMM initializations, and the results averaged to produce a final accuracy metric for a set of training parameters.  

Training sequences for a given family are chosen by randomly sampling opcode files for that family without replacement until a combined total length of~$T$ is reached.  The test set for a given run is comprised of all non-training samples from each family for that run.  

For our experiments, we use a default sequence length of~$T=10{,}000$.  
Due to this relatively low training sequence length and opcodes being selected across all families, it is not uncommon for opcodes to occur in test that are not seen during training.  This causes computation errors in HMM scores, making that test sample unusable for the SVM.  Further experiments using~$T=1000$ result in such errors for almost every test sequence.  To solve this problem, a smoothing value of~0.01 is used to prevent zero probabilities in the models.

\begin{figure}[!htb]
\centering
\begin{subfigure}{.385\textwidth}
  \centering
  \includegraphics[width=1.0\linewidth]{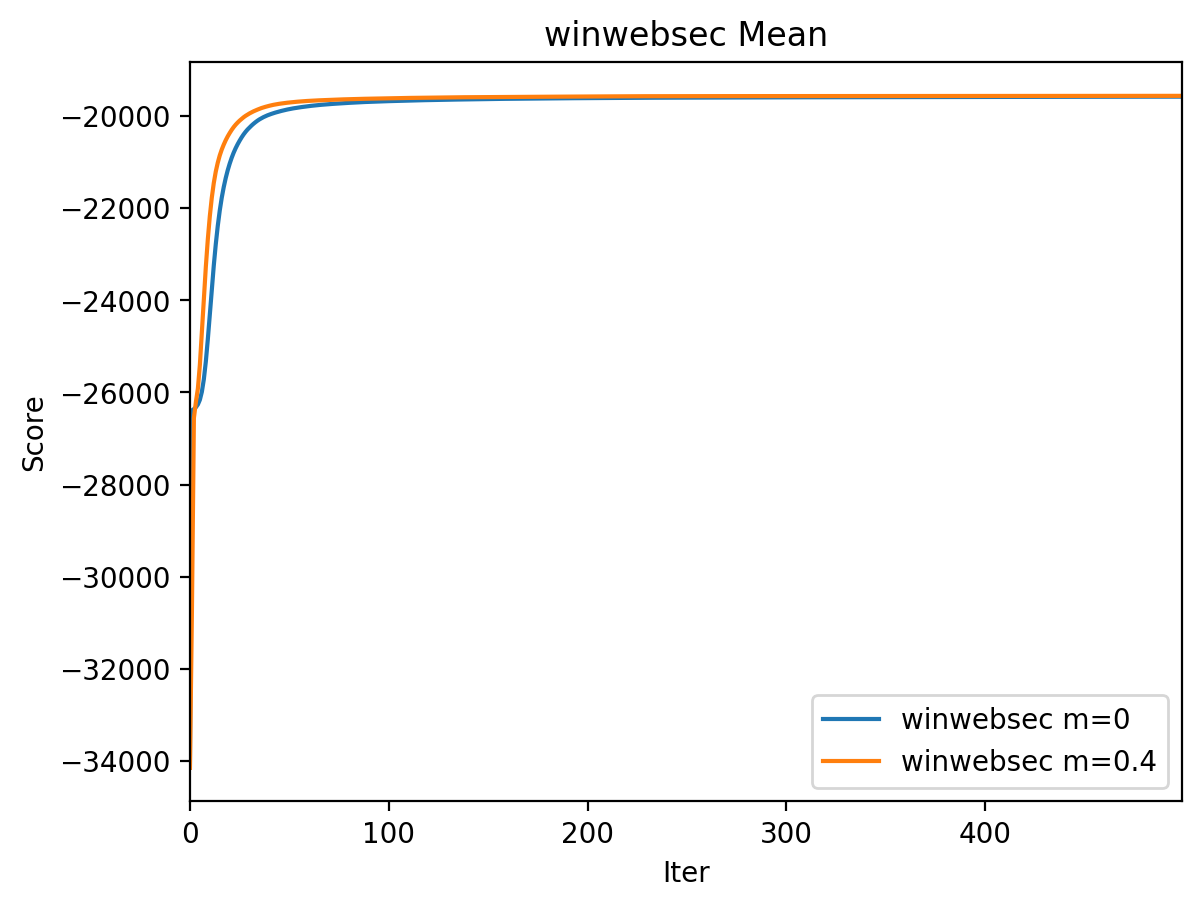}
\end{subfigure}
\begin{subfigure}{.385\textwidth}
  \centering
  \includegraphics[width=1.0\linewidth]{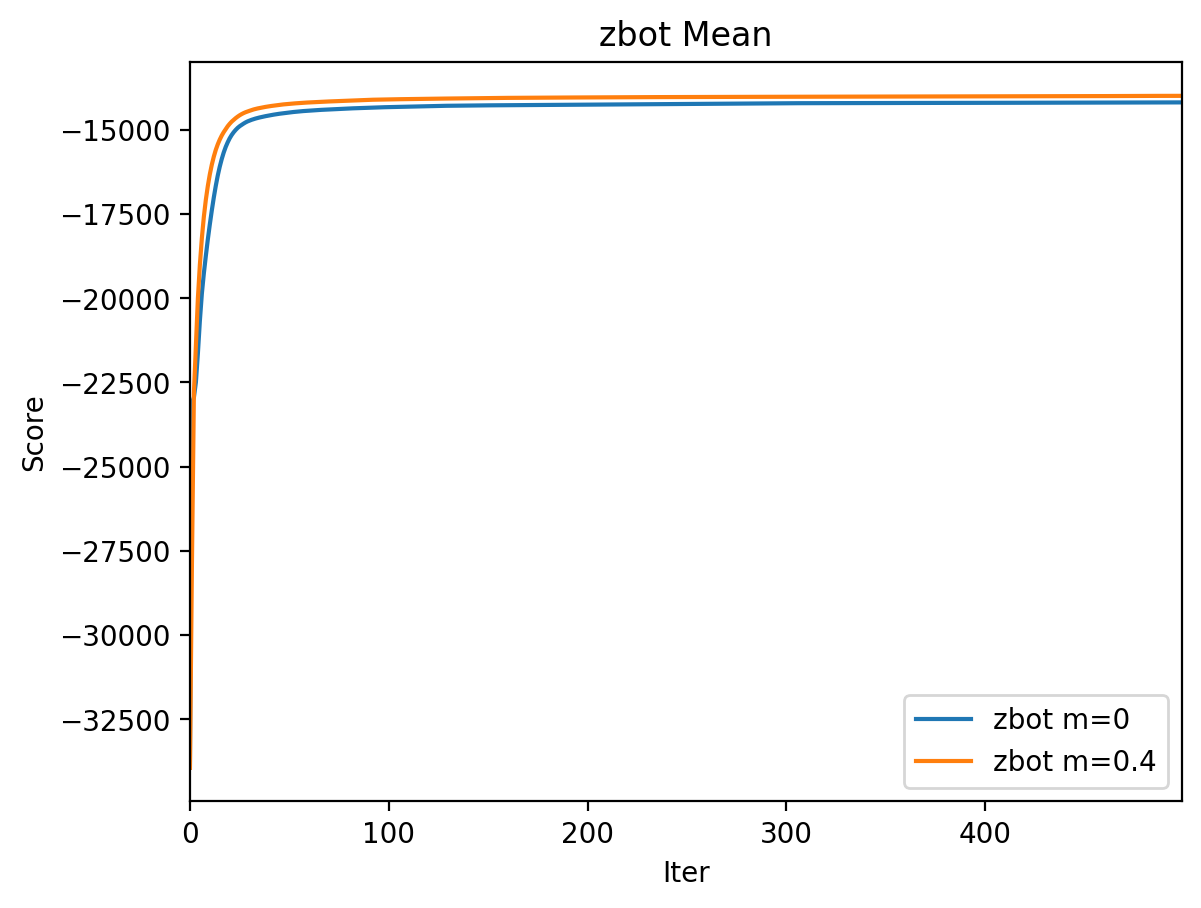}
\end{subfigure}\\[1ex]
\begin{subfigure}{\textwidth}
  \centering
  \includegraphics[width=0.4\linewidth]{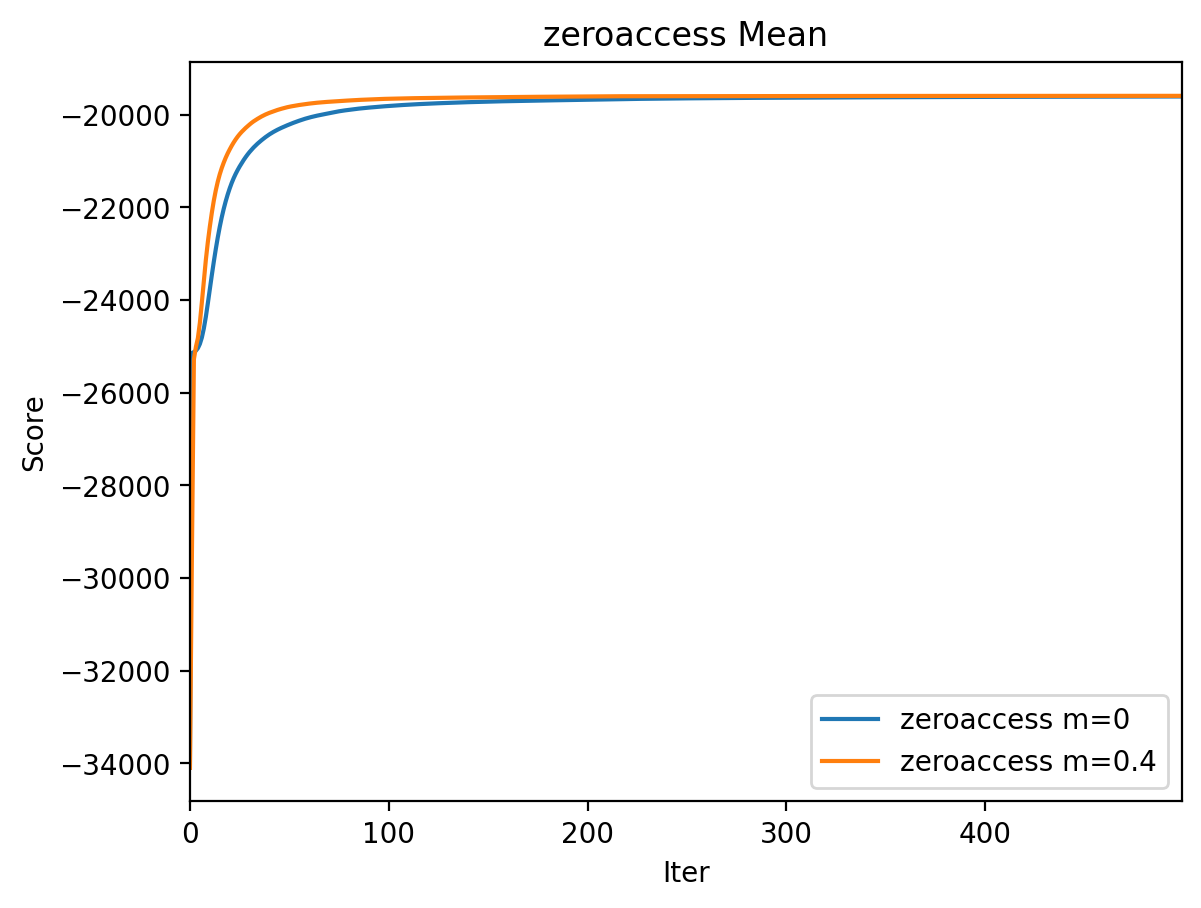}
\end{subfigure}
\caption{Mean training curve for each family with Nesterov momentum $\nest=0.4$}
\label{fig:malicia_families_500iters}
\end{figure}

Tests are performed with~$N=20$, $M=30$, $T=10{,}000$, and~$smoothing=0.01$ 
with a Nesterov momentum value of~$\nest=0.4$.  The relatively high~$N$ was chosen 
based on similar experiments in~\cite{malwareGHMM}, where more noticeable score differences are detected at higher numbers of hidden states.  Each experiment is repeated~100 times for a total of~100 models trained per family.  Figure~\ref{fig:malicia_families_500iters} shows that momentum 
produces similar changes to average training behavior as with English text.  Training curves showing the effects of momentum differ slightly between each family, but follow the general trend observed in Section~\ref{momentum_results}.  

After~500 iterations, the Zbot models show a mean score increase of~194, 
while Winwebsec and Zeroaccess increase by 17 and 13 respectively.  
However, SVM classification accuracy does not demonstrate any significant change 
with momentum, decreasing from an average balanced accuracy of 96.03\% to 95.92\%.  
Accuracies with and without momentum for models trained for varying numbers of iterations are listed in Table~\ref{svm_accuracy_iters}.  Changing the number of iterations does not significantly affect SVM classification accuracy.  Even after only two iterations, the HMM to SVM approach reaches over 90\% accuracy.  This indicates that the SVM is contributing the majority of the classification performance, not the HMMs.

\begin{table}[!htb]
\centering
\caption{SVM classification accuracy at various iterations}\label{svm_accuracy_iters}
\adjustbox{scale=0.85}{
\begin{tabular}{c|ccr}
\midrule\midrule
Iteration & \makecell{Average \\ $\Delta$score} & \makecell{No momentum \\ accuracy} & \makecell{Nesterov \\ accuracy}\\
\midrule
\zz\zz2 & $\llap{$-$}4520$ & 0.9362 & 0.9370\zz\\
\zz15 & 3501 & 0.9534 & 0.9585\zz\\
\zz25 & \zz508 & 0.9585 & 0.9600\zz\\
\zz50 & \zz265 & 0.9602 & 0.9605\zz\\
200 & \zz106 & 0.9606 & 0.9589\zz\\
500 & \zz\zz75 & 0.9603 & 0.9592\zz\\
\midrule\midrule
\end{tabular}
}
\end{table}

Because the intermediate SVM step may be hiding (or compensating for) changes in the model, tests are instead performed using the HMM scores directly via one-vs-rest classification.  
The classification performance of each model is measured by computing a 
receiver operating characteristic~(ROC) curve.  A ROC curve is obtained by plotting the true positive rate vs the false positive rate over all possible thresholds~\cite{stampBook}.  The area under the curve~(AUC) is used as a metric to produce a measurable value for each ROC curve.  

Table~\ref{tab:malicia_500_iters_AUC} shows that after~500 iterations, the combined AUC of all families is almost identical with or without momentum.  
While Zbot's larger increase in score results in a minor AUC increase of~$0.0059$,  there is minimal change for Winwebsec, and a decrease in average AUC of~$-0.0025$ for Zeroaccess.  This indicates that the small score differences on the tail end of training may not meaningfully contribute to model performance, and may even be an artifact of overfitting.  

\begin{table}[!htb]
\caption{$\Delta$score vs $\Delta$AUC at 500 iterations}
\label{tab:malicia_500_iters_AUC}
\centering
\adjustbox{scale=0.85}{
\begin{tabular}{c|ccc}
    \midrule\midrule
    Family & $\Delta$score & No momentum AUC & Nesterov AUC\\
    \midrule
    Winwebsec & \zz17 & 0.8779 & 0.8782 \\
    Zbot & 194 & 0.8750 & 0.8809 \\
    Zeroaccess & \zz13 & 0.7609 & 0.7584 \\
    \midrule
    All samples & \zz75 & 0.8379 & 0.8392 \\
    \midrule\midrule
\end{tabular}
}
\end{table}

\begin{figure}[!htb]
\centering
\begin{subfigure}{.385\textwidth}
  \centering
  \includegraphics[width=1.0\linewidth]{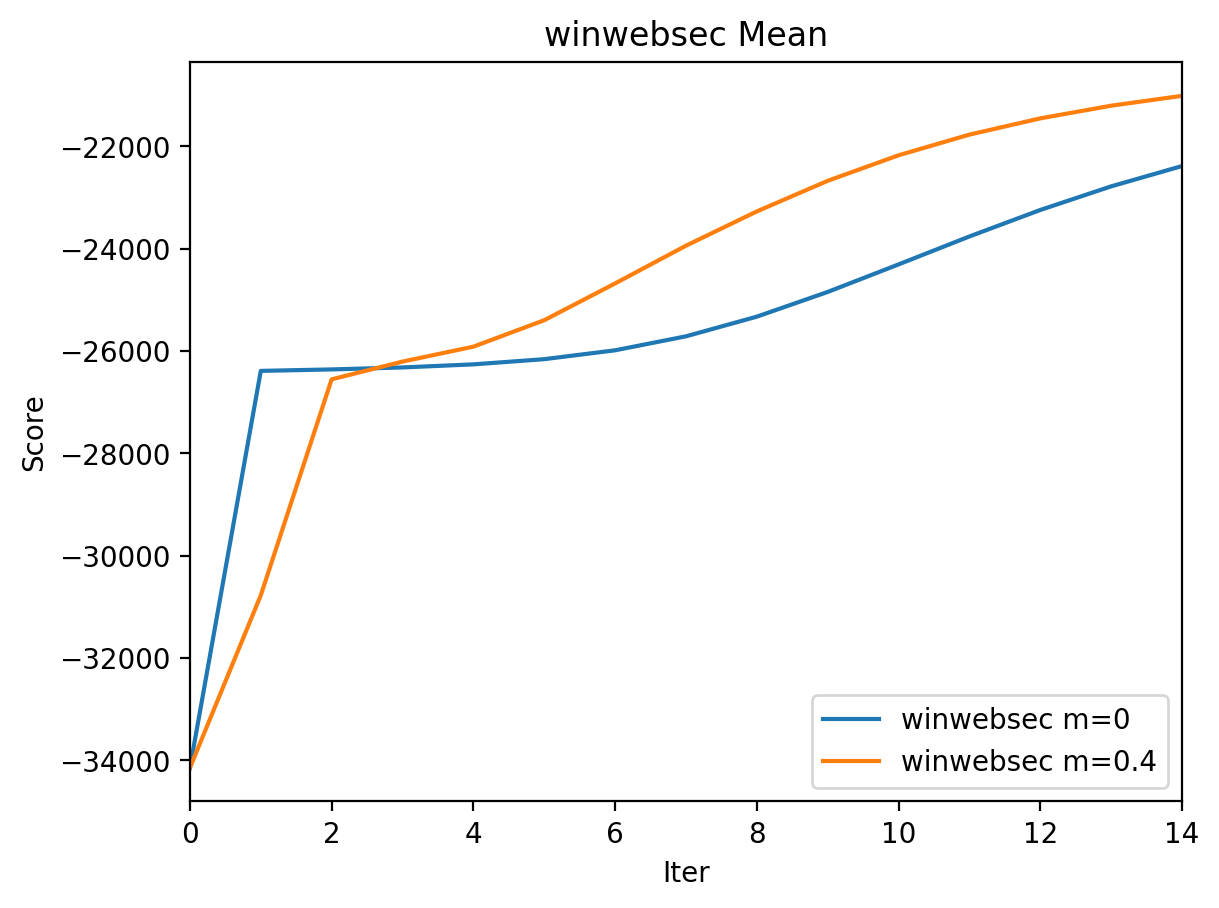}
\end{subfigure}
\begin{subfigure}{.385\textwidth}
  \centering
  \includegraphics[width=1.0\linewidth]{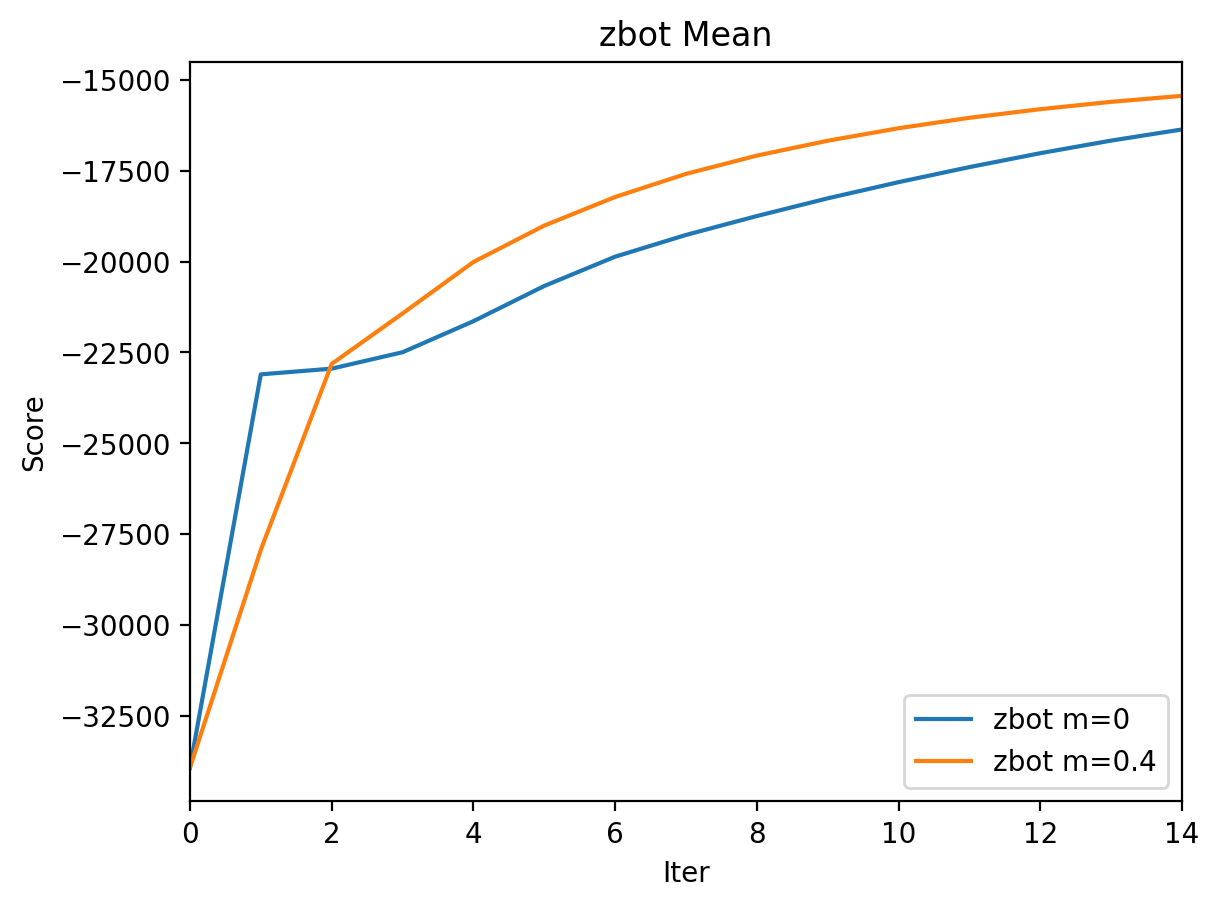}
\end{subfigure}\\[1ex]
\begin{subfigure}{\textwidth}
  \centering
  \includegraphics[width=0.4\linewidth]{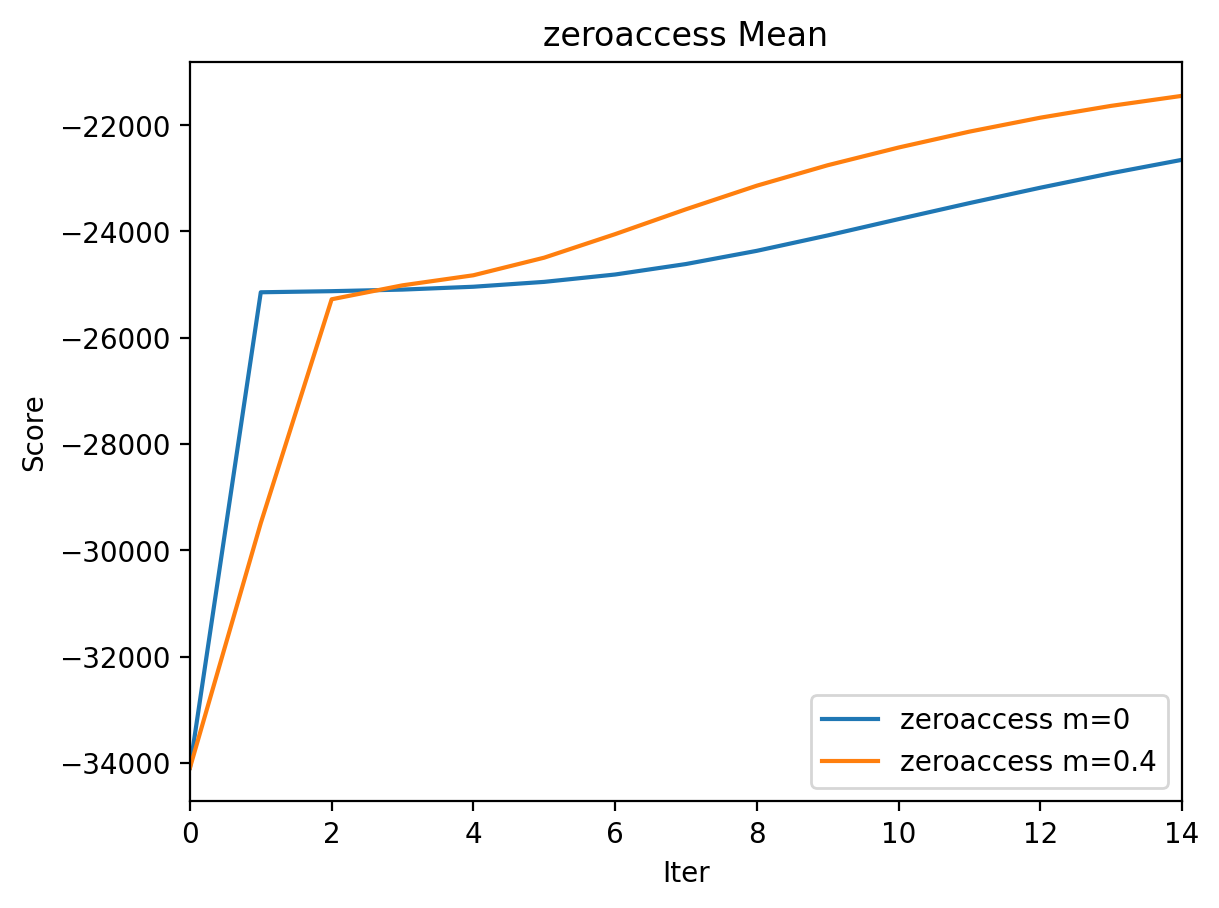}
\end{subfigure}
\caption{Mean scores for first 15 iterations with $\nest=0.4$}
\label{fig:malicia_families_15iters}
\end{figure}

To test this, we compute the AUC-ROC again after only~15 iterations, during the period of maximum changes in scores with momentum.  Figure~\ref{fig:malicia_families_15iters} depicts the mean 
training scores for each family over the first~15 iterations.  
Table~\ref{tab:malicia_15_iters} compares the mean change in score for each family to the AUC after~15 iterations.  
The significant score differences result in a much more significant combined AUC increase of 0.023 with momentum.  This shows that momentum is able to improve model classification performance during this period.  

\begin{table}[!htb]
\caption{$\Delta$score vs $\Delta$AUC at 15 iterations}
\label{tab:malicia_15_iters}
\centering
\adjustbox{scale=0.85}{
\begin{tabular}{c|ccc}
    \midrule\midrule
    Family & $\Delta$score & No momentum AUC & Nesterov AUC\\
    \midrule
    Winwebsec & 1372 & 0.8533 & 0.8606 \\
    Zbot & \zz925 & 0.8302 & 0.8501 \\
    Zeroaccess & 1204 & 0.6911 & 0.7333 \\
    \midrule
    Total & 1167 & 0.7915 & 0.8147 \\
    \midrule\midrule
\end{tabular}
}
\end{table}

Training with Nesterov momentum for the full~500 iterations results in an additional AUC increase of~0.025 as compared to only~15 iterations.  This indicates that the usefulness of momentum may be restricted to applications which train for a small number of iterations, as when training time is limited,
resources for training are limited, a large number of models must be trained in a short period of
time, and so on. Table~\ref{tab:malicia_auc_selected_iters} lists the total AUC at various points in training.  
By~50 iterations, AUC without momentum closes the gap, and the AUC only shows minor increases after that point.

\begin{table}[!htb]
\caption{Total $\Delta$score vs $\Delta$AUC at select iterations}
\label{tab:malicia_auc_selected_iters}
\centering
\adjustbox{scale=0.85}{
\begin{tabular}{c|rcc}
    \midrule\midrule
    Iterations & $\Delta$score & No momentum AUC & Nesterov AUC\\
    \midrule
    \zz\zz2 & $-4520$\zz & 0.7431 & 0.7444 \\
    \zz15 & 1167\zz & 0.7915 & 0.8147 \\
    \zz25 & 508\zz & 0.8189 & 0.8332 \\
    \zz50 & 265\zz & 0.8338 & 0.8378 \\
    200 & 107\zz & 0.8378 & 0.8387 \\
    500 & 75\zz & 0.8379 & 0.8392 \\
    \midrule\midrule
\end{tabular}
}
\end{table}

Unsurprisingly, increasing the number of restarts per run to~5 results in a minor increase in score and AUC.  Table~\ref{tab:malicia_auc_5_restarts} compares score differences and AUC after~15 iterations for~1 restart versus~5 restarts.  

\begin{table}[!htb]
\caption{Total $\Delta$score vs $\Delta$AUC at 15 iterations, 1 vs 5 restarts}
\label{tab:malicia_auc_5_restarts}
\centering
\adjustbox{scale=0.85}{
\begin{tabular}{c|ccc}
    \midrule\midrule
    Restarts & $\Delta$score & No momentum AUC & Nesterov AUC\\
    \midrule
    1 & 1167 & 0.7915 & 0.8147 \\
    5 & 1169 & 0.7955 & 0.8203 \\
    \midrule\midrule
\end{tabular}
}
\end{table}

\begin{figure}[!htb]
\centering\advance\tabcolsep by -10pt
\begin{tabular}{cc}
  \includegraphics[width=0.45\linewidth]{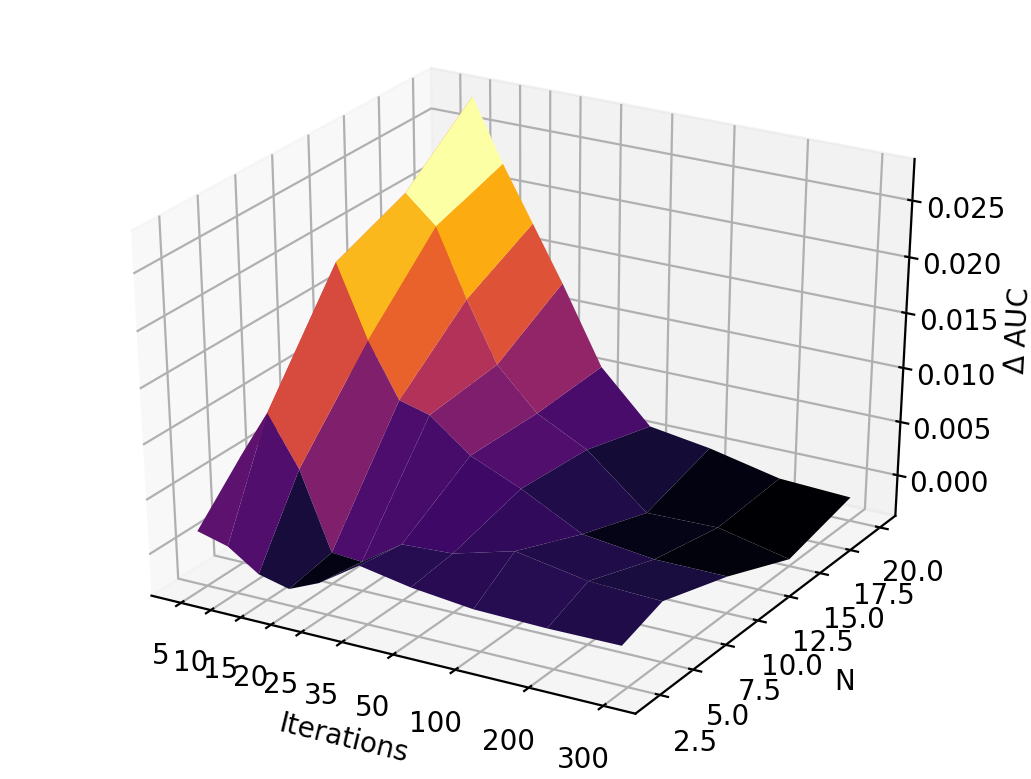} 
  &
  \includegraphics[width=0.45\linewidth]{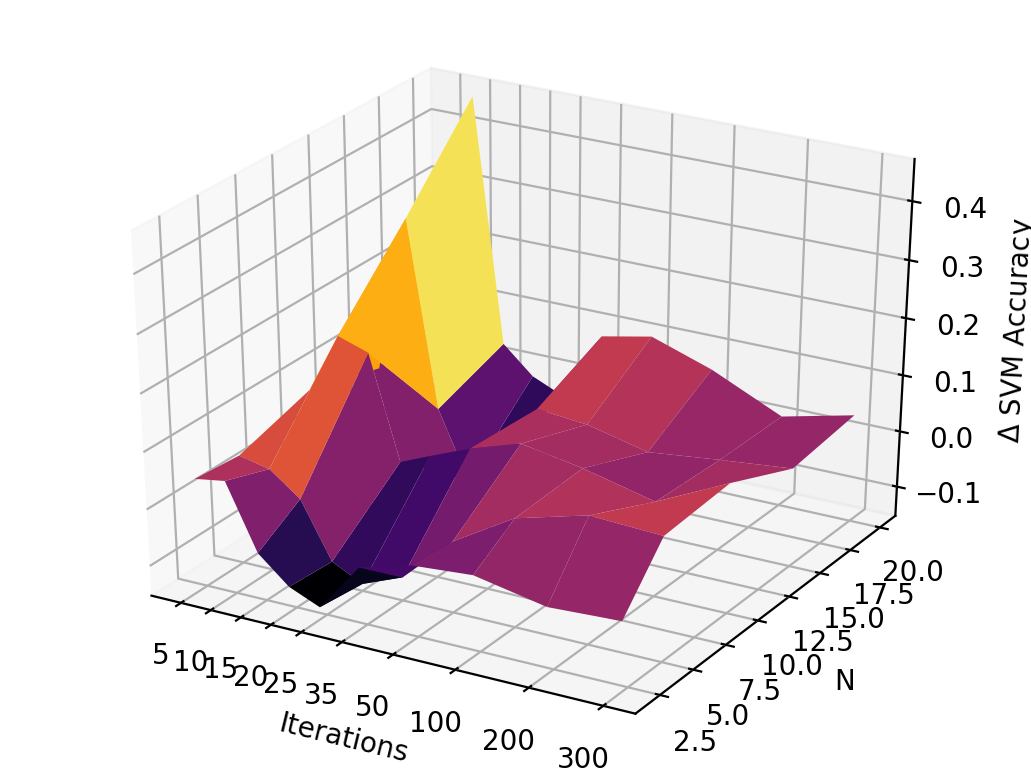} \\
  (a) AUC
  &
  (b) SVM accuracy
  \end{tabular}
\caption{Change in AUC and accuracy due to momentum}\label{fig:varyN_AUC_SVM}
\end{figure}

Additionally, we experiment with various choices of~$N$ in order to observe how the chosen number of hidden states influences the performance of momentum.  
Each model is trained for~300 iterations with $M=30$, $T=10{,}000$, 
and a Nesterov momentum of~$\nest=0.4$.  Changes in AUC and accuracy are compared 
between~$N\in\{2, 5, 10, 15, 20\}$ hidden states, with scores computed 
at iterations~$\{5,10, 15, 20, 25, 35, 50, 100, 200, 300\}$.  
Figure~\ref{fig:varyN_AUC_SVM}(a) depicts the changes in AUC 
at the specified iterations for each~$N$ tested, while
Figure~\ref{fig:varyN_AUC_SVM}(b) shows analogous results for the SVM accuracy.  
At early iterations, higher numbers of hidden states result in larger changes in AUC with momentum, while later iterations show minimal change regardless of~$N$.  SVM classification is similar, with greater increases in accuracy during early training at higher~$N$.  These results tend to indicate that momentum is 
more beneficial for more complex models (i.e., larger~$N$), but has minimal impact when dealing with simpler models, where~$N$ is small.  These findings align well with the English text experiments in Section~\ref{highN}, above.

%

\begin{figure}[!htb]
\centering\advance\tabcolsep by -10pt
\begin{tabular}{cc}
  \includegraphics[width=0.45\linewidth]{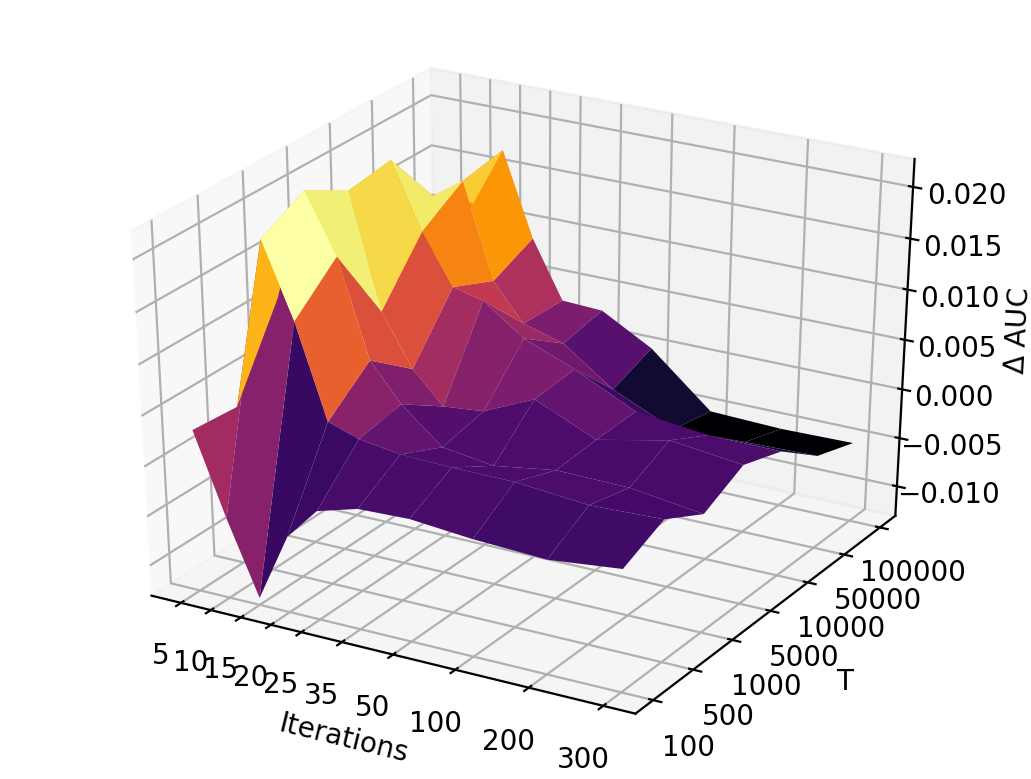}
  &
  \includegraphics[width=0.45\linewidth]{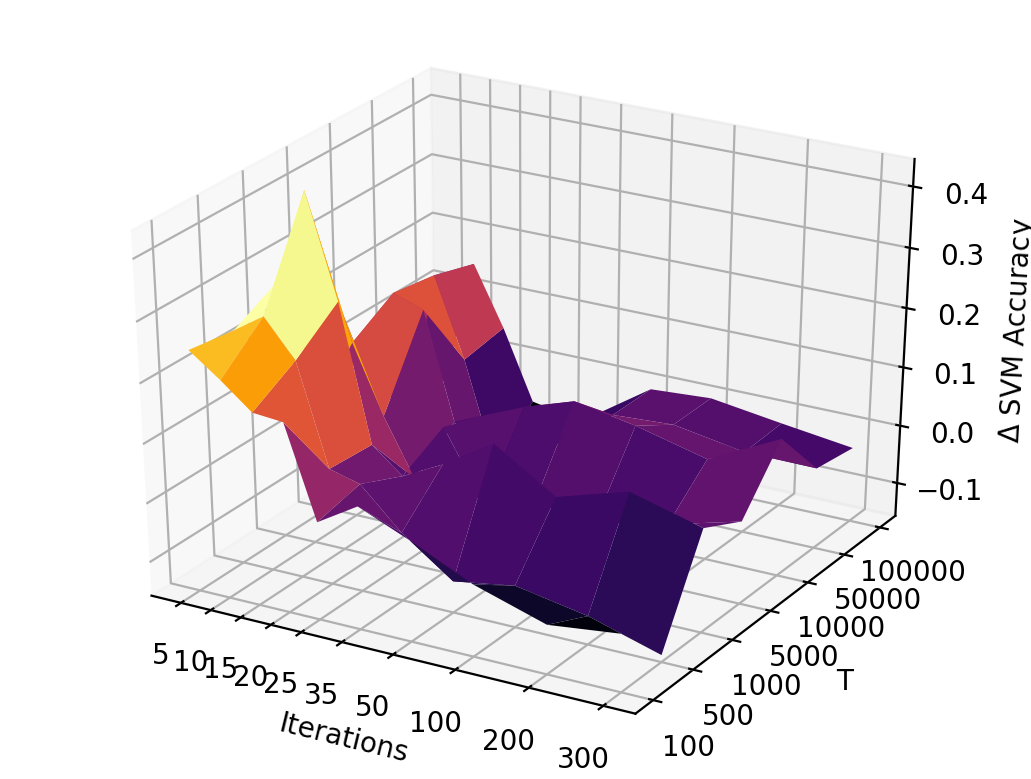}
  \\
  (a) AUC
  &
  (b) SVM accuracy
  \end{tabular}
\caption{Changes in AUC and SVM for cold start}\label{fig:varyT_AUC_SVM}
\end{figure}

Similar tests are conducted to determine how the amount of training data affects model performance with momentum.  
Models are trained with and without momentum while varying the total length of the observation sequences 
from~$T=100$ to~$T=100{,}000$.  HMMs are trained with~$N=10$ and~$M=30$, and Nesterov momentum~$\nest=0.4$, 
with scoring performed at each number of iterations in the set~$\{5, 10, 15, 20, 25, 35, 50, 100, 200, 300\}$.  
Figure~\ref{fig:varyT_AUC_SVM}(a) shows that changes in AUC remain relatively consistent despite significant variance 
in the amount of training data, with the exception of~$T=100$.  At~$T=100$, momentum does not produce the expected increase in AUC at early iterations.  The small amount of data might suggest that momentum simply causes the model to overfit more quickly.  For SVM accuracy, Figure~\ref{fig:varyT_AUC_SVM}(b) shows that early momentum actually seems to perform slightly better with lower amounts of training data, peaking at~$T=1000$.  In addition, at~$T=50{,}000$ and above, 
momentum results in a small negative dip in score between~25 and~35 iterations.  The overall curve is much rougher and less consistent than with AUC, but changes in accuracy at later iterations are minimal or slightly negative for all~$T$. These results show that momentum is worth considering for the ``cold start''
problems, where training data is limited.

\subsubsection{Extended Malware Dataset}

In order to experiment on a more challenging case, we further experiment with a large malware dataset containing~131,072 malware executables~\cite{SKim}.  Of these, 58,679 are labeled as belonging to a known malware family.  We use the IDA Free disassembler~\cite{idafree} to produce opcode sequences 
for each labeled sample by analyzing each executable and outputting a corresponding assembly file.  We then extract mnemonic opcodes in sequential order from the exported \texttt{.asm} files to construct an opcode sequence for each executable.  IDA's default settings are used when analyzing all assembly files.  Final observation sequences are constructed using the same methods as in Section~\ref{malicia_desc} for the Malicia dataset. 

\begin{figure}[!htb]
    \centering
    \includegraphics[width=0.7\linewidth]{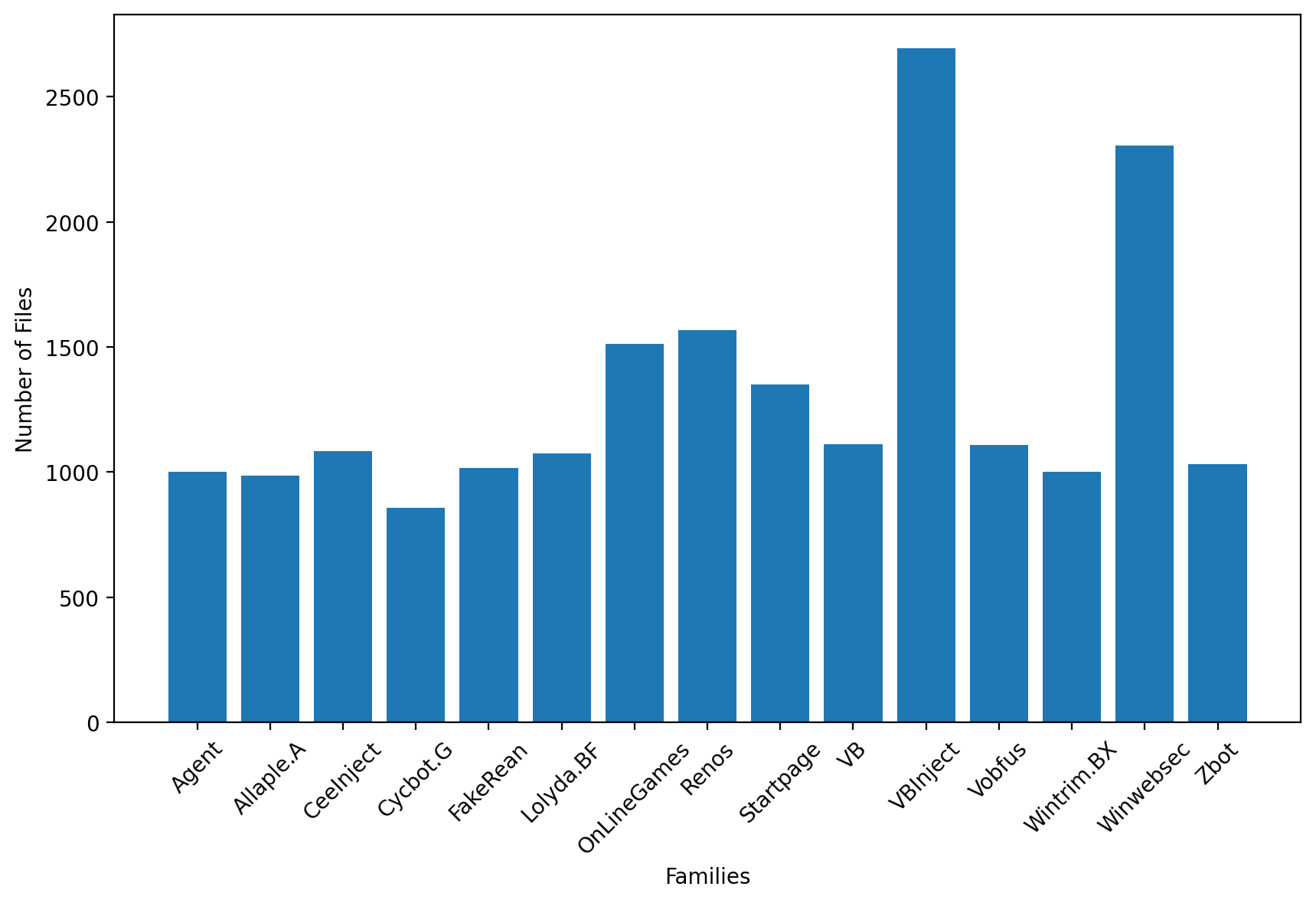}
    \caption{Number of executables for the 15 families used}
    \label{fig:malware_stats_counts}
\end{figure}

\begin{figure}[!htb]
    \centering
    \includegraphics[width=0.7\linewidth]{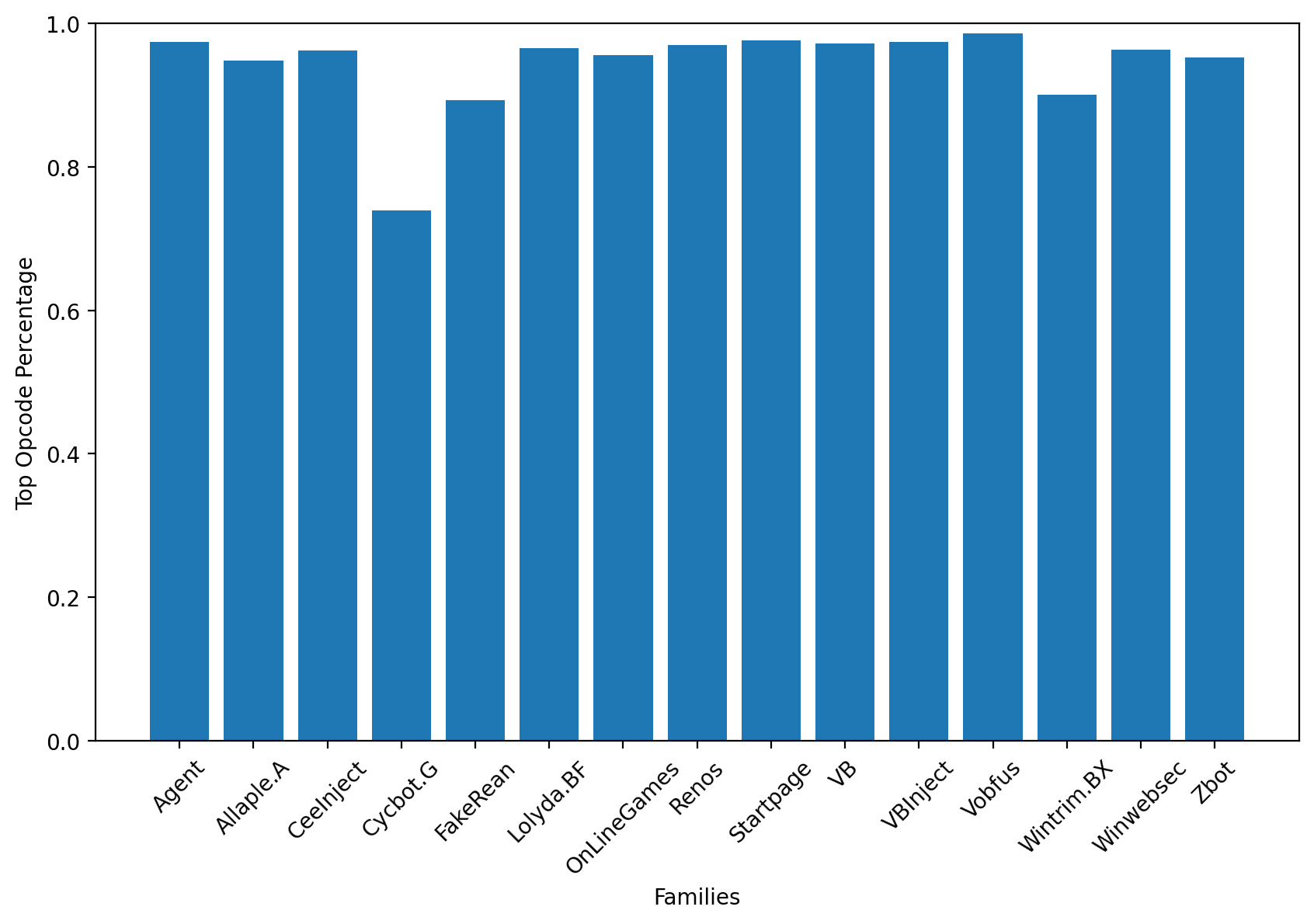}
    \caption{Top 29 opcodes percentage for each family}
    \label{fig:malware_stats_family_top_opcode_pct}
\end{figure}

\begin{figure}[!htb]
    \centering
    \includegraphics[width=0.7\linewidth]{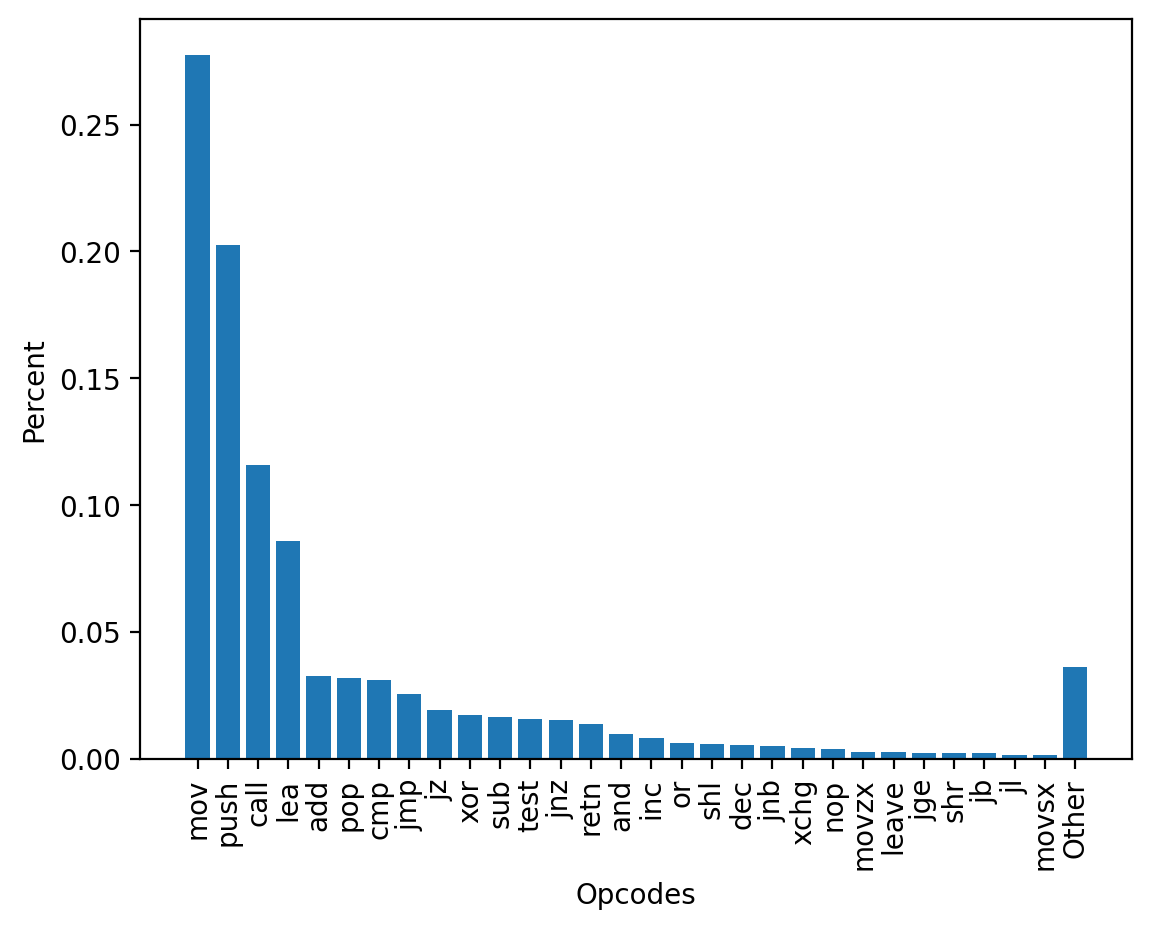}
    \caption{Opcode frequency for 15 families}
    \label{fig:malware_stats_opcodefreq}
\end{figure}

Our experiments consist of multiclass classification utilizing~15 of the largest families, for a total of~19,705 malware samples.  Figure~\ref{fig:malware_stats_counts} lists these families, 
along with the total number of samples for each.  As with the Malicia dataset, the most frequent~29 opcodes across these~15 families are considered as unique observations, with any other opcodes being categorized a single ``other'' observation.  The top~29 opcodes make up well over~90\%\ of observations for most families, as indicated in Figure~\ref{fig:malware_stats_family_top_opcode_pct}.  Cycbot.G is an outlier, with less frequent opcodes occurring at a rate of~26\%.  Figure~\ref{fig:malware_stats_opcodefreq} lists the observed opcodes and their frequencies.  

A single HMM is trained for each family based on training sequences belonging to that family of total length~$T=100{,}000$.  
Training splits for each family are generated as in Section~\ref{malicia_desc} by randomly selecting samples from a specific family until the combined sequence length reaches~$T$.  Once a sample is used in training, it cannot be selected again for training until all samples for that family are exhausted.  All samples not used for training from all families are combined into a single test set.  Test samples are scored against each family model to determine the likelihood of belonging to that family.  Each test score is normalized by sequence length to produce a log-likelihood per opcode score, ensuring that scores belonging to sequences of different lengths are comparable.  Scoring is performed after each Baum-Welch iteration 
in $\{5, 10, 15, 20, 25, 35, 50, 100, 200, 300\}$.
For each set of parameters, the entire process is repeated~100 times 
with a unique set of training sequences selected for each run.  

To compare the effects of momentum on model performance, identical experiments are conducted with and without momentum.  Experiments are performed using $N=10$, $M=30$, and $T=100{,}000$, and a smoothing value of~$s=0.001$.  Models are trained for~300 iterations with a single initial restart.  A Nesterov momentum of~$\nest=0.4$ is used for the momentum experiments, based on our findings discussed above for English text and Malicia malware experiments.  While previous experiments seem
to indicate more obvious differences at higher~$N$, we reduce the number of hidden states to~$N=10$ to lower the computation time.  

Classification performance of the models is again measured using AUC-ROC and SVM balanced accuracy.  One-vs-rest classification is used when computing the ROC curve for a model, with scores of samples belonging to the same family as the model being considered the positive class.  
SVM training is performed using five-fold stratified cross validation, with performance measured by averaging the balanced accuracy across each fold.  
SVMS are trained using the RBF kernel with~$C=10$, as it increases accuracy by about 6\% when compared to the linear kernel used in the Malicia experiments.  
The set of family test scores for each sample is used as the feature vectors for the SVM.

\begin{figure}[!htb]
    \centering
    \includegraphics[width=0.6\linewidth]{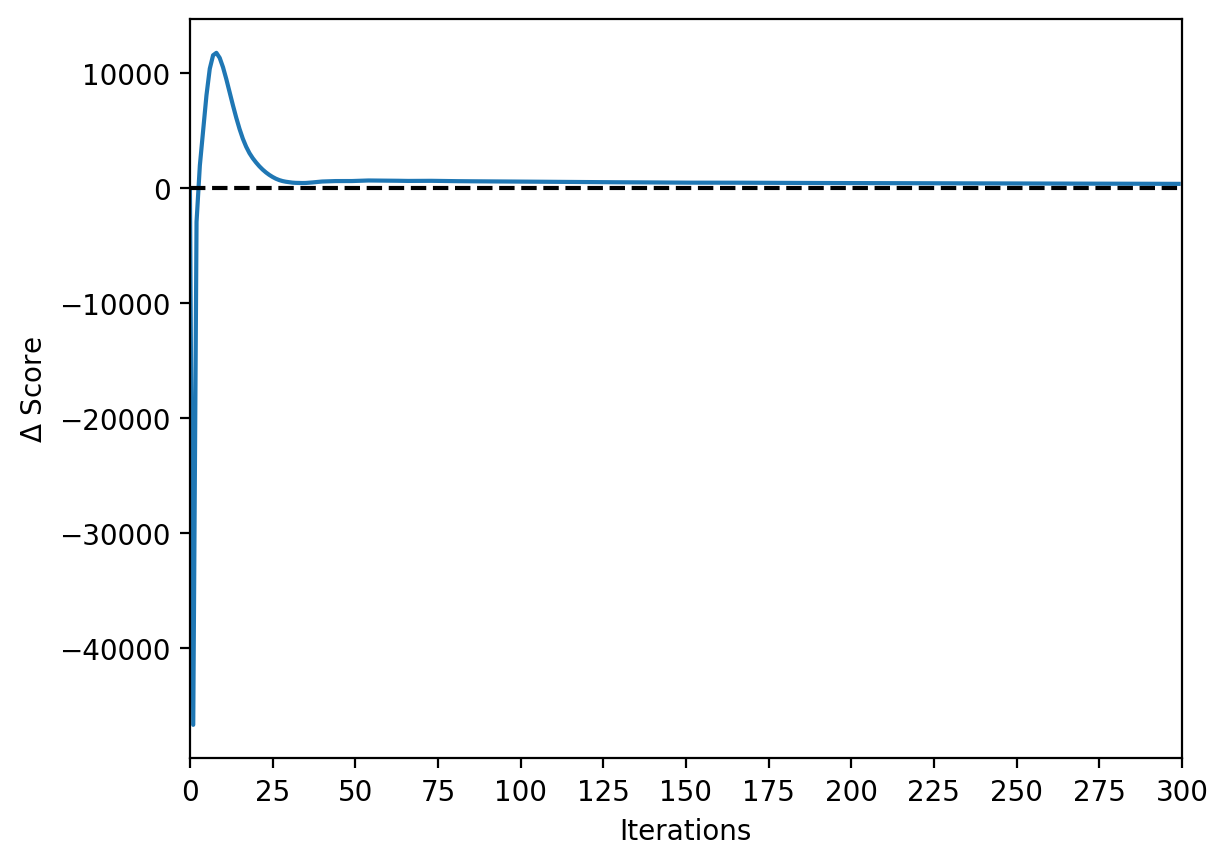}
    \caption{Mean change in model score with momentum}
    \label{fig:malware_total_mean_score_diff}
\end{figure}

Overall changes in model score due to momentum follow the pattern found in the text and Malicia experiments, with an initial sharp dip in score due to overshoot followed by improved scores during the period of initial convergence.  Figure~\ref{fig:malware_total_mean_score_diff} shows the mean change in score caused by momentum across all families, with significant increases in score leveling off after roughly 25 iterations.  

\begin{figure}[!htb]
    \centering
    \includegraphics[width=0.6\linewidth]{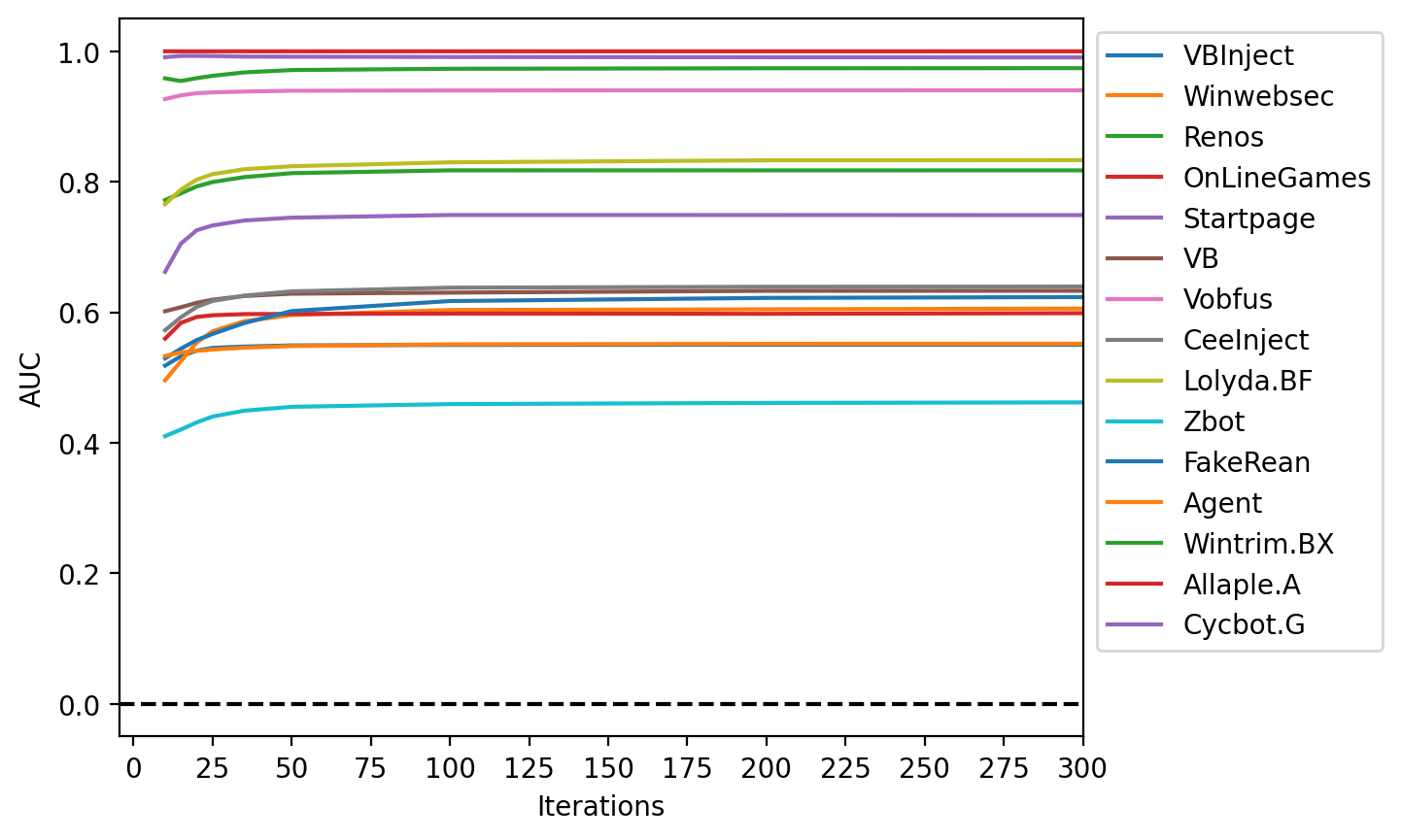}
    \caption{AUC-ROC without momentum for each family}
    \label{fig:malware_auc_all0}
\end{figure}

Inspecting the baseline models trained without momentum shows significant variance in AUC between each family, as demonstrated by Figure~\ref{fig:malware_auc_all0}.  Allaple.A and Cycbot.G have mean AUCs near~1.0, leaving little room for any positive changes with momentum.  Half of the~15 families produce models with a very poor 
average AUC between~$0.5$ and~$0.6$;  Zbot is lowest, with an average AUC just under~$0.5$.  
On average, model AUC without momentum stops increasing by~50 training iterations.  While many families demonstrate a poor AUC, this research focuses 
on the difference in AUC caused by momentum, rather than the absolute level of the AUC itself.  

\begin{figure}[!htb]
\centering
\begin{subfigure}[t]{.385\textwidth}
    \centering
    \includegraphics[width=1.0\linewidth]{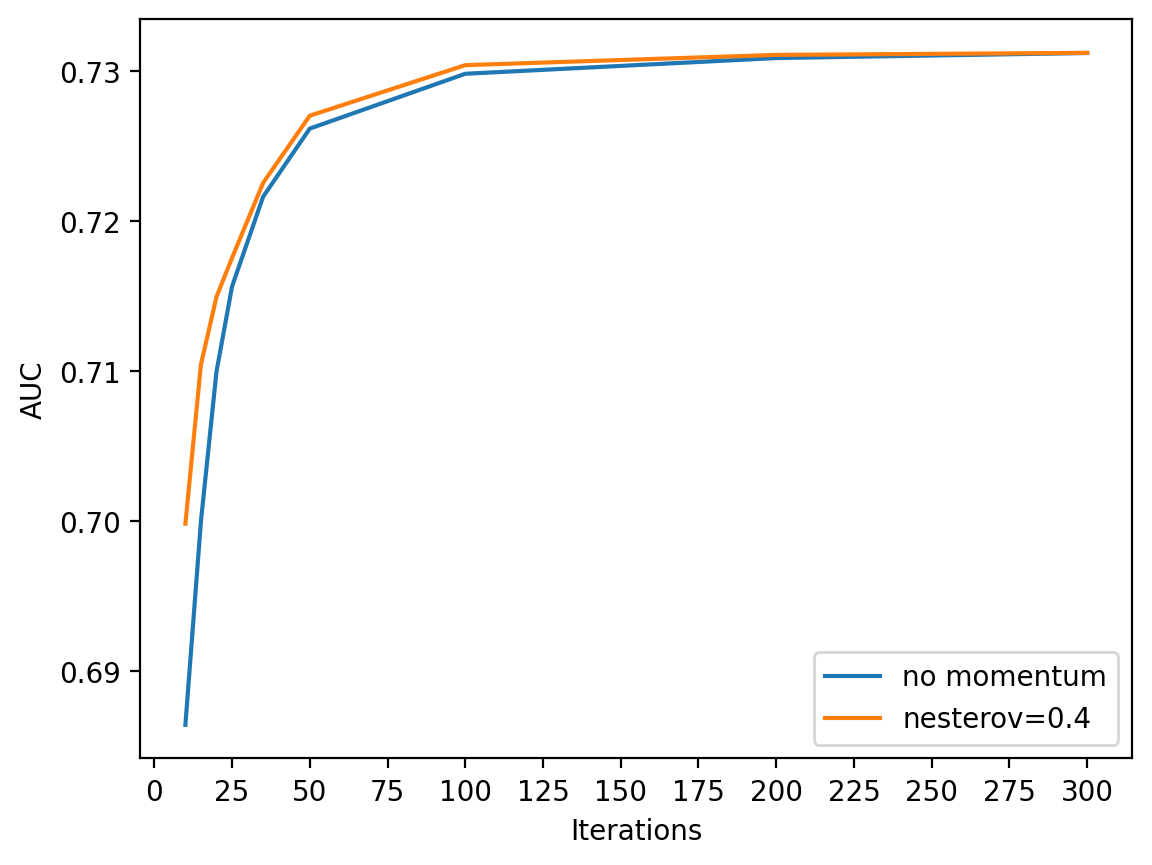}
    \caption{AUC-ROC}
\end{subfigure}
\begin{subfigure}[t]{.385\textwidth}
    \centering
    \includegraphics[width=1.0\linewidth]{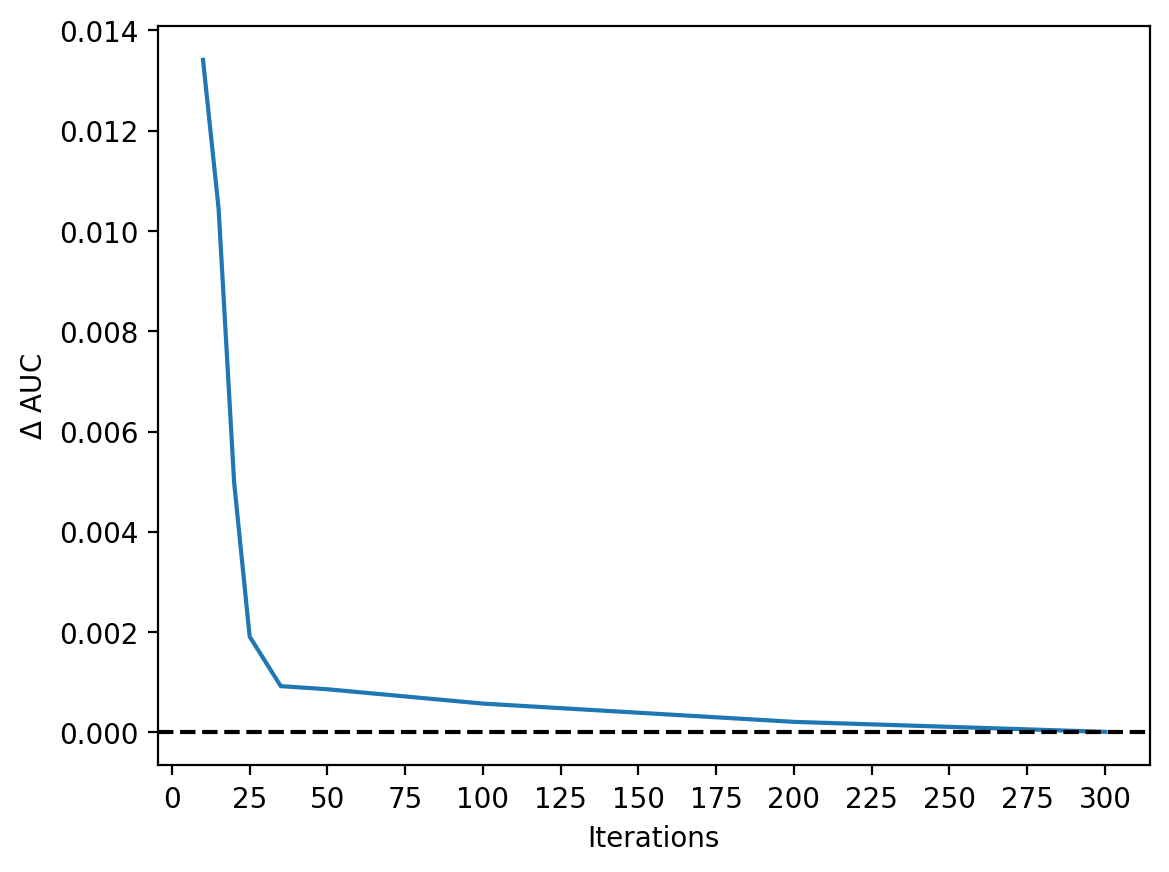}
    \caption{Change in AUC due to Momentum}
    \label{fig:malware_total_auc_diff}
\end{subfigure}
\caption{Comparison of total AUC-ROC for all families}
\label{fig:malware_aucs}
\end{figure}

Reproducing these same experiments with momentum results in small increases in AUC early in training,aligning with the period of significant score increases seen in Figure~\ref{fig:malware_total_mean_score_diff}. 
A comparison of the mean AUC across all families with and without momentum is shown in Figure~\ref{fig:malware_aucs}, with Figure~\ref{fig:malware_total_auc_diff} showing the explicit difference in AUC caused by momentum.  Momentum results in an average AUC increase of~$0.0134$ at~10 iterations, dropping to~$0.002$ by~25 iterations.  AUCs with and without momentum converge as the number 
of iterations increases, showing negligible differences by~200 iterations. 

\begin{figure}[!htb]
    \centering
    \includegraphics[width=0.75\linewidth]{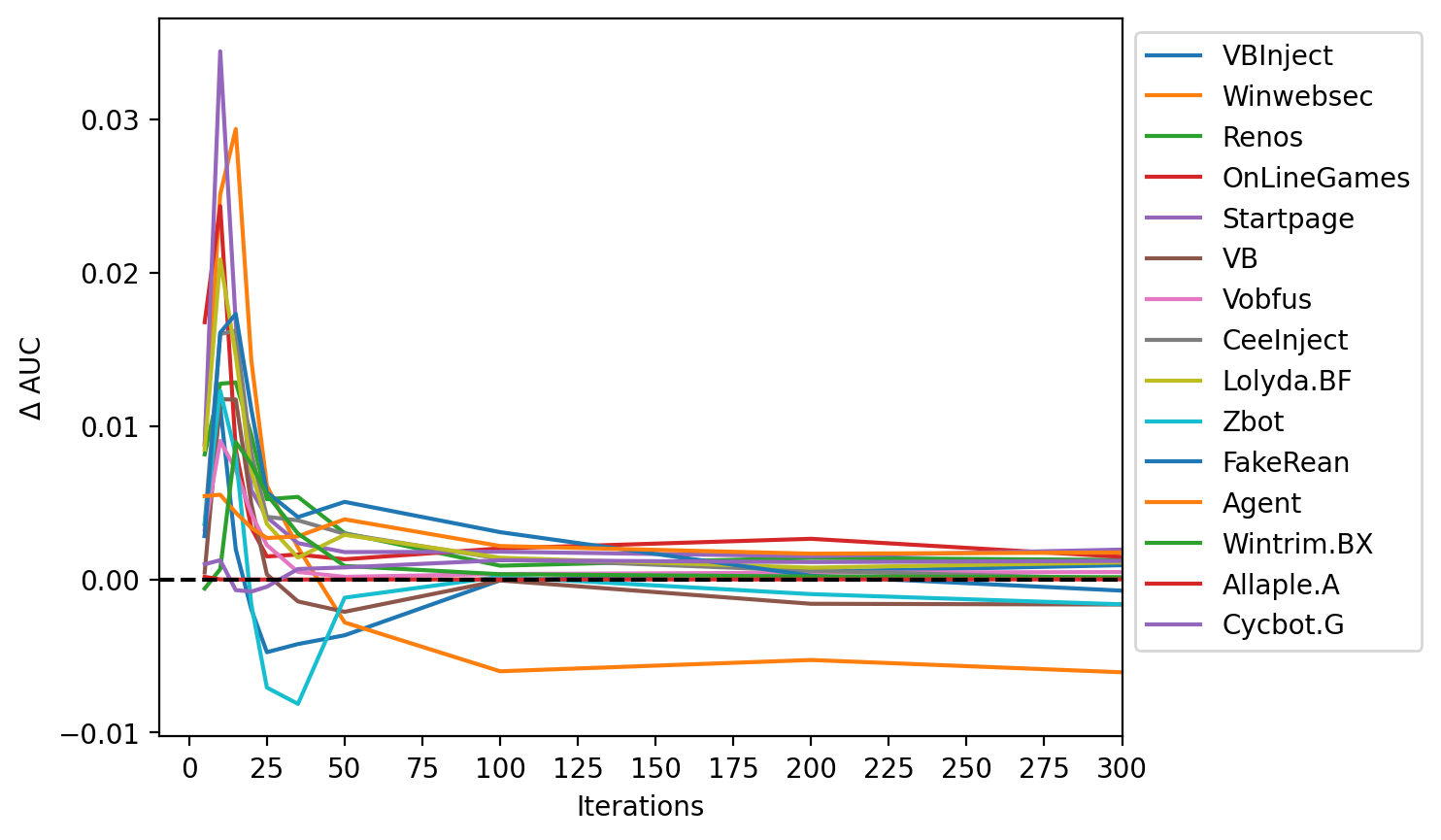}
    \caption{Change in AUC due to momentum}
    \label{fig:malware_auc_diffs}
\end{figure}

\begin{figure}[!htb]
    \centering
    \includegraphics[width=0.75\linewidth]{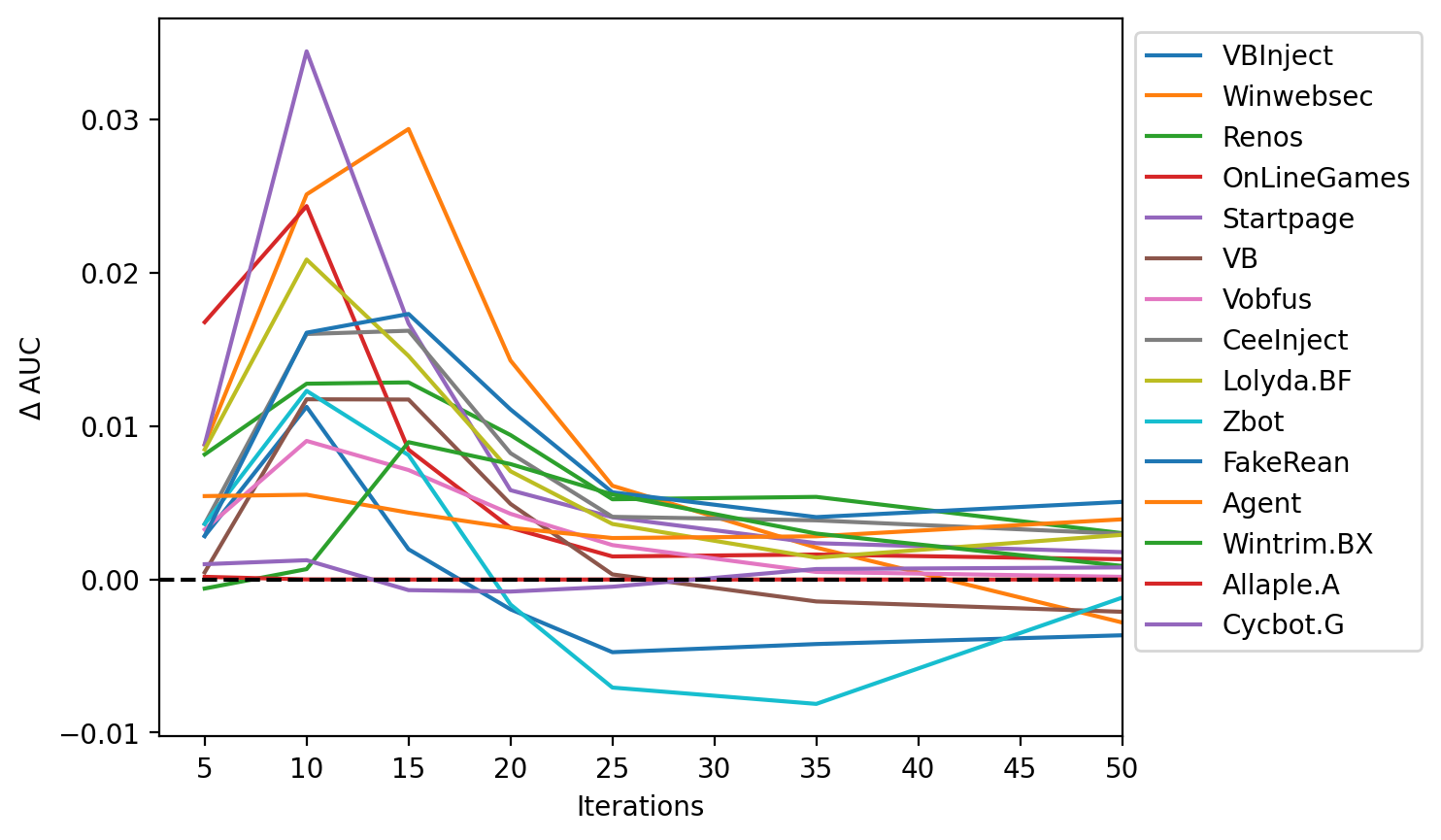}
    \caption{Change in AUC due to momentum at early iterations}
    \label{fig:malware_auc_diffs50}
\end{figure}

The difference in AUC for each family due to momentum is depicted in Figure~\ref{fig:malware_auc_diffs}.  At~10 iterations, all families show a net positive change in AUC with momentum.  Between~25 and~50 iterations, 
VBInject, Zbot, and VB show negative changes in model AUC.  All three families recover with no net difference in AUC by~100 iterations, possibly indicating that this could be the result of momentum overshooting.  On the other hand, Winwebsec consistently performs worse with momentum after~50 iterations.  Most 
families stabilize at a small net positive change in AUC at later iterations, with the exception of Winwebsec, Zbot, VB, and FakeRean.  Momentum produces a positive or neutral change in mean AUC for all other families throughout training.  The early AUC differences for each family can be viewed more clearly in Figure~\ref{fig:malware_auc_diffs50}.  

\begin{figure}[!htb]
\centering
\begin{subfigure}[t]{.385\textwidth}
    \captionsetup{justification=centering}
    \centering
    \includegraphics[width=1.0\linewidth]{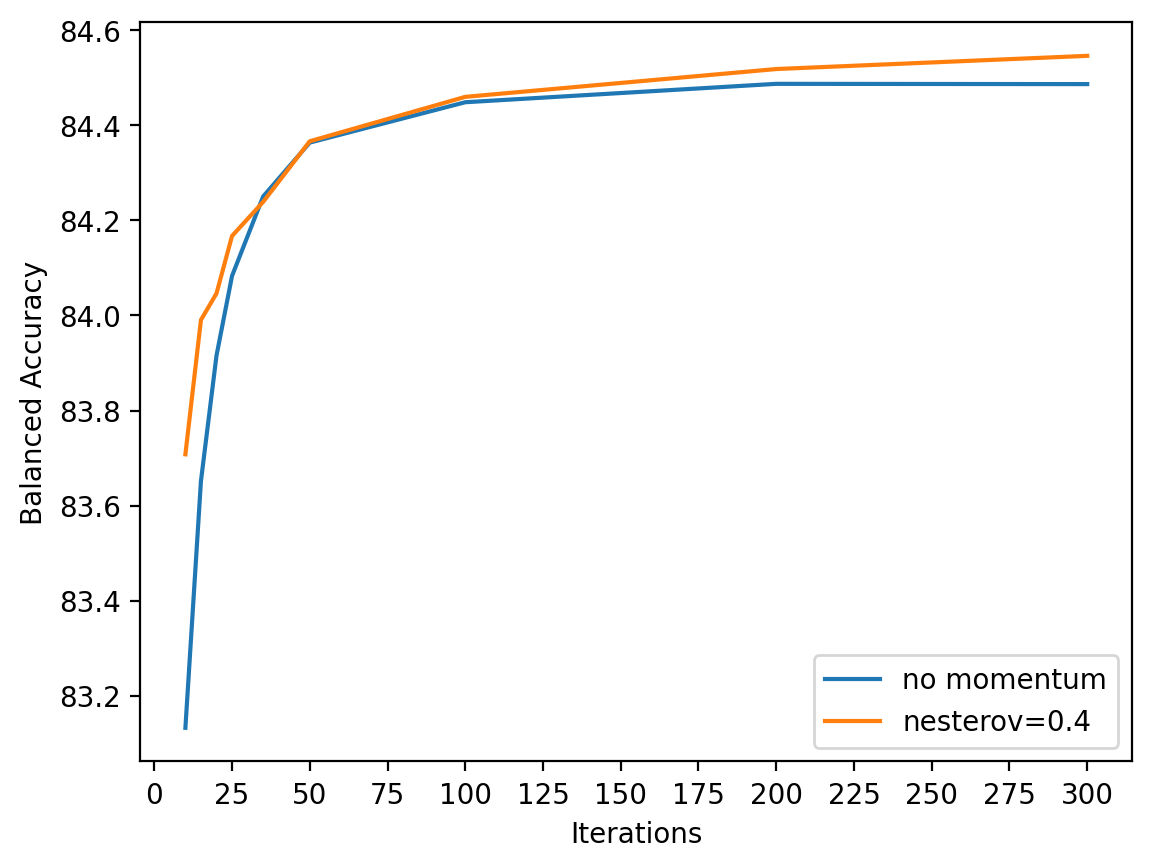}
\end{subfigure}
\begin{subfigure}[t]{.385\textwidth}
    \captionsetup{justification=centering}
    \centering
    \includegraphics[width=1.0\linewidth]{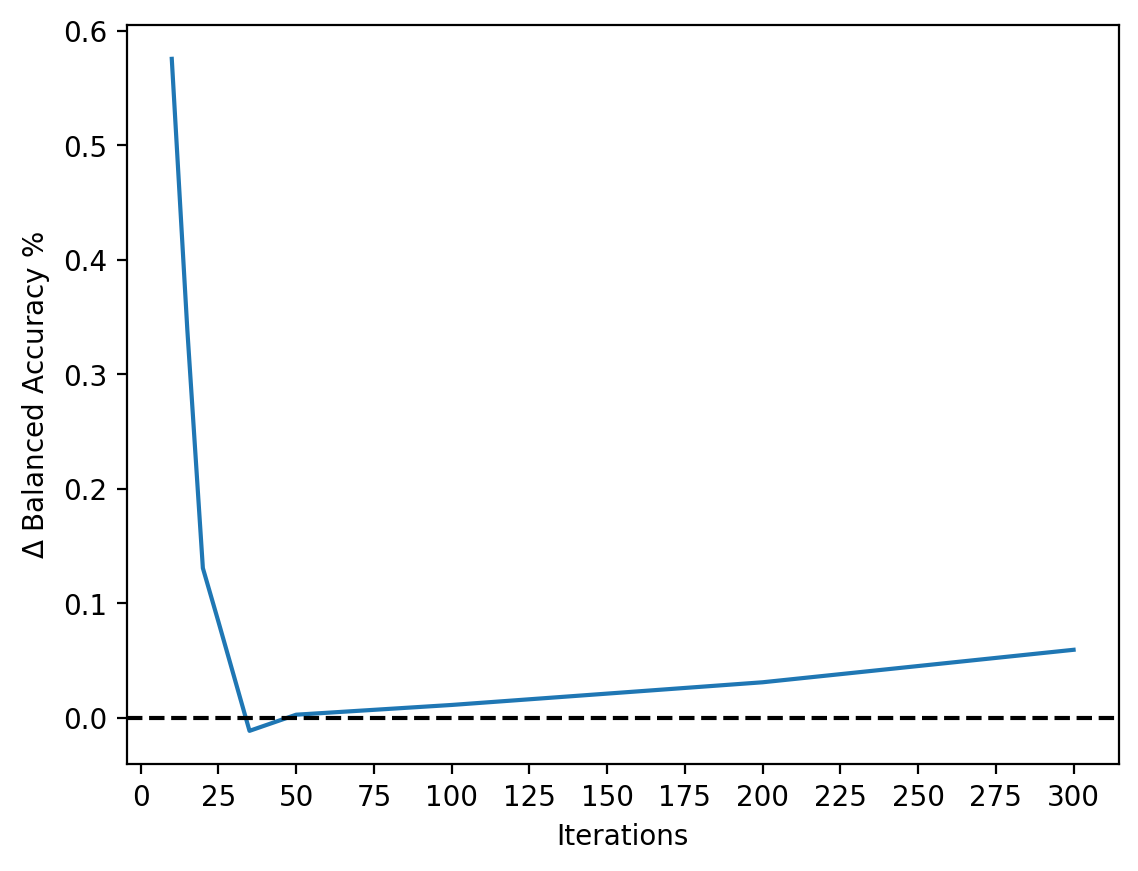}
\end{subfigure}
\caption{Comparison of SVM accuracy with and without momentum}
\label{fig:malware_svm_all}
\end{figure}

The change in SVM balanced accuracy with momentum is displayed in Figure~\ref{fig:malware_svm_all}.  
For baseline models without momentum, accuracy increases by a mere~1.36\%\ between~10 and~300 iterations.  This appears to indicate that despite the low AUC for many individual models, the combination of family scores is informative even for models trained for few iterations.  As with model scores and AUC, SVM accuracy shows the 
greatest increase at early iterations, declining to just below zero after~25 iterations.  The difference in accuracy shows continuous growth after~25 iterations, with a~0.1\%\ increase in accuracy due to momentum at~300 iterations.  
While minor, the small average score change with momentum at later iterations does seem to positively influence the SVM.  As models are only trained for~300 iterations, it is not clear for how many iterations this trend would continue.

\subsection{Discussion}\label{discussion}

In our extended malware experiments, both AUC and SVM metrics demonstrate an overall mean 
improvement in training speed of approximately~5 iterations, up to approximately~25 iterations.  
After that point, the difference in performance caused by momentum becomes negligible.  
Therefore, applications aiming to maximize performance would likely not benefit from momentum, 
assuming that a large number of iterations are performed with a large training dataset.  
However, in cases where time, computational power, or training data is severely limited, 
it may be desirable to train for fewer iterations at the cost of some model performance.  
For such applications, momentum appears to be able to decrease the number of necessary iterations
to achieve a given level of performance.  Based on the plateau examples with English text in 
Section~\ref{plateaus}, the longer it takes for a model without momentum to converge, 
the more potential there is for momentum to improve training speed.

\section{Conclusion}\label{chap:conclusion}

Our extensive experiments indicate that adding momentum to Baum-Welch re-estimation 
can be beneficial for training hidden Markov models in certain cases.  
In general, HMMs trained with momentum converged more quickly than HMMs trained without, 
leading to improved classification performance during early training iterations.  Momentum significantly 
reduced the number of iterations needed in cases where the model was slow to converge, 
such as our English text experiments, where the matrices were initialized close to~$1/N$.  

On the other hand, differences in model score and malware classification performance were 
negligible at higher numbers of iterations.  Momentum is therefore unlikely to be beneficial in 
applications aiming to maximize performance at any cost.  However, in cases involving limited resources,
or when training on limited data is necessary (e.g., the cold start problem), momentum can enable
us to train for fewer iterations at a lesser penalty in terms of model performance.  For such cases, 
momentum shows promise in reducing the number of required iterations while maintaining adequate 
performance.  

Gains from momentum were most significant for models with a relatively large number of hidden states, 
indicating that momentum may be more useful for more complex models.  Momentum did not seem to 
reduce the number of random restarts needed; while high momentum caused large jumps in parameter space, 
we found little evidence showing that new regions of the parameter space were being searched.  
In practice it would likely be beneficial to customize momentum based on behavior for each individual
model. For example, in our malware experiments, some customization for each malware family
would be needed to obtain optimal results from the use of momentum.

Future work involving momentum and HMMs could extend experiments to other HMM variants 
such Gaussian mixture hidden Markov models for continuous observations.  Momentum scheduling 
should also be be investigated in more depth, as learning rate scheduling is widely used in gradient descent, 
and changes in learning rate indirectly control momentum in such applications.  In particular, a proper adaptive 
schedule which dynamically adjusts momentum based on changes in score or some other metric could be used 
to smooth out training periods where momentum overshoots and underperforms, without requiring manual 
schedule tweaking.  It would also be beneficial to compare momentum to the parameterized EM algorithm 
discussed in Section~\ref{paramEM}.  With the proper selection of hyperparameters, parameterized EM 
functions similar to momentum, but it only takes into account the previous timestep, rather than a weighted
sum of multiple previous timesteps.  
Future experiments with momentum should also involve other types of data and HMM applications to 
ensure the conclusions found in this paper hold more broadly.

\bibliographystyle{plain}
\bibliography{references.bib}

\end{document}